\documentclass[manuscript,screen]{acmart}

\usepackage{booktabs}
\usepackage{pifont}
\usepackage{xcolor}
\usepackage{multirow}
\usepackage{array}
\usepackage{tabularx}
\usepackage{pifont}
\usepackage{enumitem}
\usepackage{graphicx}
\usepackage{wrapfig}       
\usepackage{caption}        

\usepackage{xcolor,colortbl}
\usepackage{tcolorbox}
\newtcolorbox{findingsbox}[1]{
    title={#1},
    fonttitle=\bfseries,
    fontupper=\small,
    colback=black!2,
    colbacktitle=black!7,
    coltitle=black,
    colframe=black!30,
    boxrule=0.4pt,
    arc=1pt,
    left=5pt,
    right=5pt,
    top=4pt,
    bottom=4pt,
    before skip=6pt,
    after skip=6pt
}

\AtBeginDocument{%
  }

\setcopyright{acmlicensed}
\copyrightyear{2018}
\acmYear{2027}
\acmDOI{}
\acmISBN{978-1-4503-XXXX-X/2018/06}

\makeatletter
\g@addto@macro\@titlefont{%
  \fontsize{11pt}{13pt}\selectfont
}
\makeatother
\AtBeginDocument{%
  \setlength{\skip\footinscopyrightpermission}{6pt}%
}
\begin{document}

\title{A Comprehensive Survey of Agent Distillation across Models, Artifacts, and Harnesses}

\author{Ziluowen Luo}
\affiliation{%
  \institution{Central South University}
  \country{China}
}
\orcid{0009-0002-2817-673X}
\email{lzlwddl@csu.edu.cn}

\author{Senzhang Wang}
\correspondingauthor
\affiliation{%
  \institution{Central South University}
  \country{China}
}
\orcid{0000-0002-3615-4859}
\email{szwang@csu.edu.cn}

\author{Chaozhuo Li}
\correspondingauthor
\affiliation{%
  \institution{Beijing Academy of Artificial Intelligence}
  \country{China}
}
\orcid{0000-0002-8179-7503}
\email{lichaozhuo@bupt.edu.cn}

\author{Jun Yin}
\affiliation{%
  \institution{Hong Kong Polytechnic University}
  \country{China}
}
\orcid{0009-0004-6714-3476}
\email{Junmay.yin@connect.polyu.hk}

\author{Hao Yan}
\affiliation{%
  \institution{Central South University}
  \country{China}
}
\orcid{0009-0007-7631-9375}
\email{csuyh1999@csu.edu.cn}

\author{Ming Cheng}
\affiliation{%
  \institution{Central South University}
  \country{China}
}
\orcid{0009-0003-8873-3021}
\email{chengming@csu.edu.cn}
            
\author{Chenxu Wang}
\affiliation{%
  \institution{Beijing University of Posts and Telecommunications}
  \country{China}
}
\orcid{0009-0007-5013-1972}
\email{chenxuwang@bupt.edu.cn}

\author{Songyang Liu}
\affiliation{%
  \institution{Beijing University of Posts and Telecommunications}
  \country{China}
}
\orcid{0009-0003-9592-8046}
\email{liusyang@bupt.edu.cn}



\author{Litian Zhang}
\affiliation{%
  \institution{Beijing University of Posts and Telecommunications}
  \country{China}
}
\orcid{0000-0002-6981-3873}
\email{litianzhang@bupt.edu.cn}

\author{Qiwei Ye}
\affiliation{%
  \institution{Beijing Academy of Artificial Intelligence}
  \country{China}
}
\orcid{0000-0003-4264-5846}
\email{qwye@baai.ac.cn}

\author{Zheng Liu}
\affiliation{%
  \institution{Beijing Academy of Artificial Intelligence}
  \country{China}
}
\orcid{0000-0001-7765-8466}
\email{zhengliu1026@gmail.com}


\author{Chengqi Zhang}
\affiliation{%
  \institution{Hong Kong Polytechnic University}
  \country{China}
}
\orcid{0000-0001-5715-7154}
\email{chengqi.zhang@polyu.edu.hk}

\author{Philip S. Yu}
\affiliation{%
  \institution{University of Illinois at Chicago}
  \country{USA}
}
\orcid{0000-0002-3491-5968}
\email{psyu@uic.edu}
\renewcommand{\shortauthors}{Luo et al.}

\authorsaddresses{%
Authors' Contact Information:
Ziluowen Luo (lzlwddl@csu.edu.cn),
Senzhang Wang (szwang@csu.edu.cn),
Hao Yan (csuyh1999@csu.edu.cn), and
Ming Cheng (chengming@csu.edu.cn),
Central South University, Changsha, Hunan, China;
Chaozhuo Li (lichaozhuo@bupt.edu.cn),
Qiwei Ye (qwye@baai.ac.cn), and
Zheng Liu (zhengliu1026@gmail.com),
Beijing Academy of Artificial Intelligence, Beijing, China;
Jun Yin (Junmay.yin@connect.polyu.hk) and
Chengqi Zhang (chengqi.zhang@polyu.edu.hk),
Hong Kong Polytechnic University, Hong Kong, China;
Chenxu Wang (chenxuwang@bupt.edu.cn),
Songyang Liu (liusyang@bupt.edu.cn), and
Litian Zhang (litianzhang@bupt.edu.cn),
Beijing University of Posts and Telecommunications, Beijing, China;
Philip S. Yu (psyu@uic.edu),
University of Illinois at Chicago, Chicago, Illinois, USA
}



\begin{abstract}

Knowledge distillation provides a teacher--student framework for transferring capabilities at reduced cost and has been widely studied for large language models (LLMs).
As LLM-based agents move from response generation to sustained, closed-loop task solving, distillation expands beyond model parameters to knowledge retained in reusable resources and execution workflows.
This shift from \textit{model-level} to \textit{agent-level} knowledge transfer supports efficient deployment, broader capability access, and continual self-evolution.
However, existing surveys largely examine model distillation and agent system design separately, limiting comparison of transfer strategies across agent components.
To bridge this gap, we formalize \textbf{\textit{Agent Distillation}} as persistent knowledge transfer between teacher and student agents and present its first unified survey.
Distinguishing experience used for transfer from knowledge retained for reuse, we develop a \textit{knowledge-substrate} taxonomy based on where knowledge persists: \textit{\textbf{models, artifacts, or harnesses}}.
Building on this taxonomy, we analyze cross-substrate transfer paths and hybrid strategies.
To assess the resulting capabilities, we review benchmarks and evaluation protocols for retention, causal attribution, and deployed utility without further teacher support.
This survey provides a unified basis for studying agent knowledge transfer and developing efficient, adaptable, and self-evolving agents. 

\end{abstract}

\ccsdesc[500]{General and reference~Surveys and overviews}
\ccsdesc[500]{Computing methodologies~Artificial intelligence}


\keywords{Large language model, LLM-based Agent, Knowledge Distillation}

\received{20 February 2007}
\received[revised]{12 March 2009}
\received[accepted]{5 June 2009}

\setcopyright{none}
 \maketitle
\section{Introduction}

Large language model (LLM)-based agents extend language models from response generators into systems capable of sustained task solving and interaction with external environments~\cite{schick2023toolformer}.  
During execution, an agent reasons over its current state, takes actions through tools or other artifacts, observes the resulting feedback, and revises its subsequent decisions~\cite{kang2025distillingllmagent,harnessbench2026}. 
This transition toward sustained, closed-loop interaction characterizes a broad range of agentic systems, including prominent coding agents (e.g., OpenAI Codex~\cite{openai_codex}, Anthropic Claude Code~\cite{anthropic_claude_code} and OpenClaw~\cite{openclaw2026}), web agents~\cite{zhou2024webarena,jimenez2024swebench} and general-purpose agents~\cite{wang2024autonomous,xi2023rise}. 
Consequently, the competence underlying successful task solving is increasingly expressed through the execution process rather than through individual model responses alone.

However, the knowledge supporting such competence is often coupled to the particular agent realization and task-solving context in which it emerges~\cite{shang2025agentsquare,zhuge2024gptswarm}. 
The knowledge underlying a successful execution may remain distributed across model behavior, accumulated artifacts and runtime decisions. 
Consequently, another agent may need to rediscover similar strategies at substantial cost through repeated reasoning and interaction. 
This raises a broader question: 
\emph{how can task-solving knowledge acquired by one agent be incorporated into another and retained for future use?}

\subsection{Motivation: From Model Distillation to Agent Distillation}

Although knowledge distillation offers a natural teacher--student framework for addressing this question, existing formulations operate primarily at the model level: both the teacher and student are treated as models, with the transferred knowledge ultimately absorbed into the parameters of the student~\cite{hinton2015distilling,gou2021knowledge}. 
For agentic task solving, the unit of transfer expands from the model to the complete agent realization. 
Accordingly, distillation broadens from \textbf{\textit{model-to-model}} to \textbf{\textit{agent-to-agent}} knowledge transfer, with the aim of turning teacher experience into reusable student capabilities. 

\begin{wrapfigure}[13]{r}{0.6\textwidth}
\centering
\includegraphics[width=0.95\linewidth]{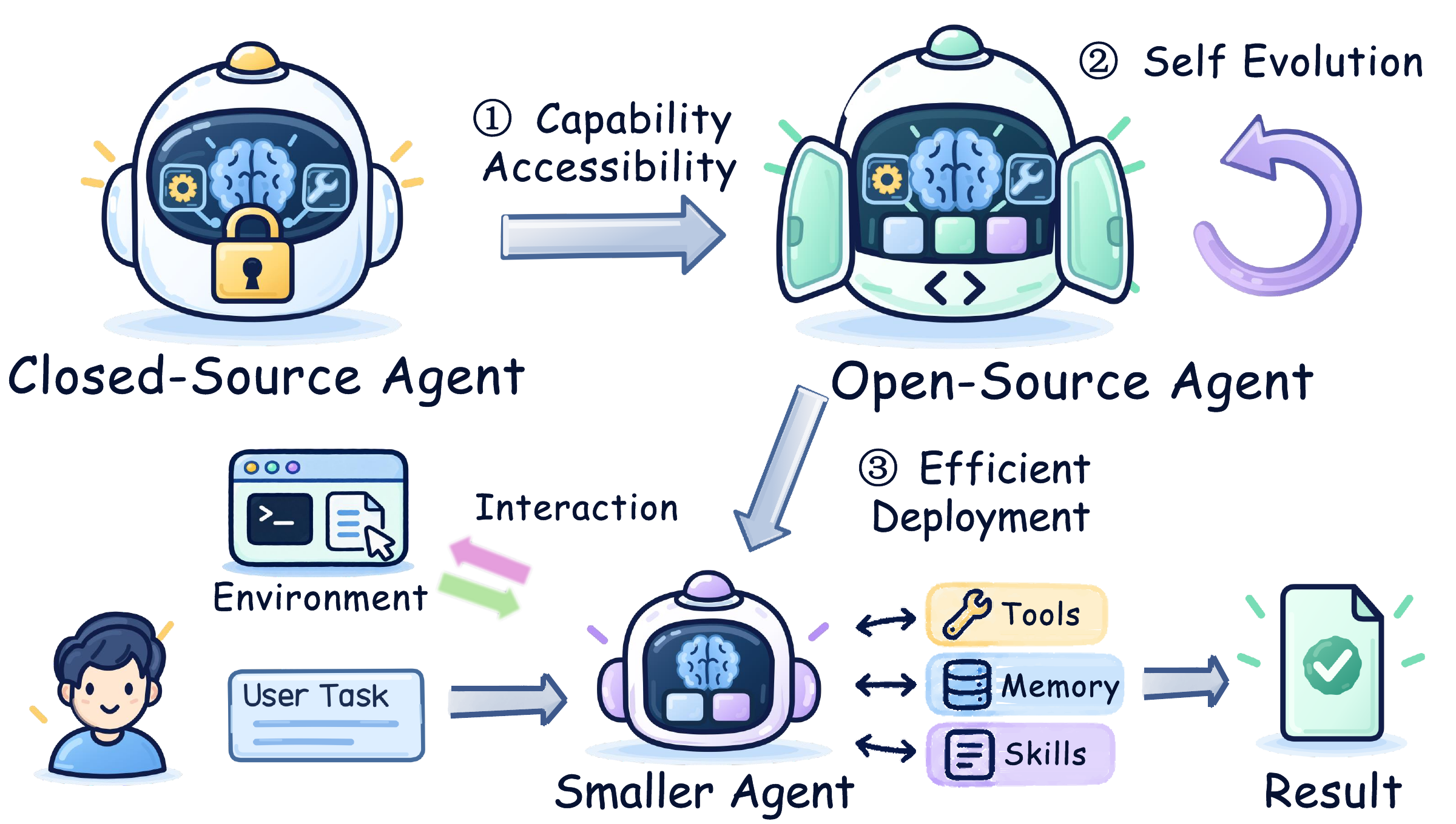} 
\caption{ From a successful agent to broader reuse of task-solving knowledge. }
\label{fig:motivation}
\end{wrapfigure}

As illustrated in Figure~\ref{fig:motivation}, such knowledge reuse serves three complementary needs, extending the role of distillation beyond model compression. 
It retains the familiar goal of \textit{efficient deployment}: knowledge from a high-capability teacher guides a compact student toward comparable task-solving competence at a lower deployment cost. 
Beyond efficiency, it expands \emph{capability accessibility} by incorporating knowledge revealed through a closed-source teacher's behavior into an open-source student without accessing the internal of the teacher~\cite{mukherjee2023orca,zhang2024xlam}. 
More distinctively, agent-level distillation supports \textit{self-evolution}.
By incorporating knowledge from prior experience into successive realizations, an agent converts isolated task-solving episodes into continual improvement~\cite{sima2025sima2,zhang2025darwin}. 
Together, these needs highlight the broader role of teacher--student transfer in reusing knowledge and improving agent competence. 

In this work, we use \textbf{\textit{Agent Distillation}} to denote the persistent incorporation of teacher-derived task-solving knowledge into a student agent, with the aim of improving the student's competence. 
Such knowledge may be incorporated into different substrates of the student realization rather than into model parameters alone. 
This broader view motivates a systematic investigation of agent-centered distillation beyond conventional model-centered distillation.

\newcommand{\cmark}{\textcolor{green!60!black}{\ding{51}}}
\newcommand{\xmark}{\textcolor{red!75!black}{\ding{55}}}
\newcommand{\pmark}{\textcolor{orange!80!black}{\ding{108}}}

\newcolumntype{Y}{>{\centering\arraybackslash}p{1.5cm}}
\newcolumntype{L}{>{\centering\raggedright\arraybackslash}X}

\begin{table*}[tbp]
\centering
\caption{
\textbf{Comparison with representative surveys from the distillation and agent literatures.} 
\cmark~denotes systematic treatment as a primary organizing perspective,
\pmark~denotes partial treatment,
and \xmark~indicates that the aspect is outside the scope of the survey.
}
\label{tab:survey_comparison}

\setlength{\tabcolsep}{5pt}
\renewcommand{\arraystretch}{1.15}
\begin{tabularx}{\textwidth}{
    @{}
    p{2.8cm}
    >{\centering\arraybackslash}p{0.7cm}
    Y
    Y Y Y
    Y
    Y
    @{}
}
\toprule

\multirow{2}{*}{\textbf{Survey}}
&
\multirow{2}{*}{\textbf{Year}}
&
\multirow{2}{*}{\textbf{T$\rightarrow$S}}
&
\multicolumn{3}{c}{\textbf{Substrate Coverage}}
&
\multirow{2}{*}{\shortstack{\textbf{Cross-}\\\textbf{Substrate}}}
&
\multirow{2}{*}{\shortstack{\textbf{Transfer}\\\textbf{Evaluation}}}
\\

\cmidrule(lr){4-6}

&
&
&
\textbf{Model}
&
\textbf{Artifact}
&
\textbf{Harness}
&
&
\\

\midrule

\multicolumn{8}{@{}l}{
    \textbf{\textit{Distillation-related Surveys}}
}
\\[-2pt]

\citeauthor{gou2021knowledge}~\cite{gou2021knowledge}
& 2021
& \cmark & \cmark & \xmark & \xmark & \xmark & \pmark \\

\citeauthor{xu2024llmkd}~\cite{xu2024llmkd}
& 2024
& \cmark & \cmark & \xmark & \xmark & \xmark & \pmark \\

\citeauthor{yang2025llmkd}~\cite{yang2025llmkd}
& 2025
& \cmark & \cmark & \xmark & \xmark & \xmark & \cmark \\

\citeauthor{mansourian2025comprehensive}~\cite{mansourian2025comprehensive}
& 2025
& \cmark & \cmark & \xmark & \xmark & \xmark & \pmark \\

\addlinespace[2pt]
\midrule

\multicolumn{8}{@{}l}{
    \textbf{\textit{Agent-related Surveys}}
}
\\[-2pt]

\citeauthor{wang2024autonomous}~\cite{wang2024autonomous}
& 2024
& \xmark & \cmark & \pmark & \pmark & \xmark & \xmark \\


\citeauthor{zhang2025memory}~\cite{zhang2025memory}
& 2025
& \xmark & \pmark & \cmark & \pmark & \xmark & \xmark \\

\citeauthor{guo2024multiagent}~\cite{guo2024multiagent}
& 2024
& \xmark & \pmark & \pmark & \cmark & \xmark & \xmark \\

\citeauthor{xu2025tool}~\cite{xu2025tool}
& 2025
& \xmark & \cmark & \xmark & \cmark & \xmark & \xmark \\

\citeauthor{gao2026selfevolving}~\cite{gao2026selfevolving}
& 2026
& \xmark & \cmark & \cmark & \cmark & \pmark & \xmark \\

\citeauthor{guo2026harness}~\cite{guo2026harness}
& 2026
& \xmark & \cmark & \pmark & \cmark & \pmark & \xmark \\

\addlinespace[2pt]
\midrule

\textbf{Ours}
& 2027
& \cmark & \cmark & \cmark & \cmark & \cmark & \cmark \\

\bottomrule
\end{tabularx}
\end{table*}

\subsection{Comparison with Existing Surveys}

\textbf{Model-Centered Distillation.} 
Existing surveys provide a well-established account of teacher-student knowledge transfer. 
Classical surveys organize this literature around the forms of transferred knowledge, distillation objectives, teacher--student architectures, and training strategies~\cite{gou2021knowledge}. 
With the rise of large language models, recent surveys further broaden this scope to generated supervision, reasoning traces, instruction data, preference signals, and evaluation protocols~\cite{xu2024llmkd,yang2025llmkd}. 
These studies therefore offer a rich view of \emph{what} knowledge can be extracted from a teacher and \emph{how} it can be incorporated into a student. 
Their organizing perspective, however, remains predominantly \emph{model-centered}: the student model serves as both the primary recipient of distilled knowledge and the basic unit of evaluation.

\textbf{System-Centered Agent Research.} 
Agent-related surveys provide a complementary \emph{system-centered} perspective. 
They examine agentic behavior from a system perspective, focusing on how agent components jointly support task execution~\cite{wang2024autonomous,xi2023rise}.  
Research along this line covers planning and tool learning~\cite{huang2024planning,xu2025tool}, memory and multi-agent coordination~\cite{zhang2025memory,guo2024multiagent}, and agent evaluation~\cite{liu2023agentbench,yehudai2026survey}.
More recent harness-oriented work further shifts attention toward workflows and dynamic execution process, emphasizing how models, tools, memory, control logic, and execution structures are composed into effective agent systems~\cite{guo2026harness}. 
Collectively, these surveys provide an increasingly comprehensive account of how agentic competence is constructed, organized, and evaluated at the system level. 
Their primary concern, however, is not the persistent transfer of task-solving knowledge through a teacher--student relation.

\textbf{The Missing Bridge.} 
Taken together, these two bodies of literature reveal a clear gap: \textit{a unified study for agent-level teacher--student knowledge transfer is still missing.} 
Distillation surveys are transfer-centric but predominantly model-centered, whereas agent surveys are system-centered but do not organize the literature around how knowledge is transferred into and retained by a complete student agent. 
Furthermore, table~\ref{tab:survey_comparison} summarizes this distinction and positions our survey at the intersection of these two bodies of literature.

\begin{figure*}[tbp]
    \centering
    \includegraphics[width=0.95\linewidth]{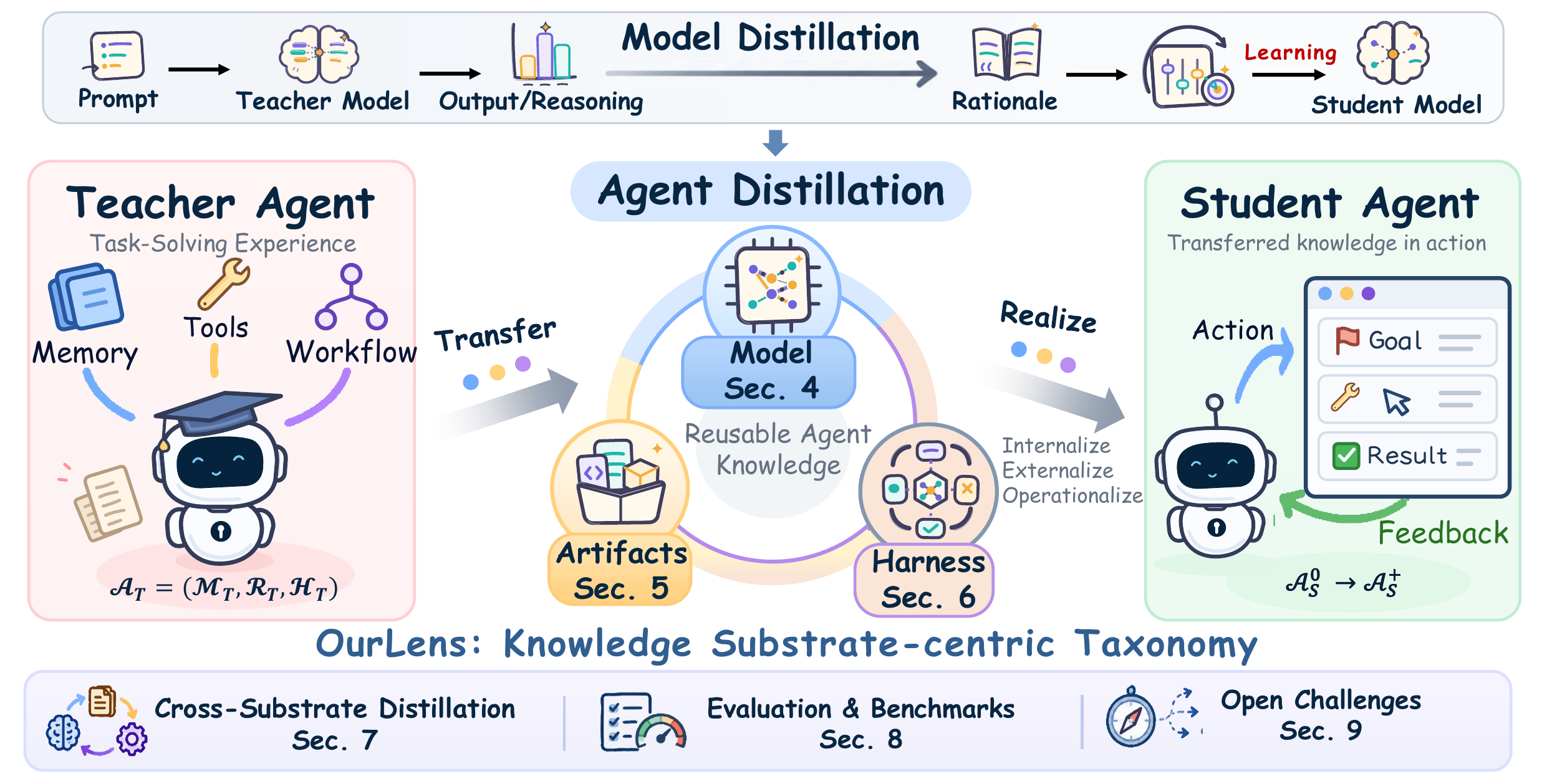}
    \caption{
\textbf{From model-centered to agent-centered knowledge transfer.} 
Agent Distillation treats the complete agent realization as the unit of teacher--student transfer. 
Our knowledge substrate-centric view organizes existing approaches according to the primary form in which transferred knowledge persists in the student: the parametric \emph{model}, reusable \emph{artifacts} or the execution \emph{harness}.
}
    \label{fig:fig1_overview}
    \vspace{-0.5cm}
\end{figure*}

\subsection{Our Perspective and Contributions}

To address this gap, we adopt a knowledge substrate centric perspective on Agent Distillation. 
Our central question is where and in what persistent form teacher-derived task-solving knowledge becomes available to the student. We use \textbf{\emph{knowledge substrate}} to denote the persistent carrier within an agent realization through which such knowledge is encoded, retained and made available to future. 
A substrate is characterized jointly by how the knowledge is encoded and how it participates in future agent execution. 
This perspective provides a common basis for analyzing knowledge that is \textit{internalized} in learned computation, \textit{externalized} as reusable objects or \textit{operationalized} through execution logic.

Under this view, we distinguish three principal knowledge substrates.
\textbf{\textit{Parametric models}} ($\mathcal{M}$) retain knowledge in learned computation that directly shapes inference or reasoning. 
\textbf{\textit{Reusable artifacts}} ($\mathcal{R}$) preserve knowledge as independently addressable objects, such as memories, skills, and tools, that can be retrieved or invoked across episodes. 
\textbf{\textit{Execution harnesses}} ($\mathcal{H}$) encode knowledge in the procedures and control structures that organize execution, including routing, verification, recovery, and orchestration. 
These substrates form our taxonomy, which organizes methods by where teacher-derived knowledge primarily persists in the student. 
Building on this taxonomy, we further review transfer mechanisms, cross-substrate paths, and evaluation principles, leading to the following main contributions:
\begin{enumerate}[leftmargin=*,noitemsep]

    \item \textbf{Definition and Scope.}
    To the best of our knowledge, we establish the first unified definition of Agent Distillation at the agent level, centered on persistent teacher-derived knowledge incorporation and briefly clarify its scope.

    \item \textbf{Knowledge-Substrate Taxonomy.}
    We develop a substrate-centered taxonomy that organizes existing approaches according to the primary substrate in which knowledge form persists after transfer: models, artifacts or harnesses.

    \item \textbf{Cross-Substrate Transfer.}
    Beyond the three primary branches, we analyze how agent knowledge can migrate across substrates and how multiple substrates can be combined within hybrid distillation strategies.

    \item \textbf{Evaluation and Research Frontiers.}
    We investigate how to assess knowledge retention, causal contribution, and deployed utility and identify open challenges and research priorities for advancing reusable agent capabilities.

\end{enumerate}

The remainder of this survey is organized as follows. 
\S~\ref{sec:2} establishes the agent formulation, defines Agent Distillation and its scope, and describes our survey methodology. 
\S~\ref{sec:3} introduces the knowledge-substrate taxonomy used to organize the literature. 
\S~\ref{sec:4} reviews knowledge internalization in parametric models, \S~\ref{sec:5} externalization as reusable artifacts, and \S~\ref{sec:6} operationalization through execution harnesses. 
\S~\ref{sec:7} analyzes cross-substrate transfer paths and hybrid realizations that retain knowledge in multiple student components. 
\S~\ref{sec:8} reviews evaluation protocols and benchmarks for retention, causal contribution, and deployed utility. 
Finally, \S~\ref{sec:9} discusses open challenges and future directions in attribution, adaptation, and knowledge maintenance, and \S~\ref{sec:10} concludes the survey. 
\section{Foundations and Scope of Agent Distillation}\label{sec:2}

To establish a common architectural foundation, this section clarifies what constitutes Agent Distillation and delineates its scope. 
In the following, we first introduce the basic concepts of agentic task solving and system realization (\S~\ref{sec:2.1}), followed by a running example that traces the progression from teacher execution to persistent student realization (\S~\ref{sec:2.2}). 
Building on these foundations, we then formalize Agent Distillation and clarify its boundaries (\S~\ref{sec:2.3}), and systematize the literature collection and analysis methodology in our investigation (\S~\ref{sec:2.4}).

\subsection{Agent Realizations and Task Solving}\label{sec:2.1}
Consistent with prior work, we view an agent as a goal-directed system that repeatedly observes and acts within an environment to accomplish a task \cite{russell2020artificial}. 
Recent work on LLM-based agents extends this interaction-centered view by incorporating reasoning, planning, memory, tool use, and runtime control \cite{yao2023react,wang2024autonomous,sumers2024cognitive}. 
Prior studies further show that agent behavior can change substantially with the dynamic harness even when the underlying model remains fixed \cite{yang2024sweagent,kapoor2025agents}. 
We therefore take the system as the basic unit of Agent Distillation and refer to it as an \emph{agent realization}:
\begin{equation}
    \mathcal{A}=(\mathcal M,\mathcal R,\mathcal  H), 
\end{equation}
where $\mathcal M$, $\mathcal R$, and $\mathcal H$ denote its parametric \textbf{\textit{M}}odel, reusable a\textbf{\textit{R}}tifacts and execution \textbf{\textit{H}}arness, respectively.

\paragraph{\textbf{Agent Components}.} 
The three components capture complementary aspects of an agent. 
The model component $\mathcal M$ comprises the parametric models that generate reasoning, decisions, or action proposals. 
The artifacts component $\mathcal R$ comprises reusable objects external to $\mathcal M$ that can be retrieved, interpreted, or invoked across episodes, such as memories, demonstrations, instructions, skill descriptions, and programs. 
The harness $\mathcal H$ comprises the runtime mechanisms that assemble context, coordinate models, invoke tools and control workflows, including verification, retry, and recovery. 
This functional decomposition  need not correspond to architecturally or physically separate modules.

\paragraph{ \textbf{Task and Agent–Environment Interaction.} } 
To characterize the capability of an agent realization, we specify both the task it seeks to accomplish and the environment with which it interacts. 
Let $q \sim \mathcal{P}$ denote a concrete task instance drawn from a task distribution $\mathcal{P}$, where $q$ specifies the objective, task constraints and criteria for successful completion. 
To pursue $q$, an agent realization $\mathcal{A}$ interacts with a stateful environment $\mathcal{E}$, whose state at step $t$ is denoted by $e_t$. 
Thus, $q$ specifies \emph{what} the agent should achieve, whereas $\mathcal{E}$ supplies the external state, dynamics, and feedback under which that objective is pursued. 
Given the interaction setting $(\mathcal{A},q,\mathcal{E})$, the interaction begins from an initial environment state $e_0$, from which $\mathcal{E}$ exposes an initial observation $o_0$ to $\mathcal{A}$. 
At step $t$, the agent need not observe the complete state $e_t$; instead, it conditions its next action on the task and the observable interaction history:
\begin{equation}
    h_t =
    \left(
        q,o_0,a_0,o_1,\ldots,a_{t-1},o_t
    \right),
    \label{eq:interaction-history}
\end{equation}
where $o_i$ is the observation available before the agent acts at step $i$, and $a_i$ is the action selected in response. 
Conditioned on $h_t$, the realization induces an action policy $\pi_{\mathcal{A}}$, while the environment evolves according to a transition function $\mathcal{T}_{\mathcal{E}}$:
\begin{equation}
    a_t \sim \pi_{\mathcal{A}}(\cdot \mid h_t),
    \qquad
    (e_{t+1},o_{t+1})
    \sim
    \mathcal{T}_{\mathcal{E}}(\cdot \mid e_t,a_t).
    \label{eq:agent-environment-interaction}
\end{equation}
The policy $\pi_{\mathcal{A}}$ reflects the joint contribution of $\mathcal{M}$, $\mathcal{R}$, and $\mathcal{H}$ to the agent-side decision, whereas $\mathcal{T}_{\mathcal{E}}$ captures how the environment converts that decision into a state transition and a new observation. 
Both mappings may also be stochastic. 
The action $a_t$ and the resulting observation $o_{t+1}$ extend $h_t$ to $h_{t+1}$, providing the context for the next interaction step. 
Furthermore, repeating this decision--transition cycle produces the task-solving episode characterized next.

\paragraph{\textbf{Execution Episode and Trace.}}
A task-solving episode comprises a finite sequence of interactions that begins with $o_0$ and terminates at step $L$ upon task completion, failure, or exhaustion of the available resource budget. 
Because agent actions may modify the environment in ways that are not fully represented by subsequent observations, we record the environmental change associated with each transition as 
\begin{equation} 
\Delta e_{t+1} = \operatorname{Diff}_{\mathcal{E}} \left(e_t,e_{t+1}\right), 
\label{eq:environment-delta} 
\end{equation} 
where $\operatorname{Diff}_{\mathcal{E}}$ is a domain-specific comparison operator for environment states rather than numerical subtraction. 
Let $B$ denote the resource budget available to the agent and we then represent the episode by the execution trace 
\begin{equation}
    \tau(\mathcal{A},q,\mathcal{E};B)
    =
    \left(
        o_0,
        \left(
            a_t,o_{t+1},\Delta e_{t+1}
        \right)_{t=0}^{L-1}
    \right).
    \label{eq:execution-trace}
\end{equation}
The trace records the actions taken, received observations and environmental changes produced during one execution. 
It therefore describes what occurred in a particular episode, but not how the agent performs across tasks and executions.

\paragraph{ \textbf{Agentic Competence.} }
Agentic competence lifts this episode-level view to the expected task-solving performance of a complete agent realization across a task distribution. 
Let $B$ denote the resource budget available for an episode, and write $\tau(\mathcal{A},q,\mathcal{E};B)$ for the trace generated under that budget. 
Given a utility function $u(q,\tau)$ that evaluates a trace with respect to its task, we operationalize agentic competence as: 
\begin{equation} 
\operatorname{Comp} \left( \mathcal{A}; \mathcal{P}, \mathcal{E}, B \right) = 
\mathbb{E}_{q \sim \mathcal{P}} \left[ u\left( q, \tau(\mathcal{A},q,\mathcal{E};B) \right) \right]. \label{eq:agentic-competence} 
\end{equation} 
Here, $u$ account for task success and resource efficiency. 
For stochastic agents or environments, the expectation may additionally be estimated over repeated executions. Moreover, competence is a property of the complete realization under $(\mathcal{P},\mathcal{E},B)$, not an object contained in any individual trace. 
Individual traces provide episode-level evidence of competence, but they are neither competence itself nor automatically reusable knowledge. 
This distinction allows teacher and student to be defined by their roles in a transfer process rather than by any particular output or trajectory.

\subsection{A Running Example of Agent Distillation}\label{sec:2.2}
\label{sec:running-example}
\begin{figure*}[tbp]
    \centering
    \includegraphics[width=1.0\linewidth]{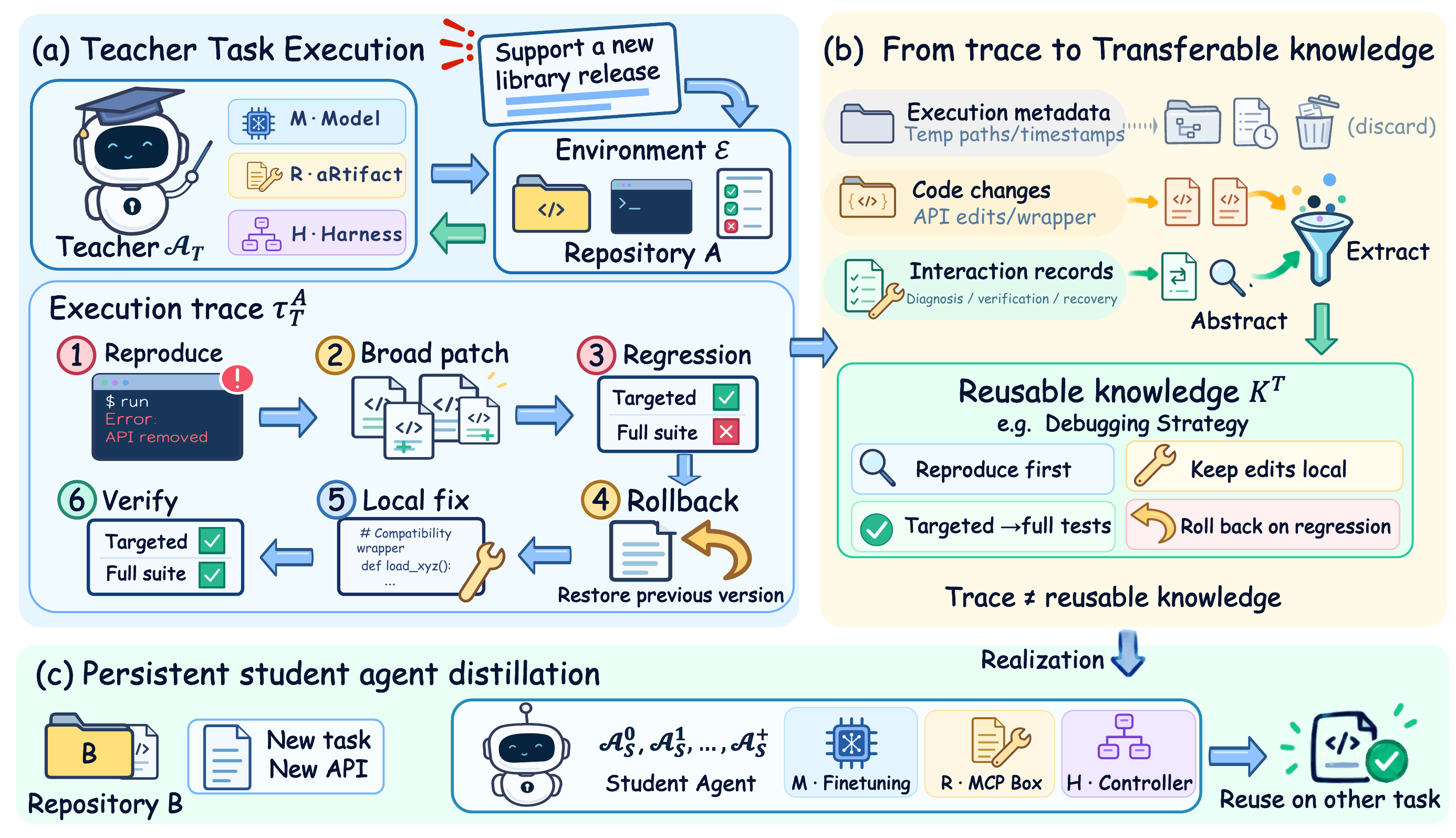}
    \caption{
\textbf{A running example of Agent Distillation.} 
(a) A teacher coding agent repairs a repository by recovering from an over-broad patch and verifying localized fix. 
(b) The trace mixes incidental residue, task-specific information and reusable patterns; filtering and abstraction turn selected evidence into a debugging strategy. 
(c) Then it is realized in student agents and reused on other tasks.
}
    \label{fig:fig3_toy_example}
\end{figure*}

To ground the preceding concepts in a concrete process, we use a software maintenance task as a running example to provide more intuitions. 
Consider a teacher coding agent $\mathcal{A}_T=(\mathcal{M}_T,\mathcal{R}_T,\mathcal{H}_T)$ assigned a source task $q_A$ in Repository~A. 
The task is to support a new release of a third-party library while preserving the repository's intended behavior intended behavior and passing the full test suite. 
The corresponding environment $\mathcal{E}_A$ contains the repository state, installed dependencies, and test infrastructure. 
Through its available tools, the teacher can inspect files, execute commands, modify code or dependencies, and run tests. 
Under a budget $B_T$, this whole interaction produces the execution trace: 
\begin{equation} 
\tau_T^A = \tau\left( \mathcal{A}_T, q_A, \mathcal{E}_A; B_T \right). 
\label{eq:teacher-example-trace} 
\end{equation} 
Figure~\ref{fig:fig3_toy_example} follows this example from teacher execution, through knowledge extraction and transformation to persistent student agent realization and reuse. 
The teacher completes $q_A$ through the six-step process illustrated in Figure~\ref{fig:fig3_toy_example}(a).

\begin{findingsbox}{A toy example of the software maintenance task by Agents. }
\noindent [Step 1] \textbf{Reproduce}:
The teacher runs the test suite and traces the failure to a library API removed in the new release.

\noindent [Step 2] \textbf{Broad patch}:
The teacher replaces the removed API with its apparent successor throughout Repository~A.

\noindent [Step 3] \textbf{Regression}:
A targeted test passes, but the full suite exposes a regression caused by the broad patch.

\noindent [Step 4] \textbf{Rollback}:
The teacher reverts the broad patch and restores the repository to its pre-patch state.

\noindent [Step 5] \textbf{Local fix}:
The teacher adds compatibility wrapper that supports the new API, preserving existing behavior.

\noindent [Step 6] \textbf{Verify}:
Both targeted and full-suite tests pass, confirming compatibility with the new release.

\end{findingsbox}

\paragraph{\textbf{From trace to transferable knowledge}} 
The actions, observations, and environmental changes produced during the six steps form the teacher agent execution trace $\tau_T^A$. 
In this example, the trace serves as a teacher-side source from which transferable knowledge can be derived, as illustrated in Figure~\ref{fig:fig3_toy_example}(b). 
The trace contains incidental metadata, task-specific code changes, and interaction records that capture diagnosis, verification, and recovery. 
Selection removes transient or irrelevant residue, while abstraction separates reusable problem-solving principles from repository-specific details. 
The resulting transferable knowledge, denoted by $K_T$, takes the form of a debugging strategy: reproduce the failure first, keep edits local, proceed from targeted to full-suite testing, and roll back when a regression is detected. 

For this knowledge to constitute Agent Distillation, it must be persistently realized in the student agent. 
Starting from $\mathcal{A}_S^0$, the debugging strategy may be internalized in $\mathcal{M}_S$ as a learned policy, externalized in $\mathcal{R}_S$ as a reusable skill or script, operationalized in $\mathcal{H}_S$ as a workflow controller, or realized jointly across multiple components. 
These alternatives produce an updated realization $\mathcal{A}_S^+$ that retains the teacher-derived knowledge beyond the source episode. 

Figure~\ref{fig:fig3_toy_example}(c) illustrates this persistence by assigning $\mathcal{A}_S^+$ a new task $q_B$ in Repository~B, whose files and failing API differ from those in Repository~A. 
Although the original patch cannot be replayed, the student can reuse the knowledge (e.g., debugging strategy, evolving policy, and so on) to reproduce the failure, localize the repair, verify the result, and roll back if a regression occurs. 
This example distinguishes the reuse of transferable knowledge from the reproduction of a task-specific solution. 
Section~\ref{sec:2.3} then expands this trace-based example into a broader definition of Agent Distillation. 

\subsection{Agent Distillation: Definition and Scope} \label{sec:2.3}

Building on the running example, we now formalize Agent Distillation at the level of teacher and student agent realizations. 
In this setting, the \emph{teacher} denotes the realization from which task-solving knowledge is derived, whereas the \emph{student} denotes the realization into which that knowledge is persistently incorporated.  
We then use $\mathcal{A}_T$ to denote the teacher, $\mathcal{A}_S^0$ the student state before distillation and $\mathcal{A}_S^+$ the student states after distillation. 
This view covers transfer between different agents, self-evolution through the reuse of prior experience, and learning from multiple teachers.

\paragraph{\textbf{Teacher Knowledge.}} 
We use $K_T$ to denote the task-solving knowledge derived from evidence produced by the teacher. 
In Figure~\ref{fig:fig3_toy_example}(b), $K_T$ consists of the reusable debugging principles abstracted from the teacher execution trace $\tau_T^A$. 
More generally, the knowledge $K_T$ may originate from individual components of the teacher realization or from their joint execution. 
At the model level, it may be expressed through generated outputs, rationales, distributions, or internal representations. 
In artifacts, it may be encoded in the contents of reusable resources, while in the harness, it may be captured by workflow and control structures. 
Moreover, the joint execution may further reveal procedural knowledge through actions, observations, feedback, and environment changes. 
Regardless of how it is expressed, agent distillation requires the knowledge $K_T$ from the teacher agent to be persistently incorporated into the student agent realization. 

\paragraph{\textbf{Student Realization.}}
The persistent incorporation of $K_T$ transforms the student from its initial realization $\mathcal{A}_S^0$ into an updated realization $\mathcal{A}_S^+$.
Let $\mathcal{Z}_S$ denote the nonempty set of student components modified through distillation:
\begin{equation}
\mathcal{A}_S^0 \
\xrightarrow[\mathcal{Z}_S]{K_T} \
\mathcal{A}_S^+,
\qquad
\varnothing
\neq
\mathcal{Z}_S
\subseteq
\left\{
\mathcal{M}_S,
\mathcal{R}_S,
\mathcal{H}_S
\right\}.
\label{eq:persistent-student-realization}
\end{equation}
Depending on $\mathcal{Z}_S$, $K_T$ may be incorporated into the student's parameters or learned policy through $\mathcal{M}_S$, retained as reusable artifacts in $\mathcal{R}_S$, or embedded into workflow and control structures through $\mathcal{H}_S$. 
These realizations are not mutually exclusive where $K_T$ can span multiple components, yielding a hybrid realization in which the model, artifacts, and harness jointly encode and enact the distilled knowledge. 
Persistence requires these changes to remain available to the student in subsequent episodes rather than being supplied only transiently within the source interaction.

\paragraph{\textbf{Definition (Agent Distillation).}}
Given the teacher $\mathcal{A}_T$ and an initial student $\mathcal{A}_S^0$, we define \emph{Agent Distillation} as the process of persistently incorporating teacher knowledge $K_T$ into the components of student, thereby producing an updated realization $\mathcal{A}_S^+$ that can reuse this knowledge to improve the agentic competence on the subsequent tasks. 
Inspired by the information bottleneck principle \cite{tishby2000information}, we characterize this objective from two complementary perspectives: \emph{sufficiency} and \emph{minimality}. 
Sufficiency requires $K_T$ to retain the teacher-derived experience that contributes to the student's competence, whereas minimality favors excluding information that does not contribute to this outcome. 
Together, these considerations cast Agent Distillation as identifying the least complex teacher-derived knowledge that is sufficient to attain the best achievable student competence:
\begin{equation}
    K_T^\star
    =
    \operatorname*{arg\,max}_{K_T}
    \left[
        \operatorname{Comp}
        \left (
            \mathcal{A}_S^+(K_T)
        \right)
         \ - \
        \mathcal{C}_K
        \left(
            K_T
        \right)
    \right],
    \label{eq:agent-distillation-objective}
\end{equation}
where the target task setting in $\operatorname{Comp}$ is omitted for brevity. 
The competence term captures the sufficiency of $K_T$, while $\mathcal{C}_K(K_T)$ measures its information complexity and encourages minimality. 
Accordingly, $K_T^\star$ represents the minimal sufficient teacher-derived knowledge for improving the competence of the persistently updated student realization.

\paragraph{\textbf{Operational Scope.}} 
A process falls within the scope of Agent Distillation when three conditions are satisfied. 
\textit{First}, an identifiable teacher agent provides the evidence from which task-solving knowledge is derived. 
The teacher may be a different agent, multiple agents, or the same agent in a self-evolution setting. 
\textit{Second}, the process derives transferable task-solving knowledge $K_T$ from this evidence. 
\textit{Third}, the knowledge $K_T$ is persistently incorporated into the student and remains available to influence future episodes after access to the teacher or source interaction has been removed.

Specifically, in-context learning~\cite{brown2020language}, inference-time consultation~\cite{madaan2023selfrefine}, and multi-agent collaboration~\cite{li2023camel,wu2023autogen} do not constitute Agent Distillation when their effects end with the current episode. 
If such interactions produce a persistent change in the student, they may serve as mechanisms for agent distillation. 
Feedback-driven adaptation that affects only the current episode and leaves no persistent change in the agent falls outside Agent Distillation \cite{yao2023react,madaan2023selfrefine}. 
In contrast, self-evolution falls within our scope when prior experience serves as teacher-side evidence from which the same agent derives reusable knowledge and retains it for future episodes \cite{shinn2023reflexion,zhao2024expel,wang2023voyager}. 
Within these boundaries, agent distillation is agnostic to both the form of distilled task-solving knowledge and the realization of the teacher and the student. 


\subsection{Survey Methodology} \label{sec:2.4}

To construct the literature corpus, we adopted a definition-guided collection process rather than relying solely on the term \emph{Agent Distillation}, which has not yet been used consistently across research communities. 
We searched the \textit{ACM Digital Library}, \textit{IEEE Xplore}, \textit{Scopus}, \textit{ACL Anthology} and supplemented these sources with \textit{arXiv} and \textit{OpenReview} to capture recent studies available through \textit{August 31, 2026}. 
Search queries combined terms related to agent learning, knowledge distillation, experience reuse, memory, tools, skills and workflow optimization. 
We further used \textit{Semantic Scholar} and \textit{Google Scholar} for backward and forward citation tracing. 
Guided by the definition in Section~\ref{sec:2.3}, we included as core literature studies in which teacher knowledge is persistently incorporated into $\mathcal{M}_S$, $\mathcal{R}_S$ or $\mathcal{H}_S$. 
Studies on transient prompting and inference-time collaboration were retained as contextual literature but were not treated as instances of agent distillation. 
We then coded each core study according to its teacher--student configuration, knowledge form, updated agent components, distillation mechanism and evaluation setting. 
The recurring patterns identified through this process provide the empirical basis for the taxonomy developed in the following section.
\section{Our Lens: A Knowledge Substrate-Centric Taxonomy}\label{sec:3}

Building on the definition of Agent Distillation in Section~\ref{sec:2}, the central taxonomic question is where and in what persistent form teacher-derived task-solving knowledge is retained for future use. 
We use \textbf{\emph{knowledge substrate}} to denote the persistent carrier within an agent realization through which such knowledge is encoded, retained, and made available to future episodes. 
Each substrate combines a mode of knowledge retention with a functional locus in the agent: learned model computation for parametric knowledge, independently addressable objects for artifact knowledge, and execution or control logic for harness knowledge. 
For organizational purposes, we then classify each method according to the primary substrate in which the student retains the transferred knowledge after distillation. 

Table~\ref{tab:knowledge_substrates} summarizes the representative forms and functional roles of these three substrates, corresponding to three complementary modes of \emph{internalization}, \emph{externalization}, and \emph{operationalization}.
The listed forms are illustrative: substrate membership is determined by how the retained knowledge participates in future task solving, rather than by its physical format or the mechanism used to obtain it.
Accordingly, an adapter stored as a separate file remains parametric, a callable script constitutes an artifact, and the logic controlling its invocation belongs to the harness.
Teacher outputs, rationales, and execution traces provide evidence for distillation, while fine-tuning, memory construction, and workflow induction describe transfer mechanisms.
The same evidence may therefore support different forms of distillation, depending on where and how the resulting knowledge persists in the student.

\begin{figure}[t]
    \centering
    \includegraphics[width=0.95\linewidth]{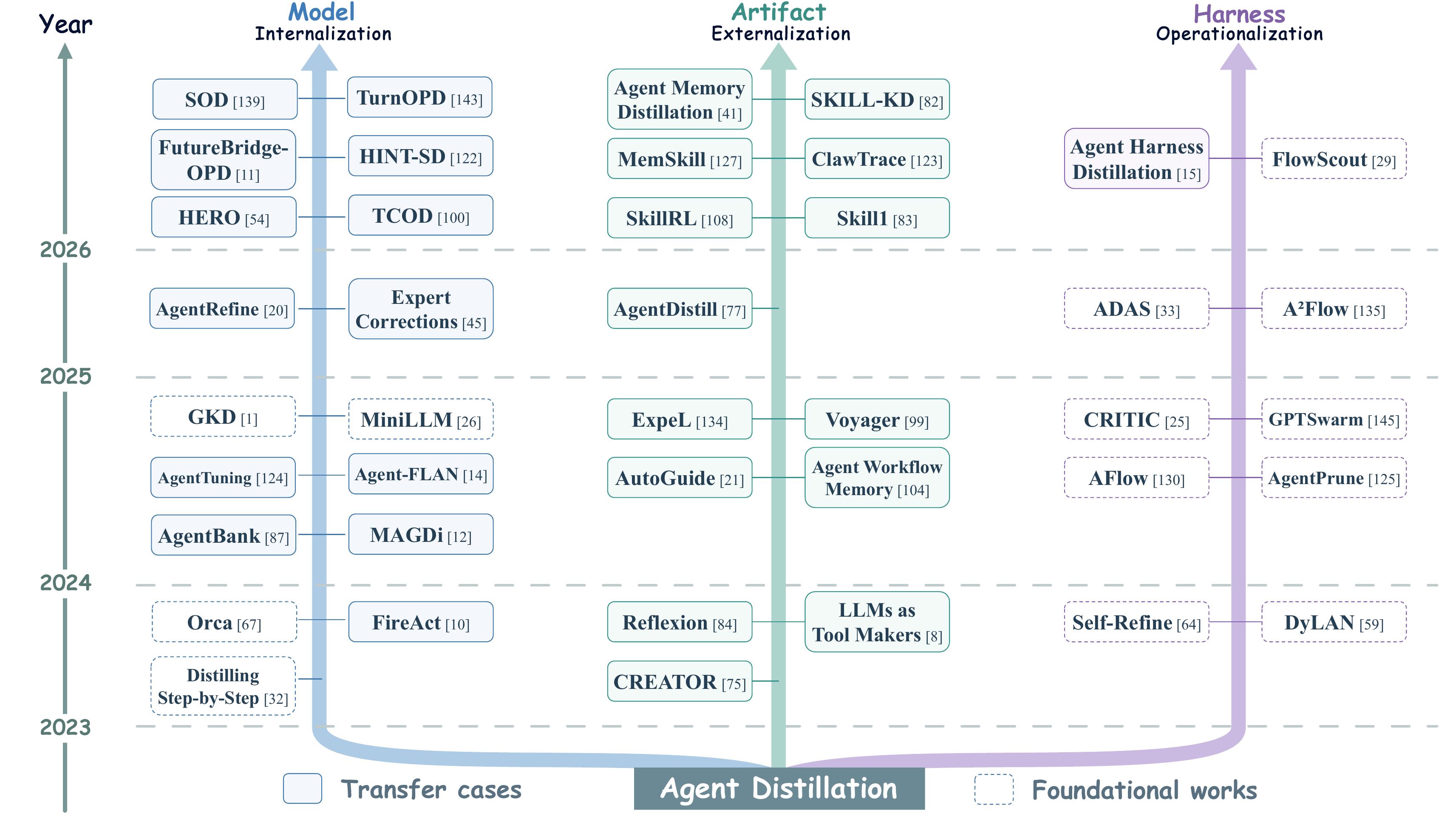}
    \caption{\textbf{Timeline of representative literature on Agent Distillation across model, artifact, and harness knowledge.}}
    \label{fig:related_work}
    \vspace{-0.2cm}
\end{figure}

Figure~\ref{fig:related_work} situates representative literature within this taxonomy and traces its development along the three branches.
In the model branch, the literature expands from response and reasoning distillation to learning from interaction trajectories and supervising student-induced histories.
The artifact branch connects the extraction of reusable memories, procedures, and tools with the curation and evolution of the resulting libraries.
The harness branch places workflow construction and multi-agent orchestration alongside emerging efforts to distill execution and control logic.
Solid boxes identify transfer cases within our operational scope, while dashed boxes mark foundational or adjacent studies that inform the surveyed mechanisms.
This distinction preserves the connection to neighboring research while keeping persistent teacher--student knowledge transfer as the criterion for inclusion as \textbf{Agent Distillation}.

The three substrates define the principal forms of distilled knowledge, but a transfer process need not remain within a single substrate. 
We use \emph{cross-substrate transfer} to describe the transformation of knowledge from one substrate form into another and \emph{hybrid realization} to describe an outcome in which distilled knowledge is jointly retained across multiple substrates. 
The former characterizes the transfer path, whereas the latter characterizes the organization of knowledge in the resulting student. 
We first examine parametric, artifact and harness distillation in Sections~\ref{sec:4}--\ref{sec:6}, and then return to the cross-substrate transfer and the hybrid realization in Section~\ref{sec:7}.

\begin{table*}[t]
    \centering
    \small
    \caption{\textbf{Principal knowledge substrates in Agent Distillation and their roles in future task solving.} }
    \label{tab:knowledge_substrates}
    \renewcommand{\arraystretch}{1.12}
    \begin{tabularx}{\textwidth}{
        @{}
        >{\raggedright\arraybackslash}p{0.09\textwidth}
        >{\raggedright\arraybackslash}p{0.25\textwidth}
        >{\raggedright\arraybackslash}p{0.31\textwidth}
        >{\raggedright\arraybackslash}X
        @{}
    }
        \toprule
        \textbf{Substrate} &
        \textbf{Embodiment} &
        \textbf{Representative forms} &
        \textbf{Role in future task solving} \\
        \midrule

        \textbf{Model} &
        Learned model computation &
        Parameters, adapters, internal representations, learned policies &
        Internalizes knowledge to directly shape inference and decisions \\
        \addlinespace[4pt]

        \textbf{Artifact} &
        Explicit reusable objects &
        Memories, demonstrations, plans, code, skills, and tools &
        Externalizes knowledge for retrieval and reuse across episodes \\
        \addlinespace[4pt]

        \textbf{Harness} &
        Dynamic execution logic &
        Workflows, routing, verification, recovery, and orchestration logic &
        Operationalizes knowledge by organizing and regulating execution \\

        \bottomrule
    \end{tabularx}
\end{table*}


\section{Internalizing Agent Knowledge in Parametric Models}\label{sec:4}

\begin{figure}[th]
    \centering
    \includegraphics[width=0.95\linewidth]{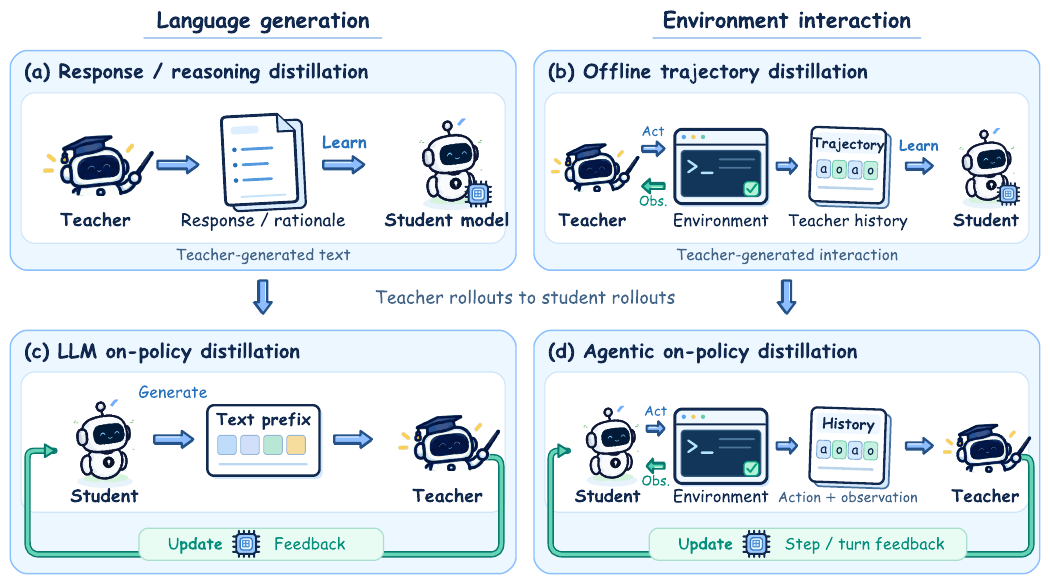}
    \caption{\textbf{The progression of parametric distillation toward environment-grounded agent policy learning.}
In the upper row, the student learns from the teacher's textual traces (a) or recorded interaction trajectories (b). 
The downward arrows mark the shift to on-policy distillation, where the teacher conditions its feedback on the student's own prefix (c) or its history of executions (d). 
    }
    \label{fig:fig4}
    \vspace{-0.3cm}
\end{figure}

In this section, we examine how the student agent internalizes teacher-derived task-solving knowledge in its parametric models.
During task execution, the agent may use LLMs, vision--language models (VLMs)~\citep{driess2023palme,hong2024cogagent}, or vision--language--action policies (VLAs)~\citep{kim2024openvla,zitkovich2023rt2} to reason over observations or propose actions.
Under the decomposition introduced in Section~\ref{sec:2}, distillation updates the student's models from $\mathcal{M}_S^0$ to $\mathcal{M}_S^+$, which the agent uses in subsequent episodes.
The central question within this branch is whose decisions shape the histories on which teacher supervision is based.
The student may learn from teacher-generated histories or receive teacher feedback on histories shaped by its own decisions.
Student-induced histories can expose deviations and recovery needs that successful teacher demonstrations may omit. 
Figure~\ref{fig:fig4} further illustrates this distinction by showing who generates the response in a language task and whose actions shape subsequent observations in an environment. 
We review response and reasoning distillation (\S~\ref{sec:parametric-language}), offline trajectory distillation (\S~\ref{sec:parametric-offline}), and on-policy distillation for LLMs (\S~\ref{sec:parametric-opd}) and interactive agents (\S~\ref{sec:parametric-agentic-opd}).

\subsection{Response and Reasoning Distillation}
\label{sec:parametric-language}

Conventional knowledge distillation establishes the basis of parametric transfer by training a student on information exposed by a teacher. 
Soft output distributions convey predictive structure beyond hard labels~\citep{hinton2015distilling}, while sequence-level distillation transfers teacher-generated sequences to an autoregressive student~\citep{kim2016sequence}. 
When teacher probabilities are unavailable, generated responses provide a practical supervision interface: a closed teacher answers an instruction collection, and an open or smaller student learns from those responses through supervised fine-tuning. 
Instruction distillation applies this approach to task-specialized students~\citep{sun2023instruction}. 
Its accessibility follows from requiring only generated text, although the resulting supervision remains tied to the prompts and teacher continuations represented.

Reasoning distillation enriches response supervision with intermediate steps that expose how a solution is constructed.
Orca uses explanation traces to support learning from complex instructions~\citep{mukherjee2023orca}, while Distilling Step-by-Step treats teacher rationales as an additional prediction target~\citep{hsieh-etal-2023-distilling}.
Related methods transfer chain-of-thought demonstrations, reasoning decompositions, and consistency-aware rationales~\citep{magister2023teaching,shridhar2023distilling,wang2023scott}.
These signals can teach reusable decomposition and inference patterns, but their value depends on the quality of the generated explanations and their compatibility with the student.
A rationale is a problem-specific textual account of reasoning, whose correctness and faithfulness need to be considered alongside its usefulness as a training target.
Analyses of rationale selection and mixing report non-uniform gains~\citep{wadhwa2024mysteries,li2024mixed}, and stronger teacher reasoning can exceed the student's learning capacity~\citep{li2025learnability}.
Curriculum design and prefix alignment further adapt supervision to student competence~\citep{jiang2025curriculum,liu2026prefixalign}.
These language-distillation techniques provide foundations for agent learning; interaction trajectories extend their supervision to how a teacher acquires observations, chooses actions, and responds to environmental consequences.

\subsection{Offline Trajectory Distillation}
\label{sec:parametric-offline}

An interaction trajectory records the reasoning, actions, and observations through which a teacher solves a task.
Using the notation of Section~\ref{sec:2}, the trace $\tau_T$ exposes how decisions depend on earlier observations and how their consequences shape subsequent choices.
Constructing a training corpus therefore involves both collecting teacher executions and selecting the parts that provide useful supervision. 
Outcome filtering identifies successful episodes, while process-level checks can detect invalid calls, redundant steps, or decisions that require repair. 
Failed attempts may also reveal useful diagnosis and recovery behavior. 
Selecting only clean successes can remove these behaviors from the corpus, while retaining a failed decision together with its correction can expose how an agent identifies and repairs a mistake. 
Exploration-based trajectory optimization compares alternative executions~\citep{song-etal-2024-trial}, step-level refinement repairs problematic decisions~\citep{xiong-etal-2024-watch}, and AgentRefine constructs refinement data to improve generalization beyond narrow imitation of successful demonstrations~\citep{fu2025agentrefine}.
Together, these approaches make trajectory quality a question of which behavioral regularities the student can use on later tasks.

Given a curated teacher corpus $\mathcal{D}_T=\{\tau_T^{(i)}\}$, the student learns to select decisions conditioned on the histories in that corpus.
Let $\pi_\theta(a\mid h)$ denote the conditional decision distribution contributed by the student model within its surrounding agent realization.
A representative offline objective is
\begin{equation}
    \mathcal{L}_{\mathrm{traj}}(\theta)
    = -\mathbb{E}_{\tau_T\sim\mathcal{D}_T}
    \left[\sum_{t=0}^{L-1} w_t\log\pi_\theta(a_t^T\mid h_t^T)\right],
    \label{eq:offline-trajectory}
\end{equation}
where $h_t^T$ and $a_t^T$ are the teacher history and decision, $L$ is the trajectory length, and $w_t$ optionally weights decisions by validity, difficulty, or importance.
Reasoning and actions can be serialized for training, with observations supplied as conditioning context and loss masks selecting the prediction targets.
FireAct instantiates this approach using ReAct-style trajectories collected across tasks and prompting methods~\citep{chen2023fireact}.
Combining trajectories from various prompting strategies exposes alternative decompositions and action patterns, although the resulting policy is still learned within the interfaces and observations represented in the corpus. 
Such supervision couples reasoning with tool use and observation-conditioned continuation, enabling the model to learn decision patterns that were demonstrated by teacher.

Broader trajectory collections seek to make these patterns useful across tasks and environments.
AgentTuning combines interaction trajectories with general instruction data~\citep{zeng-etal-2024-agenttuning}; AgentBank expands the scale and heterogeneity of the interaction corpus~\citep{song-etal-2024-agentbank}; and Agent-FLAN studies the composition and training choices needed to use such data effectively~\citep{chen-etal-2024-agentflan}.
Their shared concern is whether the student learns planning, state tracking, and revision patterns that extend beyond a particular task format.
Tool-oriented studies complement this emphasis by expanding the decisions represented in the corpus.
ToolLLM constructs instruction and solution trajectories for selecting and invoking APIs~\citep{qin2024toolllm}, while distillation with retrieval and code tools teaches smaller models when to consult external knowledge or execute computation~\citep{kang2025distillingllmagent}.
This is particularly relevant when a capacity-limited student must decide which parts of a task to solve through learned reasoning and which to delegate to available resources.
Across these settings, coverage concerns both the situations represented in the data and the actions available to the learned policy.

Offline trajectory distillation separates teacher execution from student training, allowing demonstrations to be filtered, rebalanced, and reused.
Its coverage remains constrained by the histories generated during teacher execution.
Critical-step selection~\citep{chen-etal-2025-atlas}, structured decomposition~\citep{liu2026structuredagentdistillation}, and capacity-aligned trajectory construction~\citep{tang-zhao-2026-smartad} improve the relevance and learnability of the supervision, but deployment exposes the student to histories induced by its own decisions.
An early deviation can lead to an unfamiliar observation and further errors, even when the teacher corpus contains many successful demonstrations.
The resulting gap shifts attention from collecting additional teacher trajectories to obtaining teacher guidance on the histories the student actually visits.

\subsection{LLM On-Policy Distillation}
\label{sec:parametric-opd}

On-policy distillation (OPD) obtains teacher supervision on histories generated by the current student, following the learner-state supervision principle of interactive imitation learning~\citep{ross2011dagger}.
For an autoregressive LLM, these histories consist of the task input and the student's generated prefixes.
The student produces a rollout, the teacher supplies feedback on the encountered prefixes, and parameter updates incorporate that feedback.
On-policy thus specifies the source of the training histories; updates need not occur after each generated token.
When teacher token probabilities are available, Generalized Knowledge Distillation provides a representative formulation~\citep{agarwal2024gkd}:
\begin{equation}
    \mathcal{L}_{\mathrm{OPD}}(\theta)
    = \mathbb{E}_{q\sim\mathcal{P},\,h_t\sim d_{\pi_\theta}(\cdot\mid q)}
    \left[D_f\!\left(\pi_\theta(\cdot\mid h_t),\pi_T(\cdot\mid h_t)\right)\right],
    \label{eq:opd-objective}
\end{equation}
where $d_{\pi_\theta}(\cdot\mid q)$ is the student-induced prefix distribution for task $q$, $\pi_T$ supplies the teacher distribution, and $D_f$ measures their discrepancy. 
The supervision is therefore evaluated under the conditions created by the student itself. 
The teacher then supply targets for these prefixes, allowing training to address deviations as well as successful continuations.

Within this setting, the choice of matching objective determines which aspects of teacher behavior the student is encouraged to retain.
Forward KL from teacher to student penalizes missing teacher-supported modes, whereas reverse KL penalizes assigning student probability to teacher-disfavored outputs.
MiniLLM uses reverse KL to support selective approximation of a larger teacher~\citep{gu2024minillm}.
DistiLLM combines skew-divergence objectives with adaptive off-policy sampling to balance transfer quality and the cost of generating student outputs~\citep{ko2024distillm}.
These approaches connect distribution matching with student capacity and sampling efficiency.
Covering multiple teacher-supported alternatives and concentrating on continuations already plausible for the student impose different learning demands; the preferred objective depends on the behavior being transferred and the student's ability to represent it.
Distillation thus involves choosing which histories to supervise and which teacher behaviors to emphasize within them. 

Teacher access determines how this supervision can be delivered.
Direct token-distribution matching requires exposed probabilities and an alignment between teacher and student prediction spaces; cross-tokenizer methods address cases in which their text segmentations differ~\citep{boizard2025uld}.
With a black-box teacher, generated corrections, critiques, and preferences can instead provide supervision on student continuations.
On-policy expert correction repairs decisions in student rollouts~\citep{lauffer2025oec}, while student-centered methods concentrate guidance around the student's current capability boundary~\citep{lyu2026score}.
Corrections directly specify behavior to imitate, whereas preferences and critiques require a corresponding optimization or target-construction step before they can guide parameter updates.
Equation~\ref{eq:opd-objective} describes the distribution-matching instance of OPD, rather than a requirement that all such methods obtain teacher logits.
Likewise, offline learning can use soft teacher targets: the distinction between offline and on-policy distillation concerns the histories supplying training conditions, while the feedback format can vary in either setting.

Maintaining supervision on the current student distribution introduces a collection cost.
Fresh rollouts expose current errors but require repeated generation and teacher queries. 
Cached teacher computations reduce redundant queries, while mixture sampling and replay incorporate histories collected under teacher or earlier student policies. 
Reused histories need not follow the current student distribution; their value depends on whether they still expose useful decisions and failures for the current student. 
The practical objective is to retain useful coverage of student behavior within the available supervision budget. 
This tradeoff becomes more consequential when a rollout involves external operations, because collecting a history now requires executing actions and observing their effects. 
Agentic OPD must assess whether teacher feedback is valid in the resulting states and relevant to subsequent task progress. 

\subsection{Agentic On-Policy Distillation}
\label{sec:parametric-agentic-opd}

For an interactive agent, the history $h_t$ defined in Section~\ref{sec:2} contains observations returned by an environment in response to earlier actions.
The student therefore influences both its generated decisions and the external conditions on which later decisions depend.
A tool call must satisfy the interface, apply to the current state, and support further progress; the same textual command can have different consequences in different states.
Revisiting DAgger for LLM agents connects this setting to interactive imitation learning~\citep{li2026revisitingdagger}.
Teacher guidance must consequently address decisions within the student's actual interaction history, including states that arise from unsuccessful actions.

Student rollouts expose recovery decisions that successful demonstrations may omit.
After a successful search, a teacher might summarize the retrieved document. 
If the student omits a required argument and receives an error instead, the teacher can use that observation to supervise a corrected call. 
As tool errors accumulate, however, the teacher's token-level feedback on later steps can become less reliable. 
SOD therefore reduces the weight of that feedback when student--teacher divergence grows from one step to the next~\citep{zhong2026sod}.  

The consequences of an action may become observable only after the environment responds or several further decisions are made.
This motivates supervision at different temporal scales: token-level targets address local output choices, step-level targets cover a reasoning or tool decision, turn-level guidance relates an action to its returned observation, and trajectory-level feedback assesses progress toward completion.
TurnOPD organizes supervision around complete interaction turns~\citep{zhou2026turnopd}, while temporal curricula progressively expose the student to longer-horizon dependencies~\citep{wang2026tcod}.
These choices determine which dependencies the learning signal can capture.
For example, a correctly formatted tool call may return useful information that the student subsequently ignores; evaluating the interaction turn exposes a failure that action-syntax agreement cannot detect.
Local targets remain useful for executable decisions, while longer units allow supervision to reflect consequences not directly evaluated at the token level. 

The value of teacher guidance also depends on when intervention occurs and whether the student can use it to continue successfully.
Large disagreement can indicate a useful correction opportunity, but it can also arise after the history has drifted beyond what a local correction can repair.
FutureBridge-OPD executes a short teacher bridge and validates its effect on the subsequent student continuation~\citep{chen2026futurebridge}.
This evaluates guidance through the future behavior it enables, making intervention selection a question of downstream utility.
Multi-turn prefix replay offers a complementary mechanism for reusing informative earlier histories~\citep{liao2026reopd}.
A teacher bridge supplies a continuation from the current state, whereas replay revisits an earlier prefix as a training condition.
Together, intervention selection and replay address how to allocate teacher effort and which portions of student experience provide useful conditions for further learning.

Further supervision becomes available after an interaction reveals its consequences.
Self-distillation can use this additional information to instantiate a teacher with richer context than the student receives at deployment.
The two roles may share model parameters while receiving different information, with the student learning to retain useful behavior after the privileged context is removed.
Self-Distilled Reasoner illustrates the general pattern using verified reasoning traces~\citep{zhao2026opsd}; in interactive settings, HINT-SD revisits critical decisions through targeted hindsight~\citep{yeo2026hintsd}, and HERO derives reflective supervision from later environment observations~\citep{liu2026hero}.
Related approaches combine agentic self-distillation with reinforcement learning through prior high-quality rollouts, observation-calibrated targets, or progressively improved policies~\citep{yang2026ocsd,lu2026sdar,zhang2026stepopsd,wang2026agentopsd}.
Teacher-derived targets guide behavior at encountered histories, while environment rewards optimize task-level utility; their relative influence can also change during training~\citep{tan2026atod}.
Under the scope established in Section~\ref{sec:2}, the distillation requires identifiable teacher-derived supervision that persists through student updates. 
Environment rewards alone do not identify what the student learned from the teacher.

Across these approaches, parametric distillation aligns teacher guidance with histories on which the student must learn to act, from demonstrated reasoning to the consequences of its own environment interaction.
The resulting knowledge persists in $\mathcal{M}_S^+$.
Explicit experiences, procedures, and executable resources offer a complementary retention form when inspection, revision, and reuse are important, motivating the artifact distillation methods examined next.

\vspace{-0.3cm}
\section{Externalizing Agent Knowledge as Reusable Artifacts}\label{sec:5}
\begin{figure}[tbp]
    \centering
    \includegraphics[width=0.95\linewidth]{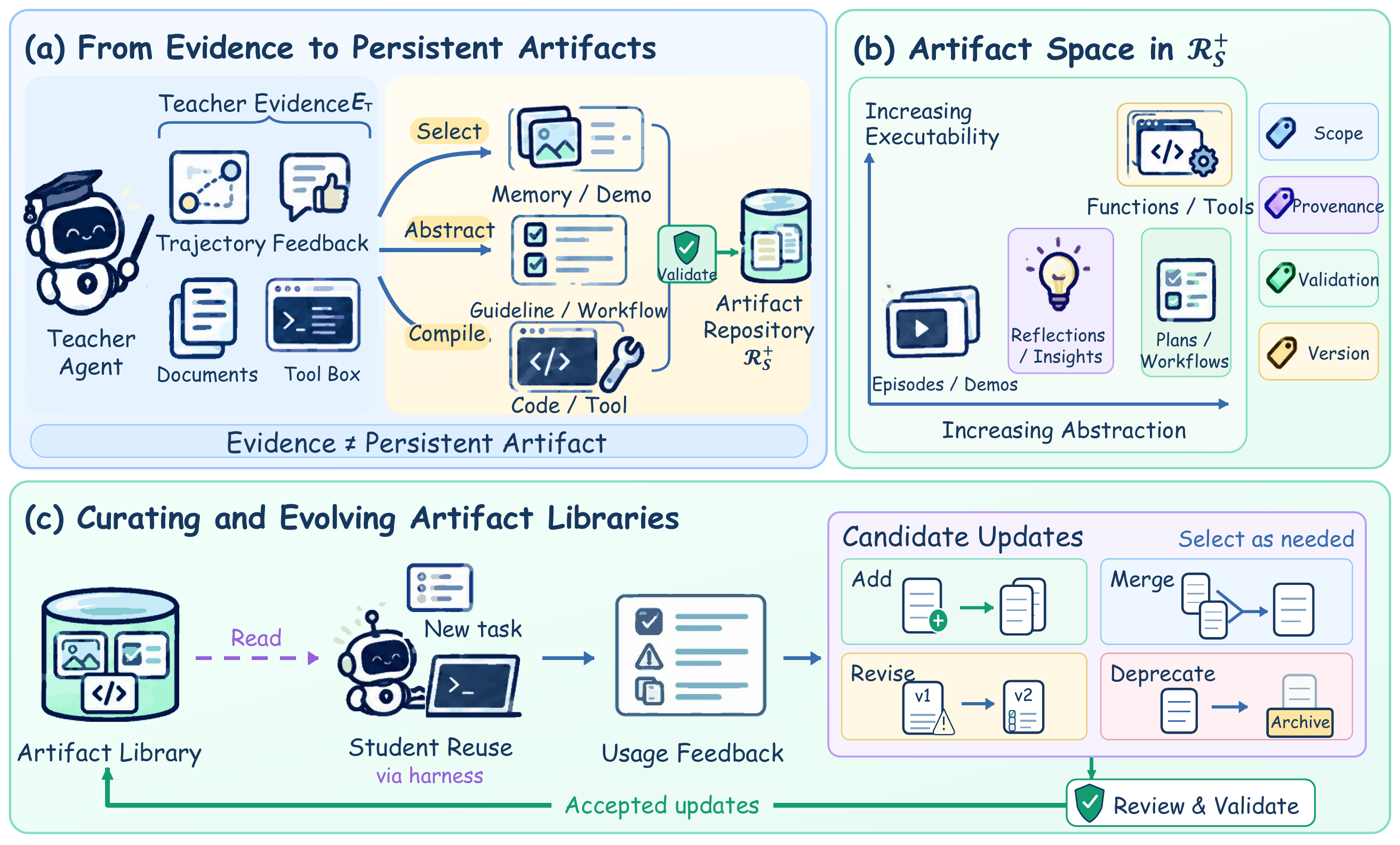}
    \caption{\textbf{Representative artifact forms in Agent Distillation.}
    (a) Teacher evidence yields artifacts retained beyond the original task. 
    (b) These range from demonstrations of individual episodes to reusable procedures the student can follow as checklists or invoke as tools. 
    (c) Feedback from reuse guides proposed library updates, which are reviewed and validated before adoption.}
    \label{fig:placeholder}
    \vspace{-0.5cm}
\end{figure}

Artifact distillation externalizes teacher-derived knowledge as independently addressable resources for reuse across episodes. 
Under the decomposition introduced in Section~\ref{sec:2}, distillation updates $\mathcal{R}_S^0$ to $\mathcal{R}_S^+$. 
Let $\mathcal{E}_T$ denote teacher evidence, including outputs, critiques, documents, and interaction traces with environment feedback.
A construction operator $G$ converts the evidence into candidate artifacts $\mathcal{C}_T$ and a repository update $U_R$ incorporates validated candidates:
\begin{equation}
    \mathcal{C}_T
    =G\!\left(\operatorname{Select}(\mathcal{E}_T)\right),
    \qquad
    \mathcal{R}_S^+
    =U_R\!\left(\mathcal{R}_S^0,\operatorname{Validate}(\mathcal{C}_T)\right).
    \label{eq:artifact-update}
\end{equation}
The update can add a new resource or revise an existing entry as further evidence becomes available.
The retained objects differ in \emph{abstraction}, from individual cases to procedures shared across tasks, and in \emph{executability}, from descriptive guidance to callable resources (Figure~\ref{fig:placeholder}(b)).
We examine distilled memories (\S~\ref{sec:artifact-memory}), reusable procedures and skills (\S~\ref{sec:artifact-procedures}), and executable tools and programs (\S~\ref{sec:artifact-executable}), followed by their organization and evolution within artifact libraries (\S~\ref{sec:artifact-lifecycle}).

\vspace{-0.3cm}
\subsection{From Episodes to Distilled Memories}
\label{sec:artifact-memory}

An execution trace mixes reusable decisions with task-specific identifiers, transient paths, and verbose tool outputs.
Memory distillation selects which parts should remain available as summaries, critical state--action segments, failure diagnoses, or structured records.
In the repository example, the regression and rollback provide reusable evidence, while a temporary file path merely identifies the source run.
Successful episodes reveal viable strategies, while failures expose invalid assumptions and recovery opportunities.
Reflexion converts task feedback into verbal reflections stored in episodic memory~\citep{shinn2023reflexion}.
A retained teacher critique can similarly explain a student's failure.
Recording the failed call, error, correction, and outcome preserves the condition that made a retry useful.

Selected experience can be retained as a case linking a state, decision, and outcome, or as a demonstration presenting a sequence to imitate.
Examples preserve argument names and calling patterns that a summary may omit, allowing the student to reuse interface conventions without changing its parameters.
Agent Memory Distillation organizes successful teacher trajectories into hierarchical memories for a smaller, frozen student, separating task-level workflows, subtask examples, and function-level calling conventions~\citep{kim2026agentmemory}.
The memory level can match the decision: a workflow supports planning, while a function-level example can help resolve tool-call errors without replaying the full trajectory.
Retaining the source context helps the student judge whether a convention applies to its current interface. 

Comparing cases can reveal regularities that extend beyond an individual episode.
ExpeL extracts natural-language insights from training-task experience and recalls them alongside relevant past trajectories~\citep{zhao2024expel}.
The resulting store combines episodic evidence with semantic memory: cases record particular outcomes, while insights express regularities inferred across them.
Abstraction removes incidental names and actions but must preserve the assumptions under which the episodes support an insight.
Keeping the supporting cases makes those assumptions inspectable when a new task differs from the source setting.
Generative Agents, MemoryBank, and MemGPT provide related mechanisms for reflection, forgetting, retrieval, and tiered context management~\citep{park2023generative,zhong2023memorybank,packer2023memgpt}.
When several episodes share a sequence of decisions, an explicit procedure can spare the student from reconstructing that sequence from separate cases.

\subsection{Abstracting Reusable Procedures and Skills}
\label{sec:artifact-procedures}

Procedural artifacts encode reusable \emph{how-to} knowledge as instructions, checklists, plan skeletons, workflow templates, or textual skills.
A procedure can specify its steps, preconditions, completion criteria, and recovery rules.
Applicability matters because an action useful in one state may be destructive in another.
AutoGuide extracts guidelines from offline experience and pairs each with its conditions of use~\citep{fu2024autoguide}.
This allows the student to retrieve a decision rule when its condition recurs, even if the preceding actions differ from the source trajectory.
The retained condition should explain why the action was appropriate, such as a particular tool error, while omitting incidental request identifiers. 

Longer procedures emerge by identifying recurring subgoals, abstracting task-specific entities, and preserving action--observation dependencies.
Agent Workflow Memory induces workflows from offline demonstrations or online experience and supplies relevant workflows on later tasks~\citep{wang2024awm}.
A workflow might instruct the student to locate an item, check its attributes against a constraint, and confirm the resulting state.
The item can vary while the need to check the constraint before confirmation persists.
This abstraction preserves consequential ordering while leaving instance-specific choices to the student.
Stored instructions and prompt templates can provide the same reusable guidance; the runtime that binds variables, advances steps, and enforces branches is examined in Section~\ref{sec:6}.

Skill granularity affects reuse and selection cost.
Fine-grained rules are easy to recombine but require accurate selection among many candidates, while a larger skill may bind together steps that should vary across tasks.
SKILL-KD uses a student's failure to identify the guidance it needs.
It compares a failed student trajectory with a teacher trajectory on the same task, distills the actionable difference into a textual skill patch, and reruns the student to refine ineffective patches~\citep{shi2026skillkd}.
Re-execution tests whether the patch repairs the motivating failure.
A nearby task can reveal whether it also adds unnecessary calls or suppresses a valid alternative, motivating a narrower scope or an explicit exception. 

A teacher's successful procedure may include steps required by its own organization.
Two teacher agents may need a handoff even when a single student can use the same domain tools without that exchange.
AdaSkill studies selective transfer from multi-agent systems into single-agent skills, preserving capability resources while limiting pipeline guidance that can overconstrain the student when several solutions satisfy the task~\citep{xu2026adaskill}.
The retained skill should therefore preserve the conditions and operations needed for reuse while allowing the student to organize its own execution.
Textual procedures still require interpretation at each use.
When a recurring operation admits a stable interface and a mechanical implementation, it can instead be compiled into an executable artifact.

\vspace{-0.3cm}
\subsection{Synthesizing Executable Tools and Programs}
\label{sec:artifact-executable}

Executable artifacts retain knowledge as functions, scripts, API wrappers, or macros whose behavior is partly determined outside the language model.
They amortize teacher reasoning across episodes: the teacher implements an operation once, and the student invokes it through a compact interface.
\emph{Large Language Models as Tool Makers} illustrates this division with a capable model constructing reusable tools and a lighter model applying the cached tools to subsequent requests~\citep{cai2023latm}.
CREATOR separates tool creation from decision execution by generating tools from task descriptions and running them on task instances~\citep{qian2023creator}.
Reuse requires an interface that exposes varying arguments while preserving the shared computation.
Hard-coding one answer limits reuse, while an underspecified interface can leave the student to reconstruct the original reasoning before invoking the tool.

Interactive environments provide feedback about what an executable skill actually does.
Voyager builds a code-skill library through environment feedback, execution errors, and self-verification for reuse in new Minecraft worlds~\citep{wang2023voyager}.
Lifelong Robot Library Learning distills recent experience into composable robot skills and uses memory and exploration to expand the tasks the library can address~\citep{tziafas2024lrll}.
The state change produced by execution reveals whether a skill realizes its intended operation and where its implementation needs repair.
For self-evolving agents, an earlier successful realization can supply teacher evidence for a later one.
Execution feedback can refine both the implementation and its applicability conditions, which remain important when the student invokes the skill in a different state.

A skill package can combine descriptive and executable resources.
AgentDistill packages teacher-generated problem-solving knowledge into Model Context Protocol (MCP) modules that bundle structured instructions and callable resources for smaller students to reuse without parameter training~\citep{qiu2025agentdistill}.
The instructions explain how to use the resource, while schemas and programs expose its callable functionality.
Validation must cover both the implementation and its use through the student's interface.
Static checks examine syntax, types, dependencies, and declared permissions; sandboxed execution and unit tests exercise functional behavior and failure handling.
A function that passes its tests may still fail when the student supplies incompatible arguments.
Linking these checks to source episodes records which behavior the artifact should preserve.
As APIs and dependencies change, previously valid tools may require revision, making continued reuse a library-level concern.

\vspace{-0.3cm}
\subsection{Curating and Evolving Artifact Libraries}
\label{sec:artifact-lifecycle}

An artifact's value depends on the other entries available to the student.
Two useful resources can duplicate one another, prescribe conflicting actions, or compete for a limited context budget.
With the model and harness fixed, curation balances task utility against the collection's storage, retrieval, context, and execution costs.
To support these choices, each entry can record preconditions, input and output schemas, dependencies, source episodes, validation results, and known failures.
These records support both selection and revision.
Two entries may be close in embedding space yet require different API versions; their recorded dependencies distinguish when each is usable.
When a tool convention changes, version records can identify affected entries without reopening every teacher trace.

As evidence accumulates, consolidation merges repeated lessons while preserving meaningful exceptions.
Several local patches may support a general rule, whereas conflicting outcomes can justify splitting a rule by its conditions or assigning precedence in a particular state.
ClawTrace summarizes model calls, tool use, sub-agent operations, and redundant steps in trace cards that guide decisions to preserve, repair, or prune a skill~\citep{yuan2026clawtrace}.
This connects a proposed edit to the execution behavior and cost that motivated it.
Removing an unnecessary operation may improve reuse, while deleting a rare recovery branch can erase an important exception.
An edit record should therefore retain the supporting tasks, student configurations, and observed outcomes.
Comparing the revised entry with its prior version provides evidence for accepting the update, and preserving that version allows rollback if later use reveals a regression.

Artifact evolution connects these updates to feedback from subsequent executions.
SkillRL distills experience into a hierarchical skill bank that co-evolves with policy learning~\citep{xia2026skillrl}.
Skill1 uses a common task-outcome signal to optimize skill selection, utilization, and new-skill distillation~\citep{shi2026skillone}.
MemSkill learns reusable routines for extracting, consolidating, and pruning information from interaction histories~\citep{zhang2026memskill}.
Selection shapes which resources receive execution feedback, and the resulting revisions change what is available on later tasks.
The evolving library must preserve useful exceptions while removing obsolete or contradictory guidance.
A resource's availability alone, however, does not determine whether the student will select it under the right conditions or respond appropriately when its use fails.
Once artifacts are retained in $\mathcal{R}_S^+$, these decisions depend on the execution harness.
Section~\ref{sec:6} examines how teacher knowledge can persist in the rules governing resource invocation, verification, and recovery.

\vspace{-0.3cm}
\section{Operationalizing Agent Knowledge through Execution Harnesses}\label{sec:6}
Harness distillation transfers a teacher's execution decisions into rules that govern future student episodes. A trajectory records the branch a teacher took, the resource it called, or the failure it recovered from. What persists in $H_S$ is the rule that tells the student when to make that decision again. This separates the harness from the artifacts in Section~\ref{sec:5}. A stored checklist belongs to $R_S$. The runtime in $H_S$ binds its arguments, enforces its order, and acts on failed checks.

\begin{figure}[tbp]
    \centering
    \includegraphics[width=0.95\linewidth]{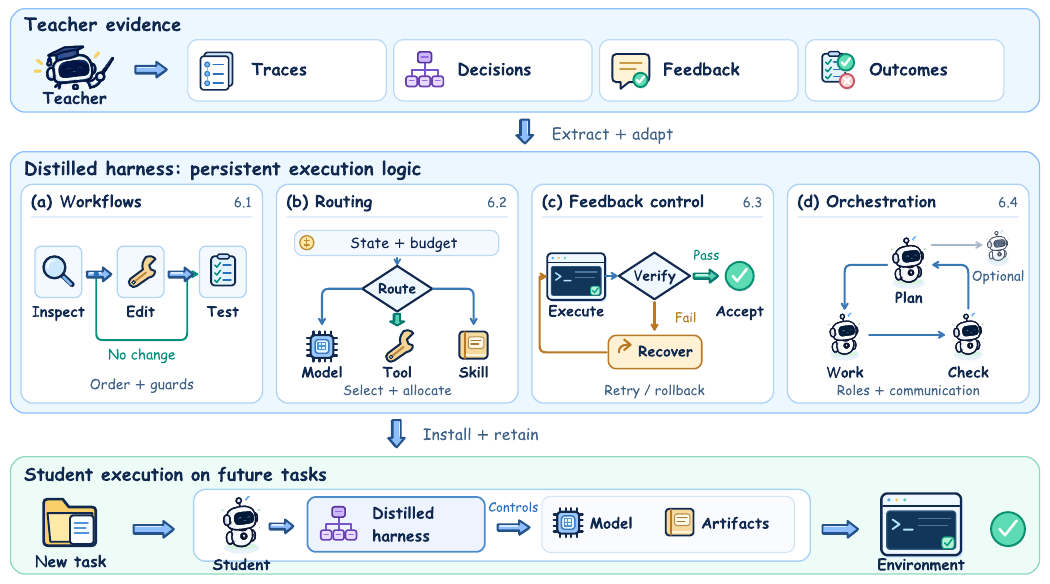}
\caption{\textbf{Harness distillation from teacher evidence to persistent execution logic.}
Inspection can lead directly to testing when no change is needed (a).
Routing chooses a resource under the remaining budget (b), feedback determines acceptance or recovery (c), and orchestration assigns work and checking to different agents (d).
Installed rules allow execution to vary with observations or budgets.}
    \label{fig:harness-distillation-overview}
\end{figure}

Let $\mathcal{E}_T^H=\{\tau_T,d_T,f_T,o_T\}$ denote teacher trajectories, exposed control decisions, feedback, and observed outcomes. To isolate the harness update, fix the model and artifacts in $A_S(H)=(M_S,R_S,H)$. Harness $H$ yields utility $U_H(q)=u(q,\tau(A_S(H),q,\mathcal{E};B))$. With construction operator $\mathcal{D}_H$, the update and deployment objective are
\begin{equation}
    \begin{aligned}
        \mathcal{H}_S^+ = \mathcal{D}_H\!\left(H_S^0;\mathcal{E}_T^H\right), \quad
        \mathcal{H}_S^+ = \arg\max_{H\in\mathcal{H}_S}
            \left\{\mathbb{E}_{q\sim\mathcal{P}}\!\left[U_H(q)\right]
            -\lambda_C C_H-\lambda_L L_H-\lambda_R\Gamma_H\right\},
    \end{aligned}
    \label{eq:harness-distillation}
\end{equation}
where $C_H$, $L_H$, and $\Gamma_H$ denote cost, latency, and operational risk. Figure~\ref{fig:harness-distillation-overview} illustrates operation ordering (\S~\ref{sec:harness-workflows}), resource selection (\S~\ref{sec:harness-routing}), feedback control (\S~\ref{sec:harness-feedback}), and agent coordination (\S~\ref{sec:harness-orchestration}). Workflow search and agent design supply candidate mechanisms, constituting distillation when identifiable teacher experience shapes the deployed controller.

\subsection{From Demonstrations to Executable Control Flows}
\label{sec:harness-workflows}

A teacher demonstration supplies one sequence of actions. To govern new executions, a workflow must preserve which operations depend on earlier results and which can be skipped, repeated, or replaced. Agent Workflow Memory illustrates this distinction by inducing routines from experience and retrieving them as guidance~\citep{wang2024awm}. While the model interprets the retrieved routine, the knowledge resides in $R_S$. Compiling its preconditions and transitions into a controller that dispatches the operations produces a change in $H_S$.

Inducing such a controller requires separating the teacher's decisions from task-specific details. 
Common paths across successful traces suggest shared subgoals, alternative paths reveal branches, and failures can expose a missing guard. In Figure~\ref{fig:harness-distillation-overview}(a), for example, inspection can lead directly to testing when no edit is needed. 
Replaying the most frequent sequence would lose that condition. Recurring order alone is weak evidence of a dependency. 
Two tools may always be called in the same order because the second consumes the first tool's output or because the teacher happened to schedule them that way. Input and output bindings help determine which edges the workflow graph needs. 
A graph can expose parallel work, while a state machine or behavior tree can express conditional retries and termination.

Existing constructors recover different parts of this structure. FlowScout mines coordination patterns from historical tool executions and refines their topology using execution feedback~\citep{hao2026flowscout}. AFlow searches over code-defined workflow graphs, using Monte Carlo tree search to revise their nodes and edges~\citep{zhang2024aflow}. A$^{2}$Flow extracts operations from demonstrations and clusters them into reusable operators for search~\citep{zhao2025a2flow}.

Deployment determines what the student has acquired. WorkflowLLM trains a model to generate orchestrations from workflow data, so training changes $M_S$~\citep{fan2024workflowllm}. Installing a generated workflow for repeated execution changes $H_S$. Such a workflow can spare a smaller student from planning each step anew, but a teacher operation may need to be split or mapped to a different tool. Exercising alternative branches on held-out tasks tests whether the student retained the dependency or merely reproduced a familiar sequence. A workflow induced from successes alone may cover the routine path well while leaving the first failed prerequisite without a valid next step.

\subsection{Distilling Conditional Routing and Resource Allocation}
\label{sec:harness-routing}

A workflow defines the available paths. Routing selects a path or resource once the student knows the current state and remaining budget. At step $t$, the harness observes history $h_t$, available components $M_S$ and $R_S$, environment state $\mathcal{E}_t$, and budget $B_t$. Its policy selects the next control action from $\mathcal{A}_H$ according to
\begin{equation}
    a_t^H \sim \pi_H\!\left(\cdot\mid h_t,M_S,R_S,\mathcal{E}_t,B_t\right),
    \qquad a_t^H\in\mathcal{A}_H .
    \label{eq:harness-routing}
\end{equation}
The student must learn when to select a resource, schedule work, escalate, or stop. A trace records which expensive model was used. Contrasting it with a similar case handled by a cheaper model can reveal what justified escalation. 

Model routing distinguishes predicting a choice from enforcing it. RouteLLM learns from preference data when to use a stronger or weaker model~\citep{ong2024routellm}. If the routing knowledge resides in learned weights, the update is to $M_S$. A runtime threshold or cascade governs which model runs, including what happens when the preferred model exceeds the budget. If both predictor and execution rule are distilled, each contributes a persistent change.

The same principle applies to artifacts, context, and scheduling. Teacher calls and their outcomes can show when a skill applies, which arguments it needs, or when a tool should be avoided. A missing skill calls for a change to $R_S$, whereas repeated selection of an available skill under invalid conditions points to the invocation rule in $H_S$. The harness can also decide which observations to include in a model call or when independent subtasks can run in parallel. A prompt template can remain fixed in the artifact library while a teacher-derived context policy changes what the model sees. That policy can use the outcomes of earlier calls to decide which observations to keep when the student's context window is smaller than the teacher's. At the task level, the scheduler may serialize dependent operations or prune a branch after a new observation makes it unnecessary. A final prompt or plan records one allocation. The policy retains what would lead the student to retrieve different evidence, postpone a subtask, or stop spending its budget.

A transferable routing rule must admit student-specific choices. The student may preserve the teacher's practice of escalating when evidence conflicts while using a different model or threshold under its own budget. Traces with explicit routing decisions support direct imitation. Under black-box access, changes in budget or resource availability can probe the decision boundary, although several policies may still agree on the tested cases. A comparison with matched resources tests whether the student reproduces a teacher choice. Withholding the preferred tool instead tests whether it can apply the same allocation principle under a tighter constraint. Once the chosen resource returns a result, the harness must determine whether to continue or intervene. 

\subsection{Operationalizing Verification, Feedback, and Recovery}
\label{sec:harness-feedback}

A feedback rule connects a call's outcome to the next action. 
Given an observation $o_t$, the harness forms an assessment $v_t=V_H(h_t,o_t)$ and decides whether to accept, revise, retry, escalate, or stop. 
Successful demonstrations show when execution can continue. 
Failed runs, teacher critiques, and paired cases contrasting teacher acceptance and revision expose the intervention boundary. 
The student can learn that boundary even if unable to reproduce long critiques.

The verifier and the rule that acts on its findings play different roles. A test suite or judge prompt is a resource in $\mathcal{R}_S$. The harness determines when to invoke it, which evidence it receives, and whether its verdict blocks the next action. A unit test can identify a specific program defect yet pass a solution that misses a requirement. A model critic can consider that requirement but may share the generator's blind spots. Learning a gate from the teacher therefore includes learning when its verdict is sufficient to proceed. Self-Refine uses a model's own feedback to revise an output~\citep{madaan2023selfrefine}, while CRITIC grounds critiques in external tools~\citep{gou2024critic}. These loops illustrate possible controls. A student distills such a loop when teacher evidence determines a persistent choice, such as which verifier to invoke or when further revision is unlikely to help. Comparing it with a generic loop tests the value of that learned rule.

Recovery depends on what the attempted action has changed. A read timeout may justify a retry; a write timeout may leave the write partly or fully committed, so retrying could duplicate its effects. A teacher's recovery trace can show which postcondition to check and when to retry, compensate, roll back, or seek approval. Recording that condition preserves the branch rationale when the same error message recurs in another state. Idempotence governs safe repetition, while reversibility governs undoing earlier effects. Tracking these conditions lets the controller bound retries for transient failures and escalate when the state cannot be verified.

Verification consumes budget and can suppress useful actions. A check is worthwhile when its expected reduction in failure loss justifies its cost under the student's error rate and deployment constraints. Injecting a partial write tests whether the learned rule checks the state before retrying, while clean completion should allow progress. This contrast links teacher intervention to controller behavior more directly than task success alone.

\vspace{-0.3cm}
\subsection{Multi-Agent Orchestration and Adaptive Harnesses}
\label{sec:harness-orchestration}

In a multi-agent harness, control actions can change who acts next or receives an observation. Write the collaboration graph as $G_H=(V,E,\mathcal{R},\mathcal{S},\mathcal{A})$, with agents or modules $V$, communication edges $E$, role assignments $\mathcal{R}$, an activation schedule $\mathcal{S}$, and an aggregation rule $\mathcal{A}$. The student may retain a fixed graph, compress a teacher graph, or learn a policy to construct one per task. Messages and activation times document one run; the retained controller governs how subsequent teams exchange information and resolve disagreements.

DyLAN estimates agent contributions to select a team for a query~\citep{liu2023dylan}. AgentPrune removes redundant or harmful messages from a communication graph~\citep{zhang2024agentprune}. GPTSwarm can optimize node prompts and edges~\citep{zhuge2024gptswarm}. Revised prompts and communication rules need not occupy the same substrate. Transferable targets include when to activate specialists, which messages affect other agents' decisions, and when to stop exchanges. Message and activation logs can distinguish roles that supply evidence from those that repeat it. Removing roles or blocking edges on matched tasks tests dependency preservation, distinguishing graph reconstruction from functional replication at lower cost.

Agent Harness Distillation queries a teacher system to infer its control behavior and iteratively refine a student harness~\citep{cui2026harness}. Agreement on black-box probes supports functional replication on the tested tasks. An always-active verifier and one triggered by disagreement may agree on routine cases. Disagreement probes can distinguish some policies; activation logs can identify the executed branch. Identifying the teacher graph requires observations distinguishing it from alternatives. Studies should distinguish porting a system with control-trace access from inferring a third party's orchestration from outputs alone, and report ownership, consent, and retention conditions for the evidence.

Automated Design of Agentic Systems uses a meta-agent to program systems and retain discoveries~\citep{hu2024adas}. Its archive, like the execution and operator experience used by AFlow and A$^{2}$Flow~\citep{zhang2024aflow,zhao2025a2flow}, can supply teacher evidence when discoveries are transferred into a deployed controller. Teacher workflows can also supply traces for model training or instructions for reusable skills, transferring knowledge into other forms examined in Section~\ref{sec:7}.

\vspace{-0.3cm}
\section{Cross-Substrate Distillation and Hybrid Realizations}\label{sec:7}
Sections~\ref{sec:4}--\ref{sec:6} classify distillation by where the student retains teacher knowledge. Transfer paths describe how that knowledge is transformed along the way. Let $K_T^x$ denote teacher knowledge primarily realized in substrate $x$, $\mathcal{E}_T$ the exposed teacher evidence, and $K_S^y$ the knowledge retained in student substrate $y$. The path is
\begin{equation}
    K_T^x \xrightarrow{\text{Expose}} \mathcal{E}_T
    \xrightarrow{\Phi_{x\to y}} K_S^y,
    \qquad x,y\in\{\mathcal{M},\mathcal{R},\mathcal{H}\},
    \label{eq:cross-substrate-transfer}
\end{equation}
where $\Phi_{x\to y}$ is cross-substrate when $x\ne y$. Intermediate representations may be discarded at deployment. If controller traces are organized as skills used only to train a model, the path is $\mathcal{H}_T\to \mathcal{R}\to \mathcal{M}_S$, with one changed substrate. Retaining both a teacher-derived skill and a teacher-derived invocation rule yields a hybrid realization. We use this distinction to examine translation fidelity (\S~\ref{sec:cross-translation}), transfers across the model boundary (\S~\ref{sec:cross-model}), transfers between artifacts and harnesses (\S~\ref{sec:cross-operational}), and knowledge allocation in hybrid students (\S~\ref{sec:cross-hybrid}).

\subsection{Transfer Paths as Representation Translation}
\label{sec:cross-translation}

The three substrates define a $3\times3$ source--destination matrix. Its diagonal preserves the primary form of knowledge. The six off-diagonal paths fall into three pairs. Artifacts or control logic can be internalized in a model, model knowledge can be expressed as a resource or a controller, and transfers between artifacts and harnesses can compile guidance or extract it from execution. Several paths may contribute to one student. Identifying a destination can be easier than establishing its source. A teacher answer may combine model computation, a retrieved document, and a verifier that triggered a revision. With only the answer available, a study can identify the resulting student update while leaving the teacher source open. In that setting the supported path is $\text{source-agnostic evidence}\to y$. A textual transcript by itself establishes neither an artifact source nor a parametric one. Component access, provenance, or an intervention that varies one teacher component while holding the others fixed can support a more specific attribution.

Translation fidelity concerns the \emph{functional invariants} that should survive the change of form. A model trained on a debate may learn to compare alternatives and revise an incorrect conclusion without reproducing each message. A workflow compiled from a checklist must preserve the ordering constraint that prevents a write before verification. The level of fidelity should therefore follow the capability being transferred. Matching a debate transcript word for word places a constraint on the student that ordinary standalone inference need never satisfy. For a controller, by contrast, losing even an infrequently exercised precondition can change what actions it permits. Let $\mathcal{B}_{\mathcal{Q}}(K)$ denote the outputs, decisions, and resource use induced by $K$ on a probe set $\mathcal{Q}$. A generic objective is
\begin{equation}
    \mathcal{L}_{\mathrm{trans}}(x\to y)
    =D\!\left(\mathcal{B}_{\mathcal{Q}}(K_T^x),\mathcal{B}_{\mathcal{Q}}(K_S^y)\right)
    +\lambda\,\Omega_y(K_S^y),
    \label{eq:cross-translation-objective}
\end{equation}
where $D$ measures disagreement on the claimed invariants, and $\Omega_y$ measures a cost of retaining knowledge in $y$, such as retrieval overhead or controller complexity. If every probe passes verification, agreement provides little evidence that the translated workflow still blocks an unsafe write. Testing a failed check makes the ordering constraint observable.

For a path with several stages, each mapping can discard a different property. A skill extracted from a trace may omit a precondition, or a compiler may fail to enforce a precondition that the skill states correctly. Recording the intermediate representation lets a study distinguish the two failures even if the deployed student retains only the controller. The record explains the path even when the deployed student discards the intermediate skill.

\subsection{Crossing the Model Boundary: Internalization and Externalization}
\label{sec:cross-model}

\emph{Internalization} ($\mathcal{R}_T\to \mathcal{M}_S$ or $\mathcal{H}_T\to \mathcal{M}_S$) moves knowledge into the student's learned computation. An explicit procedure can supply demonstrations of tool use, examples of recovery, or contrasts between states where the procedure does and does not apply. The target can be an artifact-conditioned action, an explanation derived from the resource, or a preference between applicable and inapplicable uses. Withdrawing the resource at inference tests whether the model can still choose an appropriate action. This can eliminate repeated retrieval for a frequently used procedure, at the cost of making an exact instruction harder to inspect or update.

A teacher harness can also provide structure that a final answer omits. MAGDi organizes multi-model discussions as graphs of correct and incorrect reasoning steps and conversational dependencies. It uses token-level, contrastive, and graph-aware objectives to train a smaller standalone model~\citep{chen2024magdi}. The graph exposes competing answers and the revisions they prompt. The graph and the multi-agent discussion need not run at student inference, so the model is the retained destination. Retaining a teacher-derived controller for difficult cases would add a second destination (\S~\ref{sec:cross-hybrid}).

\emph{Externalization} ($\mathcal{M}_T\to \mathcal{R}_S$ or $\mathcal{M}_T\to \mathcal{H}_S$) makes knowledge attributable to a teacher model available as a reusable resource or an execution rule. \emph{Large Language Models as Tool Makers} separates a stronger tool-making model from a lighter user of its generated functions~\citep{cai2023latm}. AgentDistill packages teacher-agent capabilities as reusable Model Context Protocol boxes that smaller students can invoke without updating their parameters~\citep{qiu2025agentdistill}. Both illustrate an artifact destination. When a teacher agent consults its own tools or library during construction, the source remains agent-level unless evidence isolates the model's contribution. The artifact must also expose an interface the intended student can use. A generated function can pass its tests yet fail when the student calls the wrong version or supplies incompatible arguments.
Executing it with that student helps separate function defects from invocation mismatches.

For $\mathcal{M}_T\to \mathcal{H}_S$, the retained output is a controller. Automated Design of Agentic Systems programs candidate systems with a meta-agent~\citep{hu2024adas}, and AFlow searches over code-defined workflows using execution feedback~\citep{zhang2024aflow}. A model-to-harness claim requires a teacher-model decision principle, such as when to verify a result, that the student enforces during execution. The choice of direction changes what survives a model update. A callable resource can be revised independently, and an installed verification rule can still constrain which actions are permitted. 

\subsection{Crystallizing and Compiling Operational Knowledge}
\label{sec:cross-operational}

Artifacts and harnesses can both contain code. Their role in future episodes distinguishes the two directions of transfer. \emph{Compilation} ($\mathcal{R}\to \mathcal{H}$) turns guidance the agent can consult into logic the runtime executes. For a checklist instruction to \emph{verify the result}, a compiler must bind a verifier to the current output and specify what happens if the check fails. The controller might block the next action, request a revision, and escalate after a bounded number of attempts. Fidelity depends on preserving the conditions under which the teacher intended the check to apply. Compiling a preference as a mandatory gate can change the procedure even when the check's wording is faithfully reproduced. The compiler should distinguish conditions stated in the artifact from defaults inferred during translation. Testing an allowed exception and a case that should be blocked is more informative than comparing an execution log with the checklist. 

\emph{Crystallization} ($\mathcal{H}\to \mathcal{R}$) extracts a reusable resource from regularities in execution. Agent Workflow Memory induces workflows for later retrieval~\citep{wang2024awm}. A$^{2}$Flow extracts reusable operators from demonstrations~\citep{zhao2025a2flow}, while FlowScout mines recurring patterns of tool coordination~\citep{hao2026flowscout}. An operator or workflow is an artifact while it remains available for independent reuse. Installing its composition as a controller yields the harness destination examined in Section~\ref{sec:6}. A trace can reveal a repeated action without identifying whether the teacher's model chose it or its harness enforced it. Attributing the source to $H_T$ requires evidence about that control decision.

The useful abstraction may preserve less than the full execution. SKILL-KD uses contrasts between student failures and teacher trajectories to construct textual skill patches~\citep{shi2026skillkd}. ClawTrace organizes evidence into trace cards whose cost attribution guides skill retention and repair~\citep{yuan2026clawtrace}. In multi-agent-to-single-agent transfer, AdaSkill shows why preserving the teacher's entire coordination structure can be counterproductive when the task rewards diverse valid solutions~\citep{xu2026adaskill}. A skill should retain the precondition or tool use needed on a new task while allowing the student to choose its own subsequent actions.

The two directions can form an improvement cycle as execution of a compiled resource supplies evidence for its next revision. At iteration $k$, this yields
\begin{equation}
    \mathcal{R}^{(k)} \xrightarrow{\text{compile}} \mathcal{H}^{(k)}
    \xrightarrow{\text{execute}} \mathcal{E}^{(k)}
    \xrightarrow{\text{crystallize}} \mathcal{R}^{(k+1)}.
    \label{eq:cross-compilation-cycle}
\end{equation}
Here $\mathcal{E}^{(k)}$ records the execution of $\mathcal{H}^{(k)}$. Linking a revision to its source executions helps identify where a guard was lost or an unnecessary restriction introduced. If only $\mathcal{R}^{(k+1)}$ survives, the student has an artifact endpoint. If both the resource and a teacher-derived controller remain in use, the endpoint spans $R_S$ and $H_S$.

\subsection{Hybrid Realizations and Substrate Allocation}
\label{sec:cross-hybrid}

The deployment endpoint is determined by the student components that retain teacher knowledge. Let $Z_S^0$ and $Z_S^+$ denote the pre- and post-distillation states of component $Z$. Set $\Delta_T(Z)=1$ exactly when teacher knowledge causes a persistent change in that component. The set of changed substrates is
\begin{equation}
    \mathcal{Z}_S
    =\bigl\{Z\in\{\mathcal{M},\mathcal{R},\mathcal{H}\}:\Delta_T(Z)=1\bigr\}.
    \label{eq:cross-hybrid-endpoint}
\end{equation}
The student is hybrid when $|\mathcal{Z}_S|>1$. Kang et al.\ train small models on agent trajectories and use retrieval and code tools at inference~\citep{kang2025distillingllmagent}. If those tools and the execution loop are unchanged, Equation~\eqref{eq:cross-hybrid-endpoint} counts only $M_S$. Conversely, one trajectory can train model parameters and supply a persistent recovery rule, yielding an $M_S+H_S$ hybrid.

The choice of a second substrate depends on what the first one leaves unresolved. In an $M_S+R_S$ student, parameters can handle a frequent decision while an artifact preserves an exact or changing procedure. Consider a teacher-derived skill that records a repair procedure. One model may learn to perform the same repair. Another may learn only which failure calls for that skill. Both students have two changed substrates. The first may still perform the repair if the skill is withdrawn. In the second, recognizing the failure and carrying out the repair depend on different components.

An $M_S+H_S$ student can instead combine learned action generation with an explicit permission check or recovery branch. A controller calibrated for a strong teacher may intervene too late for a smaller student, even when both models can issue the same tool call. An $R_S+H_S$ student can keep its model frozen while acquiring both a reusable skill and a rule for invoking it. The artifact side is illustrated by AgentDistill's protocol modules~\citep{qiu2025agentdistill}. A hybrid claim in this setting also requires a teacher-derived change to the selector or executor.

For a full $M_S+R_S+H_S$ realization, all three components must retain such changes. For example, a teacher's repair episodes might train the model to diagnose a recurring error, supply a reusable repair skill, and induce a controller rule that rolls back a patch after a failed test. Each is a separate claim about what the student retains. The same episodes may instead serve entirely as training data for the model, leaving the existing library and controller intact. The components must also remain compatible. Revising a tool schema can break a trained model's arguments or invalidate an otherwise sound routing rule. If each library revision requires retraining the model, the resource loses its practical independence. A stable call signature and a record of compatible versions allow it to be revised separately.

Allocation therefore depends on workload, access, and update cost. If the student's model weights are inaccessible and no trainable adapter is available, transfer can still change the resources its model reads or the controller that dispatches its actions. When all three destinations are available, parameters suit frequent, latency-sensitive behavior, artifacts keep exact or changing knowledge available for inspection, and harnesses enforce conditions over actions as an episode unfolds. Reverting one changed component at a time under a matched budget tests whether the deployed behavior depends on it. Comparing combinations of original and updated components can then reveal gains that require a particular pairing. Section~\ref{sec:8} develops the evaluation protocols and benchmarks needed to test those claims.

\section{Evaluation and Benchmarks for Agent Distillation}\label{sec:8}

\begin{table*}[t]
    \centering
    \caption{Representative benchmarks for Agent Distillation. $\mathcal{M}$, $\mathcal{R}$, and $\mathcal{H}$ denote model, artifact, and harness substrates.}
    \label{tab:evaluation-benchmarks-survey}
    \scriptsize
    \setlength{\tabcolsep}{2.7pt}
    \renewcommand{\arraystretch}{0.94}
    \begin{tabularx}{\textwidth}{@{}>{\raggedright\arraybackslash}p{0.17\textwidth} >{\raggedright\arraybackslash}p{0.20\textwidth} >{\centering\arraybackslash}p{0.085\textwidth} >{\raggedright\arraybackslash}p{0.24\textwidth} >{\raggedright\arraybackslash}X@{}}
        \toprule
        \textbf{Benchmark} & \textbf{Setting} & \textbf{Substrate focus} & \textbf{Measures} & \textbf{Distillation reading / caveat} \\
        \midrule
        \multicolumn{5}{@{}l@{}}{\textbf{Parametric reasoning / process}} \\
            ERASER~\citep{deyoung2020eraser} & Static text; offline & M & Evidence selection; sufficiency; comprehensiveness & Process supervision and error analysis; faithfulness needs intervention. \\
            Process\-Bench~\citep{zheng2025processbench} & Static text; offline & M & Step validity; first-error localization & Reasoning supervision; no closed-loop recovery. \\
        \midrule
        \multicolumn{5}{@{}l@{}}{\textbf{Tool and stateful interaction}} \\
            API\-Bank~\citep{li2023apibank} & API calls; offline & M/H & API retrieval; call generation; response production & Tool-use transfer; model, tool, and harness effects confounded. \\
            BFCL~\citep{patil2025bfcl} & Structured/multi-turn function calls & M/H & Relevance; call validity; multi-turn success; latency/cost & Structured action supervision; interface version matters. \\
            Tool\-Sandbox~\citep{lu2025toolsandbox} & Stateful conversational tools; on-policy & H/M & Milestones; prohibited states; terminal predicates; recovery & Stateful control; substrate attribution confounded. \\
            $\tau$\-bench~\citep{yao2025taubench} & Customer-service tasks; online & H/M & Task success; policy compliance; $\mathrm{pass}^{k}$ & Multi-turn policy transfer; environment state and evaluator matter. \\
        \midrule
        \multicolumn{5}{@{}l@{}}{\textbf{Web, computer, and software}} \\
            Web\-Arena~\citep{zhou2024webarena} & Websites; executable on-policy & M/H & Task success; steps; browser state & Web interaction transfer; interface version sensitive. \\
            OSWorld~\citep{xie2024osworld} & Desktop GUI; executable on-policy & M/H & Task success; side effects; UI state & Computer-use policy transfer; environment configuration matters. \\
            SWE\-bench~\citep{jimenez2024swebench} & Code repositories; executable & M/H & Patch validity; regression tests; issue resolution & Software-agent utility; repository and harness controls matter. \\
            App\-World~\citep{trivedi2024appworld} & Tool-rich applications; online & M/H & Goal completion; app state; side effects & End-to-end utility; model, tools, and harness interact. \\
        \midrule
        \multicolumn{5}{@{}l@{}}{\textbf{Memory and skill artifacts}} \\
            Long\-Mem\-Eval~\citep{wu2025longmemeval} & Long-context; multi-session memory & R & Information extraction; retrieval; temporal reasoning; updates; abstention & Memory value; retrieval and routing remain confounded. \\
            Memory\-Agent\-Bench~\citep{hu2025memoryagentbench} & Incremental multi-turn interactions & R/H & Retrieval; test-time learning; long-range use; selective forgetting & Memory construction/use; routing controls are required. \\
            Skills\-Bench~\citep{li2026skillsbench} & No-skill/curated/self-generated skills; matched tasks & R & Skill reuse; marginal gain; verifier success & Skill transfer; selection and execution are control variables. \\
            Skill\-Learn\-Bench~\citep{zhong2026skilllearnbench} & Skill learning; real-world tasks & R/H & Skill quality; execution trajectory; task outcome & Continual-learning protocol; portability needs model/harness swaps. \\
        \midrule
        \multicolumn{5}{@{}l@{}}{\textbf{Workflow and multi-agent}} \\
            Flow\-Bench~\citep{xiao2024flowbench} & Workflow-guided tool planning; turn/session & H & Tool choice; parameter validity; workflow success/progress & Guided planning; structural match does not prove functional control. \\
            Multi\-Agent\-Bench~\citep{zhu2025multiagentbench} & Collaborative/competitive multi-agent tasks & H & Milestones; planning; communication; task outcomes & Orchestration transfer; topology alone is insufficient. \\
            MAST~\citep{cemri2025mast} & Failure-analysis taxonomy; trace-based & H & Specification; inter-agent alignment; verification; termination failures & Diagnostic failure localization; not a task-utility benchmark or teacher provenance test. \\
        \midrule
        \multicolumn{5}{@{}l@{}}{\textbf{Harness and system attribution}} \\
            Harness\-Bench~\citep{harnessbench2026} & Crossed model $\times$ harness & M/H & Usage; execution alignment; validators; traces & Crossed model--harness attribution and portability; controlled teacher evidence required. \\
            Exgentic~\citep{exgentic2026} & Unified protocol across models/environments & M/H & Cross-model/environment comparability; unified protocol & System-level evaluation; harness versions affect comparability. \\
        \midrule
        \multicolumn{5}{@{}l@{}}{\textbf{Safety and security}} \\
            Agent\-Dojo~\citep{debenedetti2024agentdojo} & Adversarial tools; dynamic online environment & M/R/H & Benign success; attack success; policy violations & Safety-aware transfer; utility--safety trade-off remains. \\
            ST\-Web\-Agent\-Bench~\citep{levy2024stwebagentbench} & Web tasks; policy-constrained/adversarial & M/H & Policy compliance; task completion; unsafe-action rate & Secure policy transfer; policy and evaluator configuration matters. \\
            Agent\-Harm~\citep{andriushchenko2025agentharm} & Harmful requests; adversarial multi-step tasks & M/H & Harmful-task completion; unsafe actions & Safety behavior transfer; refusal is complementary evidence. \\
            ASB~\citep{zhang2025asb} & Agent security; adversarial attack surfaces & M/R/H & Attack success/defense robustness across prompt, memory, planning, and tool surfaces & Security-oriented harness transfer; evaluator assumptions matter. \\
        \bottomrule
    \end{tabularx}
    \vspace{-0.3cm}
\end{table*}

Evaluating Agent Distillation requires connecting a persistent change in the student to its behavior on subsequent tasks.
In the repository example of Section~\ref{sec:2}, suppose a targeted test passes but the full suite reveals a regression. The student may roll back the patch because its model has learned the teacher's recovery rule, a retrieved skill recommends rollback, or the harness enforces it.
A passing patch is compatible with all three explanations.
We examine what evidence distinguishes these destinations (\S~\ref{sec:eval-objectives}), how controlled comparisons measure their contributions to deployed performance (\S~\ref{sec:eval-protocol}), and which of these comparisons existing benchmarks support (\S~\ref{sec:benchmark-landscape}).

\subsection{Evidence of Knowledge Transfer}
\label{sec:eval-objectives}

A teacher trace records actions within an episode. Attributing an action to a teacher component requires access to it, a record linking it to the action, or an intervention isolating its effect.
On the student side, inspect the components in $\mathcal{Z}_S$ that retain a change after distillation (\S~\ref{sec:7}). A workflow used only to generate training examples can leave its sole persistent change in the model.
Evidence of \emph{retention} identifies the behavior, resource, or control rule still available after supervision ends. Intervening on the changed component under matched conditions estimates its \emph{causal contribution} to later execution, while performance on held-out tasks measures the \emph{utility} of the complete student.

\textbf{Model parameters.}
For a parametric destination, compare $\mathcal{M}_S^0$ and $\mathcal{M}_S^+$ with the same student artifacts and harness, after withdrawing demonstrations or privileged feedback supplied only during training.
At a shared interaction history, comparing teacher and student actions measures local behavioral fidelity. When the action spaces align, logits permit comparison of their action distributions~\citep{hinton2015distilling,stanton2021does}. Access to hidden states also permits comparison of their representations~\citep{kornblith2019similarity}.
A student that reproduces the teacher's explanation may still rely on a different cue. Perturbing the evidence cited in that explanation and observing the decision probes its faithfulness~\citep{deyoung2020eraser,turpin2023unfaithful}.
Held-out execution then tests whether the student can apply the learned behavior to histories it encounters itself.

\textbf{Reusable artifacts.}
For a memory, tool, or skill retained in $\mathcal{R}_S$, evaluation follows the resource from storage to use.
A debugging skill may prescribe the correct verification procedure yet never be selected. Even when retrieved, it may have no effect if the model already follows that procedure.
Checking the skill's contents or running its tests establishes validity, logging retrieval on applicable tasks establishes access, and comparing executions with the original and distilled resources estimates the benefit of making the new skill available.
To assess maintenance after a skill is revised or revoked, check affected tasks for the intended change and unrelated tasks for regressions.

\textbf{Execution harnesses.}
A verification rule retained in $\mathcal{H}_S$ governs what the agent can do after a test result arrives.
Exercising the pass and fail branches reveals whether the controller proceeds, retries, or rolls back under the intended conditions. The same principle applies to handoffs and stopping rules in a multi-agent harness.
Controllers with the same workflow graph can act differently if a branch condition differs. Conversely, distinct graphs can implement the same recovery behavior.
When the teacher controller is accessible, its guards and transitions provide a reference for structural recovery. With only teacher traces, agreement supports behavioral fidelity over the conditions exercised. 

\subsection{Controlled Evaluation Protocols}
\label{sec:eval-protocol}

The teacher $\mathcal{A}_T$ provides a reference for capability preservation. The undistilled student $\mathcal{A}_S^0$ provides the baseline for measuring what the deployed student $\mathcal{A}_S^+$ gained.
Evaluate these agents on common tasks and initial states, and record which tools, resources, and supervision each can access. For a comparison intended to isolate a student component, hold the other components, permissions, and execution budgets fixed.
When the claim concerns a particular teacher signal, compare against alternative supervision matched in volume and training budget. This tests what the signal contributes beyond additional training.
Report the raw teacher and student scores. The fraction of the teacher--student gap closed is informative when the teacher has a positive advantage large enough for the ratio to be stable.

Offline evaluation presents each agent with the same logged history and compares its next decision. Once the student chooses a different action, the log may contain no observation of the state it reaches.
On-policy evaluation lets the student generate that history from a controlled initial state. Executable tests and terminal-state predicates can then credit different trajectories that satisfy the same goal.
Milestones reveal partial progress~\citep{ma2024agentboard,lu2025toolsandbox}. For a failed run, annotating the first error that compromised the goal can identify a missed opportunity for recovery.
For open-ended outcomes, calibrate model judgments against held-out human assessments or executable checks, report the rubric and agreement, and randomize response order in paired judgments~\citep{zheng2023judging,liu2023geval}.

\textbf{Component contribution.}
Let $U(\mathcal{A})$ denote expected task utility under a fixed task distribution, environment, and budget, as in Section~\ref{sec:2}.
Restoring a changed component $Z$ from its distilled state $Z_S^+$ to its original state $Z_S^0$ on the same tasks gives the paired utility difference
\begin{equation}
    \Delta_Z = U(\mathcal{A}_S^+) - U\!\left(\mathcal{A}_S^+[Z_S^+ \leftarrow Z_S^0]\right).
    \label{eq:component-contribution}
\end{equation}
For example, restoring the original skill library while retaining the distilled router measures the new library's benefit under that router.
Repeating the comparison with another compatible router tests whether the benefit carries over. A swap that breaks the skill interface cannot distinguish loss of the benefit from an execution failure.

\textbf{Interaction between components.}
When both a skill library and its router change, either component's measured benefit can depend on the version of the other.
Evaluate all four combinations of their original and distilled versions. For components $Z_1$ and $Z_2$, let $U_{ab}$ denote expected utility with $Z_1$ in version $a$ and $Z_2$ in version $b$, where $a,b\in\{0,+\}$ and all other conditions are fixed.
Their interaction on this utility scale is
\begin{equation}
    \operatorname{Syn}(Z_1,Z_2) = (U_{++}-U_{0+}) - (U_{+0}-U_{00}).
    \label{eq:substrate-synergy}
\end{equation}
Taking the library as $Z_1$, a positive interaction means that updating the router increases the library's marginal contribution. An interaction near zero is consistent with two useful, additive changes.
The hybrid designation concerns where teacher-derived changes persist. The four comparisons reveal how those changes work together.

\textbf{Generalization.}
Hold out the unit to which the transfer claim applies. Reusing a debugging strategy across projects calls for a repository-level split, since traces from the same repair can share the solution.
Transferring a skill across models instead requires testing the same resource on a held-out compatible model and measuring its gain there~\citep{zhang2026generalizability}.
Record the origin and filtering of the teacher evidence, including which observations came from tools or the environment, and preserve versions of the initial states, artifacts, and controllers used in the comparisons~\citep{gebru2021datasheets,bender2018datastatements,pushkarna2022datacards}.
These records make it possible to reproduce the original comparison before introducing a new environment or tool schema.

\textbf{Reliability and cost.}
Repeat stochastic runs on the same task instances and report uncertainty across tasks while keeping repeated runs for each task together.
Average success describes a typical run. The $\mathrm{pass}^{k}$ criterion in $\tau$-bench asks whether all $k$ independent runs on the same task succeed~\citep{yao2025taubench}.
Report the one-time costs of collecting teacher evidence, training the model, or constructing artifacts separately from the tokens, tool calls, and elapsed time consumed during deployment~\citep{kapoor2025agents}.
If the agents require different budgets, compare utility across budgets. When ease of maintenance motivates the transfer, include the cost of revising the resource or controller and validating it after the revision.

\subsection{Benchmark Coverage}
\label{sec:benchmark-landscape}

Table~\ref{tab:evaluation-benchmarks-survey} groups representative evaluation resources by setting and records what they measure. The substrate focus indicates which student component a study could investigate with each resource.
The comparisons below illustrate which conclusions each resource supports on its own and which require additional controls.

\textbf{From local decisions to task outcomes.}
ProcessBench asks a model to locate the first error in a supplied solution. Success demonstrates error recognition on a fixed reasoning history~\citep{zheng2025processbench}.
ToolSandbox tracks milestones and prohibited states during tool interaction, while $\tau$-bench checks the resulting database state~\citep{lu2025toolsandbox,yao2025taubench}. Both expose consequences of the student's own actions that a next-action label cannot capture.
The repository example illustrates a further limit. SWE-bench can validate a submitted patch through tests~\citep{jimenez2024swebench}. To evaluate a distilled rollback strategy on such a task, the experiment must also present an opportunity to recover and record how the student responds.

\textbf{Memory and skill use.}
LongMemEval tests whether an agent can answer questions about earlier interactions, including after information has been updated. MemoryAgentBench also examines memory use as new interactions arrive~\citep{wu2025longmemeval,hu2025memoryagentbench}.
A correct answer can draw on information already available in context, so attributing it to a distilled memory requires varying access to the relevant entry.
SkillsBench includes matched conditions with and without skills, and SkillLearnBench relates the quality of a generated skill to its execution and outcome~\citep{li2026skillsbench,zhong2026skilllearnbench}.
For a teacher-derived skill, record its provenance and keep the selector fixed. If its text also lengthens the prompt, a length-matched control helps distinguish the value of its instructions from the effect of additional context.

\textbf{Workflow and harness behavior.}
FlowBench measures workflow-guided tool choices, while MultiAgentBench tracks milestones and communication in multi-agent tasks~\citep{xiao2024flowbench,zhu2025multiagentbench}. MAST provides categories for diagnosing failures in these traces, including coordination and termination errors~\citep{cemri2025mast}.
To test how much the controller contributes, Harness-Bench varies models and harnesses on shared tasks~\citep{harnessbench2026}. Exgentic provides a common evaluation protocol across models and environments~\citep{exgentic2026}.
Such comparisons can reveal that a controller helps one model more than another. Tracing the gain to distillation additionally requires the teacher evidence used to construct the controller and a comparison with its pre-distillation version.

\textbf{Safety under deployment.}
AgentDojo pairs benign tasks with prompt-injection attacks, while ST-WebAgentBench evaluates task completion under specified policies~\citep{debenedetti2024agentdojo,levy2024stwebagentbench}.
Report benign completion alongside unsafe actions or attack success. Declining every tool call can suppress attacks at the cost of legitimate work.
If a distilled permission rule is credited with the improvement, repeat the benign and adversarial tasks with the prior version of that rule. For a claim about memory privacy, test whether revoked entries remain accessible or information can be retrieved across users.

Across these settings, tasks and evaluators can be reused across studies. Reproducing a transfer claim also requires a record linking the teacher evidence to the student component versions being compared.
Maintaining that record as agents and environments change is one of the open challenges examined in the next section.
\section{Open Challenges and Future Directions}
\label{sec:9}

Controlled evaluation estimates a retained component's contribution under specified conditions. Open challenges concern attribution, environmental adaptation, and knowledge allocation and maintenance across substrates.

\textbf{ Identifiability under partial access.}
Component interventions require accessible versions and compatible interfaces. Harness-Bench and Exgentic offer starting points for such comparisons~\citep{harnessbench2026,exgentic2026}. Teacher traces alone may be consistent with distinct component combinations that produce the same observed behavior. The challenge is to determine which claims about the transfer source and retained destination are identifiable under partial access, and what additional provenance or interventions can resolve the ambiguity.

\textbf{ Adaptation to environmental change.} \label{sec:challenge-reproducibility}
Distilled knowledge may become obsolete as tool schemas, permissions, or task distributions change. ToolSandbox and AppWorld illustrate how explicit state and executable validators support controlled baselines~\citep{lu2025toolsandbox,trivedi2024appworld}. Extending these environments to successive changes could distinguish temporary recovery from knowledge useful in later episodes. The key questions are how much new teacher evidence adaptation requires, which components need revision, and how the accumulated cost compares across transfer destinations. 

\textbf{ Joint design across knowledge substrates.} \label{sec:challenge-allocation}
Most methods fix the destination before learning. Joint design would optimize retained knowledge and its allocation across model parameters, artifacts, and harnesses as workloads and update needs change. A model could learn to invoke a distilled skill while a controller validates its use, yet the gain may depend on that pairing. Allocation objectives should account for deployment costs, revision costs, and portability. Resource schemas and tool contracts would support comparisons across compatible models and controllers.

\textbf{ Safety across the knowledge lifecycle.} \label{sec:challenge-lifecycle-safety}
Distilled knowledge may remain in use after facts or permissions change. Provenance and version records make revisions traceable, but deleting a library entry may leave its effects in a model or dependent workflow. The challenge is to propagate updates and revocations across these dependencies while preserving legitimate capabilities. AgentDojo, ST-WebAgentBench, and Agent Security Bench provide attack and policy-compliance settings~\citep{debenedetti2024agentdojo,levy2024stwebagentbench,zhang2025asb} that could be extended to study persistent vulnerabilities across revisions. Combining dependency tracking, deletion, rollback, and least-privilege execution could support safer maintenance.

\textbf{ Shared foundations for comparison.} \label{sec:challenge-science}
Parameter counts, artifact size, and harness complexity quantify different resources. A common theory should relate functional equivalence to resource requirements under specified tasks and constraints. Systems with interchangeable models, artifacts, and harnesses could test these relationships under common workloads, accounting for tool dependencies and acquisition, deployment, and maintenance costs.

\section{Conclusion}
\label{sec:10}

Agent Distillation derives task-solving knowledge from teacher-agent evidence and transfers it to a student for use in future episodes. This persistence distinguishes distillation from teacher consultation whose effects end with the interaction. Under our decomposition $\mathcal{A}=(\mathcal{M},\mathcal{R},\mathcal{H})$, model parameters support self-contained execution, artifacts preserve resources for revision and reuse, and harnesses encode explicit rules for control and recovery. A cross-substrate path describes how knowledge changes form during transfer. In a hybrid realization, teacher-derived changes persist in more than one student component. Tracing a path requires evidence about the teacher components that produced the observed behavior. On the student side, inspection characterizes what a component retains, intervention tests its contribution, and the student's own interactions reveal its utility in context. 
A portfolio of benchmarks can test whether the capability generalizes. 
Reporting reliability, safety and lifecycle cost alongside task performance clarifies deployment trade-offs and provides a common basis for comparing transfer strategies and designing reusable agents.

\clearpage
\bibliographystyle{ACM-Reference-Format-short}
\bibliography{reference/sample-base,reference/sec4_references,reference/sec8_references,reference/sec5_references,reference/sec6_references,reference/sec7_references}

@inproceedings{agarwal2024gkd,
  author        = {Agarwal, Rishabh and Vieillard, Nino and Zhou, Yongchao and Stanczyk, Piotr and Ramos Garea, Sabela and Geist, Matthieu and Bachem, Olivier},
  title         = {On-Policy Distillation of Language Models: Learning from Self-Generated Mistakes},
  booktitle     = {The Twelfth International Conference on Learning Representations},
  volume        = {2024},
  pages         = {21246--21263},
  year          = {2024},
  url           = {https://proceedings.iclr.cc/paper_files/paper/2024/hash/5be69a584901a26c521c2b51e40a4c20-Abstract-Conference.html}
}

@misc{anthropic_claude_code,
  author        = {{Anthropic}},
  title         = {Claude Code Documentation},
  year          = {2026},
  url           = {https://code.claude.com/docs/en/overview},
  howpublished  = {Official software documentation},
  note          = {Accessed 2026-09-27}
}

@inproceedings{brown2020language,
  author        = {Brown, Tom B. and Mann, Benjamin and Ryder, Nick and Subbiah, Melanie and Kaplan, Jared D. and Dhariwal, Prafulla and Neelakantan, Arvind and Shyam, Pranav and Sastry, Girish and Askell, Amanda and Agarwal, Sandhini and Herbert-Voss, Ariel and Krueger, Gretchen and Henighan, Tom and Child, Rewon and Ramesh, Aditya and Ziegler, Daniel M. and Wu, Jeffrey and Winter, Clemens and Hesse, Christopher and Chen, Mark and Sigler, Eric and Litwin, Mateusz and Gray, Scott and Chess, Benjamin and Clark, Jack and Berner, Christopher and McCandlish, Sam and Radford, Alec and Sutskever, Ilya and Amodei, Dario},
  title         = {Language Models are Few-Shot Learners},
  booktitle     = {Advances in Neural Information Processing Systems},
  volume        = {33},
  pages         = {1877--1901},
  year          = {2020},
  url           = {https://papers.neurips.cc/paper/2020/hash/1457c0d6bfcb4967418bfb8ac142f64a-Abstract.html}
}

@misc{cui2026harness,
  author        = {Cui, Yu and Yang, Wuli and Shi, Yirui and Xia, Junhao and Jiang, Hui and Gao, Lei and Bao, Chenfu},
  title         = {Agent Harness Distillation: Inference-Time Harness Extraction and Exploitation in Autonomous Multi-Agent Systems},
  year          = {2026},
  doi           = {10.48550/arXiv.2607.28147},
  url           = {https://arxiv.org/abs/2607.28147},
  eprint        = {2607.28147},
  archiveprefix = {arXiv},
  howpublished  = {arXiv preprint}
}

@article{gao2026selfevolving,
  author        = {Gao, Huan-ang and Geng, Jiayi and Hua, Wenyue and Hu, Mengkang and Juan, Xinzhe and Liu, Hongzhang and Liu, Shilong and Qiu, Jiahao and Qi, Xuan and Wu, Yiran and Wang, Hongru and Xiao, Han and Zhou, Yuhang and Zhang, Shaokun and Zhang, Jiayi and Xiang, Jinyu and Fang, Yixiong and Zhao, Qiwen and Liu, Dongrui and Ren, Qihan and Qian, Cheng and Wang, Zhenhailong and Hu, Minda and Wang, Huazheng and Wu, Qingyun and Ji, Heng and Wang, Mengdi},
  title         = {A Survey of Self-Evolving Agents: What, When, How, and Where to Evolve on the Path to Artificial Super Intelligence},
  journal       = {Transactions on Machine Learning Research},
  year          = {2026},
  url           = {https://arxiv.org/abs/2507.21046},
  issn          = {2835-8856}
}

@article{gou2021knowledge,
  author        = {Gou, Jianping and Yu, Baosheng and Maybank, Stephen J. and Tao, Dacheng},
  title         = {Knowledge Distillation: A Survey},
  journal       = {International Journal of Computer Vision},
  volume        = {129},
  number        = {6},
  pages         = {1789--1819},
  publisher     = {Springer US},
  year          = {2021},
  doi           = {10.1007/s11263-021-01453-z},
  url           = {https://link.springer.com/article/10.1007/s11263-021-01453-z?error=cookies_not_supported&code=9a0af4ca-7905-4aed-a571-46e58c258cba}
}

@inproceedings{gu2024minillm,
  author        = {Gu, Yuxian and Dong, Li and Wei, Furu and Huang, Minlie},
  title         = {MiniLLM: Knowledge Distillation of Large Language Models},
  booktitle     = {International Conference on Learning Representations},
  volume        = {2024},
  pages         = {32694--32717},
  year          = {2024},
  url           = {https://proceedings.iclr.cc/paper_files/paper/2024/hash/8ac015d409635f196f9e3e9dcfb9a94e-Abstract-Conference.html}
}

@inproceedings{guo2024multiagent,
  author        = {Guo, Taicheng and Chen, Xiuying and Wang, Yaqi and Chang, Ruidi and Pei, Shichao and Chawla, Nitesh V. and Wiest, Olaf and Zhang, Xiangliang},
  title         = {Large Language Model Based Multi-agents: A Survey of Progress and Challenges},
  booktitle     = {Proceedings of the Thirty-Third International Joint Conference on Artificial Intelligence},
  volume        = {9},
  pages         = {8048--8057},
  year          = {2024},
  doi           = {10.24963/ijcai.2024/890},
  url           = {https://www.ijcai.org/proceedings/2024/890}
}

@misc{guo2026harness,
  author        = {Guo, Jianyuan and Hao, Zhiwei and Wang, Chengcheng and Fan, Cheng and Luo, Tingzhang and Li, Hongguang and Gao, Ying and Mei, Hefei and Peng, Jiankun and Xu, Rongjian and Dong, Minjing and Wu, Han and Zheng, Mengyu and Han, Kai and Wang, Shiqi and Xu, Chang and Wang, Yunhe},
  title         = {From Question Answering to Task Completion: A Survey on Agent System and Harness Design},
  year          = {2026},
  doi           = {10.48550/arXiv.2606.20683},
  url           = {https://arxiv.org/abs/2606.20683},
  eprint        = {2606.20683},
  archiveprefix = {arXiv},
  howpublished  = {arXiv preprint}
}

@misc{hinton2015distilling,
  author        = {Hinton, Geoffrey and Vinyals, Oriol and Dean, Jeff},
  title         = {Distilling the Knowledge in a Neural Network},
  year          = {2015},
  doi           = {10.48550/arXiv.1503.02531},
  url           = {https://arxiv.org/abs/1503.02531},
  eprint        = {1503.02531},
  archiveprefix = {arXiv},
  primaryclass  = {stat.ML},
  howpublished  = {arXiv preprint}
}

@misc{huang2024planning,
  author        = {Huang, Xu and Liu, Weiwen and Chen, Xiaolong and Wang, Xingmei and Wang, Hao and Lian, Defu and Wang, Yasheng and Tang, Ruiming and Chen, Enhong},
  title         = {Understanding the Planning of LLM Agents: A Survey},
  year          = {2024},
  doi           = {10.48550/arXiv.2402.02716},
  url           = {https://arxiv.org/abs/2402.02716},
  eprint        = {2402.02716},
  archiveprefix = {arXiv},
  howpublished  = {arXiv preprint}
}

@inproceedings{jimenez2024swebench,
  author        = {Jimenez, Carlos E. and Yang, John and Wettig, Alexander and Yao, Shunyu and Pei, Kexin and Press, Ofir and Narasimhan, Karthik},
  title         = {SWE-bench: Can Language Models Resolve Real-World GitHub Issues?},
  booktitle     = {International Conference on Learning Representations},
  volume        = {2024},
  pages         = {54107--54157},
  year          = {2024},
  url           = {https://proceedings.iclr.cc/paper_files/paper/2024/hash/edac78c3e300629acfe6cbe9ca88fb84-Abstract-Conference.html}
}

@article{kapoor2025agents,
  author        = {Kapoor, Sayash and Stroebl, Benedikt and Siegel, Zachary S. and Nadgir, Nitya and Narayanan, Arvind},
  title         = {{AI} Agents That Matter},
  journal       = {Transactions on Machine Learning Research},
  year          = {2025},
  url           = {https://openreview.net/forum?id=Zy4uFzMviZ},
  issn          = {2835-8856}
}

@inproceedings{kim2016sequence,
  author        = {Kim, Yoon and Rush, Alexander M.},
  title         = {Sequence-Level Knowledge Distillation},
  booktitle     = {Proceedings of the 2016 Conference on Empirical Methods in Natural Language Processing},
  pages         = {1317--1327},
  publisher     = {Association for Computational Linguistics},
  address       = {Austin, Texas},
  year          = {2016},
  doi           = {10.18653/v1/D16-1139},
  url           = {https://aclanthology.org/D16-1139/}
}

@inproceedings{ko2024distillm,
  author        = {Ko, Jongwoo and Kim, Sungnyun and Chen, Tianyi and Yun, Se-Young},
  title         = {{D}isti{LLM}: Towards Streamlined Distillation for Large Language Models},
  booktitle     = {Proceedings of the 41st International Conference on Machine Learning},
  series        = {Proceedings of Machine Learning Research},
  volume        = {235},
  pages         = {24872--24895},
  publisher     = {PMLR},
  year          = {2024},
  url           = {https://proceedings.mlr.press/v235/ko24c.html}
}

@inproceedings{li2023apibank,
  author        = {Li, Minghao and Zhao, Yingxiu and Yu, Bowen and Song, Feifan and Li, Hangyu and Yu, Haiyang and Li, Zhoujun and Huang, Fei and Li, Yongbin},
  title         = {API-Bank: A Comprehensive Benchmark for Tool-Augmented {LLM}s},
  booktitle     = {Proceedings of the 2023 Conference on Empirical Methods in Natural Language Processing},
  pages         = {3102--3116},
  publisher     = {Association for Computational Linguistics},
  address       = {Singapore},
  year          = {2023},
  doi           = {10.18653/v1/2023.emnlp-main.187},
  url           = {https://aclanthology.org/2023.emnlp-main.187/}
}

@inproceedings{li2023camel,
  author        = {Li, Guohao and Hammoud, Hasan Abed Al Kader and Itani, Hani and Khizbullin, Dmitrii and Ghanem, Bernard},
  title         = {{CAMEL}: Communicative Agents for ``Mind'' Exploration of Large Language Model Society},
  booktitle     = {Advances in Neural Information Processing Systems},
  volume        = {36},
  pages         = {51991--52008},
  year          = {2023},
  doi           = {10.52202/075280-2264},
  url           = {https://proceedings.neurips.cc/paper/2023/hash/a3621ee907def47c1b952ade25c67698-Abstract-Conference.html}
}

@inproceedings{liu2023agentbench,
  author        = {Liu, Xiao and Yu, Hao and Zhang, Hanchen and Xu, Yifan and Lei, Xuanyu and Lai, Hanyu and Gu, Yu and Ding, Hangliang and Men, Kaiwen and Yang, Kejuan and Zhang, Shudan and Deng, Xiang and Zeng, Aohan and Du, Zhengxiao and Zhang, Chenhui and Shen, Sheng and Zhang, Tianjun and Su, Yu and Sun, Huan and Huang, Minlie and Dong, Yuxiao and Tang, Jie},
  title         = {AgentBench: Evaluating {LLM}s as Agents},
  booktitle     = {International Conference on Learning Representations},
  volume        = {2024},
  pages         = {52989--53046},
  year          = {2024},
  url           = {https://proceedings.iclr.cc/paper_files/paper/2024/hash/e9df36b21ff4ee211a8b71ee8b7e9f57-Abstract-Conference.html}
}

@inproceedings{madaan2023selfrefine,
  author        = {Madaan, Aman and Tandon, Niket and Gupta, Prakhar and Hallinan, Skyler and Gao, Luyu and Wiegreffe, Sarah and Alon, Uri and Dziri, Nouha and Prabhumoye, Shrimai and Yang, Yiming and Gupta, Shashank and Majumder, Bodhisattwa Prasad and Hermann, Katherine and Welleck, Sean and Yazdanbakhsh, Amir and Clark, Peter},
  title         = {{Self-Refine}: Iterative Refinement with Self-Feedback},
  booktitle     = {Advances in Neural Information Processing Systems},
  volume        = {36},
  pages         = {46534--46594},
  year          = {2023},
  url           = {https://arxiv.org/abs/2303.17651}
}

@article{mansourian2025comprehensive,
  author        = {Mansourian, Amir M. and Ahmadi, Rozhan and Ghafouri, Masoud and Babaei, Amir Mohammad and Golezani, Elaheh Badali and Ghamchi, Zeynab Yasamani and Ramezanian, Vida and Taherian, Alireza and Dinashi, Kimia and Miri, Amirali and Kasaei, Shohreh},
  title         = {A Comprehensive Survey on Knowledge Distillation},
  journal       = {Transactions on Machine Learning Research},
  year          = {2025},
  url           = {https://arxiv.org/abs/2503.12067},
  issn          = {2835-8856}
}

@misc{mukherjee2023orca,
  author        = {Mukherjee, Subhabrata and Mitra, Arindam and Jawahar, Ganesh and Agarwal, Sahaj and Palangi, Hamid and Awadallah, Ahmed},
  title         = {{Orca}: Progressive Learning from Complex Explanation Traces of {GPT}-4},
  year          = {2023},
  doi           = {10.48550/arXiv.2306.02707},
  url           = {https://arxiv.org/abs/2306.02707},
  eprint        = {2306.02707},
  archiveprefix = {arXiv},
  primaryclass  = {cs.CL},
  howpublished  = {arXiv preprint}
}

@misc{openai_codex,
  author        = {{OpenAI}},
  title         = {Codex},
  year          = {2026},
  url           = {https://developers.openai.com/codex/},
  howpublished  = {Official software documentation},
  note          = {Accessed 2026-09-27}
}

@misc{openclaw2026,
  author        = {{OpenClaw Contributors}},
  title         = {{OpenClaw}: Personal AI Assistant and Agent Runtime},
  year          = {2026},
  url           = {https://github.com/openclaw/openclaw},
  howpublished  = {Software repository},
  note          = {Accessed 2026-09-27}
}

@misc{packer2023memgpt,
  author        = {Packer, Charles and Wooders, Sarah and Lin, Kevin and Fang, Vivian and Patil, Shishir G. and Stoica, Ion and Gonzalez, Joseph E.},
  title         = {{MemGPT}: Towards {LLM}s as Operating Systems},
  year          = {2023},
  doi           = {10.48550/arXiv.2310.08560},
  url           = {https://arxiv.org/abs/2310.08560},
  eprint        = {2310.08560},
  archiveprefix = {arXiv},
  primaryclass  = {cs.AI},
  howpublished  = {arXiv preprint}
}

@inproceedings{park2023generative,
  author        = {Park, Joon Sung and O'Brien, Joseph C. and Cai, Carrie J. and Morris, Meredith Ringel and Liang, Percy and Bernstein, Michael S.},
  title         = {Generative Agents: Interactive Simulacra of Human Behavior},
  booktitle     = {Proceedings of the 36th Annual ACM Symposium on User Interface Software and Technology},
  pages         = {1--22},
  publisher     = {ACM},
  year          = {2023},
  doi           = {10.1145/3586183.3606763},
  url           = {https://doi.org/10.1145/3586183.3606763}
}

@inproceedings{ross2011dagger,
  author        = {Ross, Stephane and Gordon, Geoffrey and Bagnell, Drew},
  title         = {A Reduction of Imitation Learning and Structured Prediction to No-Regret Online Learning},
  booktitle     = {Proceedings of the Fourteenth International Conference on Artificial Intelligence and Statistics},
  series        = {Proceedings of Machine Learning Research},
  volume        = {15},
  pages         = {627--635},
  publisher     = {JMLR Workshop and Conference Proceedings},
  address       = {Fort Lauderdale, FL, USA},
  year          = {2011},
  url           = {https://proceedings.mlr.press/v15/ross11a.html}
}

@book{russell2020artificial,
  author        = {Russell, Stuart J. and Norvig, Peter},
  title         = {Artificial Intelligence: A Modern Approach},
  edition       = {4},
  publisher     = {Pearson},
  year          = {2020},
  url           = {https://aima.eecs.berkeley.edu/translations.html},
  isbn          = {9780134610993}
}

@inproceedings{schick2023toolformer,
  author        = {Schick, Timo and Dwivedi-Yu, Jane and Dess{\`i}, Roberto and Raileanu, Roberta and Lomeli, Maria and Hambro, Eric and Zettlemoyer, Luke and Cancedda, Nicola and Scialom, Thomas},
  title         = {Toolformer: Language Models Can Teach Themselves to Use Tools},
  booktitle     = {Advances in Neural Information Processing Systems},
  volume        = {36},
  pages         = {68539--68551},
  year          = {2023},
  doi           = {10.52202/075280-2997},
  url           = {https://proceedings.neurips.cc/paper_files/paper/2023/hash/d842425e4bf79ba039352da0f658a906-Abstract-Conference.html}
}

@inproceedings{shang2025agentsquare,
  author        = {Shang, Yu and Li, Yu and Zhao, Keyu and Ma, Likai and Liu, Jiahe and Xu, Fengli and Li, Yong},
  title         = {AgentSquare: Automatic LLM Agent Search in Modular Design Space},
  booktitle     = {International Conference on Learning Representations},
  year          = {2025},
  url           = {https://openreview.net/forum?id=mPdmDYIQ7f}
}

@inproceedings{shinn2023reflexion,
  author        = {Shinn, Noah and Cassano, Federico and Gopinath, Ashwin and Narasimhan, Karthik and Yao, Shunyu},
  title         = {Reflexion: Language Agents with Verbal Reinforcement Learning},
  booktitle     = {Advances in Neural Information Processing Systems},
  volume        = {36},
  pages         = {8634--8652},
  year          = {2023},
  doi           = {10.52202/075280-0377},
  url           = {https://proceedings.neurips.cc/paper_files/paper/2023/hash/1b44b878bb782e6954cd888628510e90-Abstract-Conference.html}
}

@misc{sima2025sima2,
  author        = {{SIMA team} and Bolton, Adrian and Lerchner, Alexander and Cordell, Alexandra and Moufarek, Alexandre and Bolt, Andrew and Lampinen, Andrew and Mitenkova, Anna and Hallingstad, Arne Olav and Vujatovic, Bojan and Li, Bonnie and Lu, Cong and Wierstra, Daan and Sawyer, Daniel P. and Slater, Daniel and Reichert, David and Vercelli, Davide and Hassabis, Demis and Hudson, Drew A. and Williams, Duncan and Hirst, Ed and Pardo, Fabio and Hill, Felix and Besse, Frederic and Openshaw, Hannah and Chan, Harris and Soyer, Hubert and Wang, Jane X. and Clune, Jeff and Agapiou, John and Reid, John and Marino, Joseph and Kim, Junkyung and Gregor, Karol and Sridhar, Kaustubh and McKinney, Kay and Kampis, Laura and Zhang, Lei M. and Matthey, Loic and Wang, Luyu and Raad, Maria Abi and Loks-Thompson, Maria and Engelcke, Martin and Kecman, Matija and Jackson, Matthew and Gazeau, Maxime and Purkiss, Ollie and Knagg, Oscar and Stys, Peter and Mendolicchio, Piermaria and Hadsell, Raia and Ke, Rosemary and Faulkner, Ryan and Chakera, Sarah and Baveja, Satinder Singh and Legg, Shane and Kashem, Sheleem and Terzi, Tayfun and Keck, Thomas and Harley, Tim and Scholtes, Tim and Roberts, Tyson and Mnih, Volodymyr and Liu, Yulan and Wang, Zhengdong and Ghahramani, Zoubin},
  title         = {{SIMA} 2: A Generalist Embodied Agent for Virtual Worlds},
  year          = {2025},
  doi           = {10.48550/arXiv.2512.04797},
  url           = {https://arxiv.org/abs/2512.04797},
  eprint        = {2512.04797},
  archiveprefix = {arXiv},
  howpublished  = {arXiv preprint}
}

@article{sumers2024cognitive,
  author        = {Sumers, Theodore R. and Yao, Shunyu and Narasimhan, Karthik and Griffiths, Thomas L.},
  title         = {Cognitive Architectures for Language Agents},
  journal       = {Transactions on Machine Learning Research},
  year          = {2024},
  url           = {https://openreview.net/forum?id=1i6ZCvflQJ},
  issn          = {2835-8856}
}

@misc{tishby2000information,
  author        = {Tishby, Naftali and Pereira, Fernando C. and Bialek, William},
  title         = {The Information Bottleneck Method},
  year          = {2000},
  doi           = {10.48550/arXiv.physics/0004057},
  url           = {https://arxiv.org/abs/physics/0004057},
  eprint        = {physics/0004057},
  archiveprefix = {arXiv},
  howpublished  = {arXiv preprint}
}

@article{wang2023voyager,
  author        = {Wang, Guanzhi and Xie, Yuqi and Jiang, Yunfan and Mandlekar, Ajay and Xiao, Chaowei and Zhu, Yuke and Fan, Linxi and Anandkumar, Anima},
  title         = {Voyager: An Open-Ended Embodied Agent with Large Language Models},
  journal       = {Transactions on Machine Learning Research},
  year          = {2024},
  url           = {https://arxiv.org/abs/2305.16291},
  issn          = {2835-8856}
}

@article{wang2024autonomous,
  author        = {Wang, Lei and Ma, Chen and Feng, Xueyang and Zhang, Zeyu and Yang, Hao and Zhang, Jingsen and Chen, Zhiyuan and Tang, Jiakai and Chen, Xu and Lin, Yankai and Zhao, Wayne Xin and Wei, Zhewei and Wen, Ji-Rong},
  title         = {A Survey on Large Language Model Based Autonomous Agents},
  journal       = {Frontiers of Computer Science},
  volume        = {18},
  number        = {6},
  pages         = {186345},
  publisher     = {Higher Education Press},
  year          = {2024},
  doi           = {10.1007/s11704-024-40231-1},
  url           = {https://link.springer.com/article/10.1007/s11704-024-40231-1?error=cookies_not_supported&code=25d9584b-c435-4ef2-a472-347691e729b5}
}

@inproceedings{wu2023autogen,
  author        = {Wu, Qingyun and Bansal, Gagan and Zhang, Jieyu and Wu, Yiran and Li, Beibin and Zhu, Erkang and Jiang, Li and Zhang, Xiaoyun and Zhang, Shaokun and Liu, Jiale and Awadallah, Ahmed Hassan and White, Ryen W. and Burger, Doug and Wang, Chi},
  title         = {{AutoGen}: Enabling Next-Gen {LLM} Applications via Multi-Agent Conversation},
  booktitle     = {First Conference on Language Modeling},
  year          = {2024},
  url           = {https://openreview.net/forum?id=BAakY1hNKS}
}

@misc{xi2023rise,
  author        = {Xi, Zhiheng and Chen, Wenxiang and Guo, Xin and He, Wei and Ding, Yiwen and Hong, Boyang and Zhang, Ming and Wang, Junzhe and Jin, Senjie and Zhou, Enyu and Zheng, Rui and Fan, Xiaoran and Wang, Xiao and Xiong, Limao and Zhou, Yuhao and Wang, Weiran and Jiang, Changhao and Zou, Yicheng and Liu, Xiangyang and Yin, Zhangyue and Dou, Shihan and Weng, Rongxiang and Cheng, Wensen and Zhang, Qi and Qin, Wenjuan and Zheng, Yongyan and Qiu, Xipeng and Huang, Xuanjing and Gui, Tao},
  title         = {The Rise and Potential of Large Language Model Based Agents: A Survey},
  year          = {2023},
  doi           = {10.48550/arXiv.2309.07864},
  url           = {https://arxiv.org/abs/2309.07864},
  eprint        = {2309.07864},
  archiveprefix = {arXiv},
  primaryclass  = {cs.AI},
  howpublished  = {arXiv preprint}
}

@misc{xu2024llmkd,
  author        = {Xu, Xiaohan and Li, Ming and Tao, Chongyang and Shen, Tao and Cheng, Reynold and Li, Jinyang and Xu, Can and Tao, Dacheng and Zhou, Tianyi},
  title         = {A Survey on Knowledge Distillation of Large Language Models},
  year          = {2024},
  doi           = {10.48550/arXiv.2402.13116},
  url           = {https://arxiv.org/abs/2402.13116},
  eprint        = {2402.13116},
  archiveprefix = {arXiv},
  howpublished  = {arXiv preprint}
}

@article{xu2025tool,
  author        = {Xu, Weikai and Huang, Chengrui and Gao, Shen and Shang, Shuo},
  title         = {LLM-Based Agents for Tool Learning: A Survey},
  journal       = {Data Science and Engineering},
  volume        = {10},
  number        = {4},
  pages         = {533--563},
  publisher     = {Springer Nature Singapore},
  year          = {2025},
  doi           = {10.1007/s41019-025-00296-9},
  url           = {https://link.springer.com/article/10.1007/s41019-025-00296-9?error=cookies_not_supported&code=1944c8eb-b3d5-47dc-bdfe-61a143a2cdd8}
}

@inproceedings{yang2024sweagent,
  author        = {Yang, John and Jimenez, Carlos E. and Wettig, Alexander and Lieret, Kilian and Yao, Shunyu and Narasimhan, Karthik and Press, Ofir},
  title         = {{SWE-agent}: Agent-Computer Interfaces Enable Automated Software Engineering},
  booktitle     = {Advances in Neural Information Processing Systems},
  volume        = {37},
  pages         = {50528--50652},
  year          = {2024},
  doi           = {10.52202/079017-1601},
  url           = {https://proceedings.neurips.cc/paper_files/paper/2024/hash/5a7c947568c1b1328ccc5230172e1e7c-Abstract-Conference.html}
}

@article{yang2025llmkd,
  author        = {Yang, Chuanpeng and Zhu, Yao and Lu, Wang and Wang, Yidong and Chen, Qian and Gao, Chenlong and Yan, Bingjie and Chen, Yiqiang},
  title         = {Survey on Knowledge Distillation for Large Language Models: Methods, Evaluation, and Application},
  journal       = {ACM Transactions on Intelligent Systems and Technology},
  volume        = {16},
  number        = {6},
  pages         = {143:1--143:27},
  publisher     = {Association for Computing Machinery (ACM)},
  year          = {2025},
  doi           = {10.1145/3699518},
  url           = {https://doi.org/10.1145/3699518}
}

@inproceedings{yao2023react,
  author        = {Yao, Shunyu and Zhao, Jeffrey and Yu, Dian and Du, Nan and Shafran, Izhak and Narasimhan, Karthik and Cao, Yuan},
  title         = {ReAct: Synergizing Reasoning and Acting in Language Models},
  booktitle     = {International Conference on Learning Representations},
  year          = {2023},
  url           = {https://arxiv.org/abs/2210.03629}
}

@misc{yehudai2026survey,
  author        = {Yehudai, Asaf and Eden, Lilach and Li, Alan and Uziel, Guy and Zhao, Yilun and Bar-Haim, Roy and Cohan, Arman and Shmueli-Scheuer, Michal},
  title         = {Survey on Evaluation of {LLM}-based Agents},
  year          = {2025},
  doi           = {10.48550/arXiv.2503.16416},
  url           = {https://arxiv.org/abs/2503.16416},
  eprint        = {2503.16416},
  archiveprefix = {arXiv},
  primaryclass  = {cs.CL},
  howpublished  = {arXiv preprint}
}

@misc{zhang2024xlam,
  author        = {Zhang, Jianguo and Lan, Tian and Zhu, Ming and Liu, Zuxin and Hoang, Thai and Kokane, Shirley and Yao, Weiran and Tan, Juntao and Prabhakar, Akshara and Chen, Haolin and Liu, Zhiwei and Feng, Yihao and Awalgaonkar, Tulika and Murthy, Rithesh and Hu, Eric and Chen, Zeyuan and Xu, Ran and Niebles, Juan Carlos and Heinecke, Shelby and Wang, Huan and Savarese, Silvio and Xiong, Caiming},
  title         = {{xLAM}: A Family of Large Action Models to Empower {AI} Agent Systems},
  year          = {2024},
  doi           = {10.48550/arXiv.2409.03215},
  url           = {https://arxiv.org/abs/2409.03215},
  eprint        = {2409.03215},
  archiveprefix = {arXiv},
  howpublished  = {arXiv preprint}
}

@inproceedings{zhang2025darwin,
  author        = {Zhang, Jenny and Hu, Shengran and Lu, Cong and Lange, Robert and Clune, Jeff},
  title         = {Darwin {G{\"o}del} Machine: Open-Ended Evolution of Self-Improving Agents},
  booktitle     = {International Conference on Learning Representations},
  year          = {2026},
  url           = {https://arxiv.org/abs/2505.22954}
}

@article{zhang2025memory,
  author        = {Zhang, Zeyu and Dai, Quanyu and Bo, Xiaohe and Ma, Chen and Li, Rui and Chen, Xu and Zhu, Jieming and Dong, Zhenhua and Wen, Ji-Rong},
  title         = {A Survey on the Memory Mechanism of Large Language Model-based Agents},
  journal       = {ACM Transactions on Information Systems},
  volume        = {43},
  number        = {6},
  pages         = {155:1--155:47},
  publisher     = {Association for Computing Machinery (ACM)},
  year          = {2025},
  doi           = {10.1145/3748302},
  url           = {https://doi.org/10.1145/3748302}
}

@inproceedings{zhao2024expel,
  author        = {Zhao, Andrew and Huang, Daniel and Xu, Quentin and Lin, Matthieu and Liu, Yong-Jin and Huang, Gao},
  title         = {{ExpeL}: {LLM} Agents Are Experiential Learners},
  booktitle     = {Proceedings of the AAAI Conference on Artificial Intelligence},
  volume        = {38},
  pages         = {19632--19642},
  publisher     = {Association for the Advancement of Artificial Intelligence (AAAI)},
  year          = {2024},
  doi           = {10.1609/aaai.v38i17.29936},
  url           = {https://doi.org/10.1609/aaai.v38i17.29936}
}

@inproceedings{zhong2023memorybank,
  author        = {Zhong, Wanjun and Guo, Lianghong and Gao, Qiqi and Ye, He and Wang, Yanlin},
  title         = {MemoryBank: Enhancing Large Language Models with Long-Term Memory},
  booktitle     = {Proceedings of the AAAI Conference on Artificial Intelligence},
  volume        = {38},
  pages         = {19724--19731},
  publisher     = {AAAI Press},
  year          = {2024},
  doi           = {10.1609/aaai.v38i17.29946},
  url           = {https://ojs.aaai.org/index.php/AAAI/article/view/29946}
}

@inproceedings{zhou2024webarena,
  author        = {Zhou, Shuyan and Xu, Frank F. and Zhu, Hao and Zhou, Xuhui and Lo, Robert and Sridhar, Abishek and Cheng, Xianyi and Ou, Tianyue and Bisk, Yonatan and Fried, Daniel and Alon, Uri and Neubig, Graham},
  title         = {WebArena: A Realistic Web Environment for Building Autonomous Agents},
  booktitle     = {International Conference on Learning Representations},
  year          = {2024},
  url           = {https://openreview.net/forum?id=oKn9c6ytLx}
}

@inproceedings{zhuge2024gptswarm,
  author        = {Zhuge, Mingchen and Wang, Wenyi and Kirsch, Louis and Faccio, Francesco and Khizbullin, Dmitrii and Schmidhuber, J{\"u}rgen},
  title         = {{GPTSwarm}: Language Agents as Optimizable Graphs},
  booktitle     = {Proceedings of the 41st International Conference on Machine Learning},
  series        = {Proceedings of Machine Learning Research},
  volume        = {235},
  pages         = {62743--62767},
  publisher     = {PMLR},
  year          = {2024},
  url           = {https://proceedings.mlr.press/v235/zhuge24a.html}
}

@article{boizard2025uld,
  author        = {Nicolas Boizard and Kevin El Haddad and C{\'e}line Hudelot and Pierre Colombo},
  title         = {Towards Cross-Tokenizer Distillation: The Universal Logit Distillation Loss for {LLM}s},
  journal       = {Transactions on Machine Learning Research},
  year          = {2025},
  url           = {https://arxiv.org/abs/2402.12030},
  eprint        = {2402.12030},
  archiveprefix = {arXiv},
  primaryclass  = {cs.CL},
  issn          = {2835-8856}
}

@inproceedings{chen-etal-2024-agentflan,
  author        = {Zehui Chen and Kuikun Liu and Qiuchen Wang and Wenwei Zhang and Jiangning Liu and Dahua Lin and Kai Chen and Feng Zhao},
  title         = {Agent-{FLAN}: Designing Data and Methods of Effective Agent Tuning for Large Language Models},
  booktitle     = {Findings of the Association for Computational Linguistics: ACL 2024},
  pages         = {9354--9366},
  publisher     = {Association for Computational Linguistics},
  address       = {Bangkok, Thailand},
  year          = {2024},
  doi           = {10.18653/v1/2024.findings-acl.557},
  url           = {https://aclanthology.org/2024.findings-acl.557/}
}

@inproceedings{chen-etal-2025-atlas,
  author        = {Zhixun Chen and Ming Li and Yuxuan Huang and Yali Du and Meng Fang and Tianyi Zhou},
  title         = {{ATLAS}: Agent Tuning via Learning Critical Steps},
  booktitle     = {Findings of the Association for Computational Linguistics: ACL 2025},
  pages         = {25334--25349},
  publisher     = {Association for Computational Linguistics},
  address       = {Vienna, Austria},
  year          = {2025},
  doi           = {10.18653/v1/2025.findings-acl.1299},
  url           = {https://aclanthology.org/2025.findings-acl.1299/},
  isbn          = {979-8-89176-256-5}
}

@misc{chen2023fireact,
  author        = {Baian Chen and Chang Shu and Ehsan Shareghi and Nigel Collier and Karthik Narasimhan and Shunyu Yao},
  title         = {{FireAct}: Toward Language Agent Fine-Tuning},
  year          = {2023},
  doi           = {10.48550/arXiv.2310.05915},
  url           = {https://arxiv.org/abs/2310.05915},
  eprint        = {2310.05915},
  archiveprefix = {arXiv},
  primaryclass  = {cs.CL},
  howpublished  = {arXiv preprint}
}

@misc{chen2026futurebridge,
  author        = {Chishui Chen and Yaoyou Fan and Te Sun and Yi Yang and Chenghao Sun and Delin Mao and Hongbo Qiao and Zuowei Zhang and Junxi Wang and Chenxing Sun and Yangen Hu and Lu Pan and Xuyang Liu and Linfeng Zhang},
  title         = {Look Ahead Before You Distill: Future Trajectory Validation of Teacher Guidance for Agentic On-Policy Distillation},
  year          = {2026},
  doi           = {10.48550/arXiv.2608.01953},
  url           = {https://arxiv.org/abs/2608.01953},
  eprint        = {2608.01953},
  archiveprefix = {arXiv},
  primaryclass  = {cs.CL},
  howpublished  = {arXiv preprint}
}

@inproceedings{driess2023palme,
  author        = {Driess, Danny and Xia, Fei and Sajjadi, Mehdi S. M. and Lynch, Corey and Chowdhery, Aakanksha and Ichter, Brian and Wahid, Ayzaan and Tompson, Jonathan and Vuong, Quan and Yu, Tianhe and Huang, Wenlong and Chebotar, Yevgen and Sermanet, Pierre and Duckworth, Daniel and Levine, Sergey and Vanhoucke, Vincent and Hausman, Karol and Toussaint, Marc and Greff, Klaus and Zeng, Andy and Mordatch, Igor and Florence, Pete},
  title         = {{PaLM-E}: An Embodied Multimodal Language Model},
  booktitle     = {Proceedings of the 40th International Conference on Machine Learning},
  series        = {Proceedings of Machine Learning Research},
  volume        = {202},
  pages         = {8469--8488},
  publisher     = {PMLR},
  year          = {2023},
  url           = {https://proceedings.mlr.press/v202/driess23a.html}
}

@inproceedings{fu2025agentrefine,
  author        = {Dayuan Fu and Keqing He and Yejie Wang and Wentao Hong and Zhuoma GongQue and Weihao Zeng and Wei Wang and Jingang Wang and Xunliang Cai and Weiran Xu},
  title         = {{AgentRefine}: Enhancing Agent Generalization through Refinement Tuning},
  booktitle     = {International Conference on Learning Representations},
  volume        = {2025},
  pages         = {65185--65204},
  year          = {2025},
  url           = {https://proceedings.iclr.cc/paper_files/paper/2025/hash/a3cc50126338b175e56bb3cad134db0b-Abstract-Conference.html}
}

@inproceedings{hong2024cogagent,
  author        = {Hong, Wenyi and Wang, Weihan and Lv, Qingsong and Xu, Jiazheng and Yu, Wenmeng and Ji, Junhui and Wang, Yan and Wang, Zihan and Dong, Yuxiao and Ding, Ming and Tang, Jie},
  title         = {CogAgent: A Visual Language Model for GUI Agents},
  booktitle     = {Proceedings of the IEEE/CVF Conference on Computer Vision and Pattern Recognition},
  pages         = {14281--14290},
  year          = {2024},
  url           = {https://openaccess.thecvf.com/content/CVPR2024/html/Hong_CogAgent_A_Visual_Language_Model_for_GUI_Agents_CVPR_2024_paper.html}
}

@inproceedings{hsieh-etal-2023-distilling,
  author        = {Cheng-Yu Hsieh and Chun-Liang Li and Chih-kuan Yeh and Hootan Nakhost and Yasuhisa Fujii and Alex Ratner and Ranjay Krishna and Chen-Yu Lee and Tomas Pfister},
  title         = {Distilling Step-by-Step! Outperforming Larger Language Models with Less Training Data and Smaller Model Sizes},
  booktitle     = {Findings of the Association for Computational Linguistics: ACL 2023},
  pages         = {8003--8017},
  publisher     = {Association for Computational Linguistics},
  address       = {Toronto, Canada},
  year          = {2023},
  month         = {jul},
  doi           = {10.18653/v1/2023.findings-acl.507},
  url           = {https://aclanthology.org/2023.findings-acl.507/}
}

@inproceedings{jiang2025curriculum,
  author        = {Wangyi Jiang and Yaojie Lu and Hongyu Lin and Xianpei Han and Le Sun},
  title         = {Teach Small Models to Reason by Curriculum Distillation},
  booktitle     = {Proceedings of the 2025 Conference on Empirical Methods in Natural Language Processing},
  pages         = {7412--7422},
  publisher     = {Association for Computational Linguistics},
  address       = {Suzhou, China},
  year          = {2025},
  doi           = {10.18653/v1/2025.emnlp-main.376},
  url           = {https://aclanthology.org/2025.emnlp-main.376/},
  isbn          = {979-8-89176-332-6}
}

@inproceedings{kang2025distillingllmagent,
  author        = {Minki Kang and Jongwon Jeong and Seanie Lee and Jaewoong Cho and Sung Ju Hwang},
  title         = {Distilling {LLM} Agent into Small Models with Retrieval and Code Tools},
  booktitle     = {Advances in Neural Information Processing Systems},
  volume        = {38},
  pages         = {106501--106538},
  publisher     = {Curran Associates, Inc.},
  year          = {2025},
  doi           = {10.52202/085713-3553},
  url           = {https://proceedings.neurips.cc/paper_files/paper/2025/hash/99263f9bf46874e2b8b7f104b5063864-Abstract-Conference.html}
}

@misc{kim2024openvla,
  author        = {Kim, Moo Jin and Pertsch, Karl and Karamcheti, Siddharth and Xiao, Ted and Balakrishna, Ashwin and Nair, Suraj and Rafailov, Rafael and Foster, Ethan and Lam, Grace and Sanketi, Pannag and Vuong, Quan and Kollar, Thomas and Burchfiel, Benjamin and Tedrake, Russ and Sadigh, Dorsa and Levine, Sergey and Liang, Percy and Finn, Chelsea},
  title         = {{OpenVLA}: An Open-Source Vision-Language-Action Model},
  year          = {2024},
  doi           = {10.48550/arXiv.2406.09246},
  url           = {https://arxiv.org/abs/2406.09246},
  eprint        = {2406.09246},
  archiveprefix = {arXiv},
  howpublished  = {arXiv preprint}
}

@misc{lauffer2025oec,
  author        = {Niklas Lauffer and Xiang Deng and Srivatsa Kundurthy and Brad Kenstler and Jeff Da},
  title         = {Imitation Learning for Multi-Turn {LM} Agents via On-Policy Expert Corrections},
  year          = {2025},
  doi           = {10.48550/arXiv.2512.14895},
  url           = {https://arxiv.org/abs/2512.14895},
  eprint        = {2512.14895},
  archiveprefix = {arXiv},
  primaryclass  = {cs.LG},
  howpublished  = {arXiv preprint}
}

@inproceedings{li2024mixed,
  author        = {Chenglin Li and Qianglong Chen and Liangyue Li and Caiyu Wang and Feng Tao and Yicheng Li and Zulong Chen and Yin Zhang},
  title         = {Mixed Distillation Helps Smaller Language Models Reason Better},
  booktitle     = {Findings of the Association for Computational Linguistics: EMNLP 2024},
  pages         = {1673--1690},
  publisher     = {Association for Computational Linguistics},
  address       = {Miami, Florida, USA},
  year          = {2024},
  doi           = {10.18653/v1/2024.findings-emnlp.91},
  url           = {https://aclanthology.org/2024.findings-emnlp.91/}
}

@inproceedings{li2025learnability,
  author        = {Yuetai Li and Xiang Yue and Zhangchen Xu and Fengqing Jiang and Luyao Niu and Bill Yuchen Lin and Bhaskar Ramasubramanian and Radha Poovendran},
  title         = {Small Models Struggle to Learn from Strong Reasoners},
  booktitle     = {Findings of the Association for Computational Linguistics: ACL 2025},
  pages         = {25366--25394},
  publisher     = {Association for Computational Linguistics},
  address       = {Vienna, Austria},
  year          = {2025},
  doi           = {10.18653/v1/2025.findings-acl.1301},
  url           = {https://aclanthology.org/2025.findings-acl.1301/},
  isbn          = {979-8-89176-256-5}
}

@misc{li2026revisitingdagger,
  author        = {Changhao Li and Rushi Qiang and Jiawei Huang and Chenxiao Gao and Chao Zhang and Niao He and Bo Dai},
  title         = {Revisiting {DAgger} in the Era of {LLM}-Agents},
  year          = {2026},
  doi           = {10.48550/arXiv.2605.12913},
  url           = {https://arxiv.org/abs/2605.12913},
  eprint        = {2605.12913},
  archiveprefix = {arXiv},
  primaryclass  = {cs.LG},
  howpublished  = {arXiv preprint}
}

@misc{liao2026reopd,
  author        = {Baohao Liao and Hanze Dong and Christof Monz and Xinxing Xu and Li Dong and Furu Wei},
  title         = {Multi-Turn On-Policy Distillation with Prefix Replay},
  year          = {2026},
  doi           = {10.48550/arXiv.2607.04763},
  url           = {https://arxiv.org/abs/2607.04763},
  eprint        = {2607.04763},
  archiveprefix = {arXiv},
  primaryclass  = {cs.LG},
  howpublished  = {arXiv preprint}
}

@misc{liu2026hero,
  author        = {Haoran Liu and Yuwei Zhang and Xiyao Li and Bohan Lyu and Jingbo Shang},
  title         = {{HERO}: Hindsight-Enhanced Reflection from Environment Observations for Agentic Self-Distillation},
  year          = {2026},
  doi           = {10.48550/arXiv.2606.11559},
  url           = {https://arxiv.org/abs/2606.11559},
  eprint        = {2606.11559},
  archiveprefix = {arXiv},
  primaryclass  = {cs.AI},
  howpublished  = {arXiv preprint}
}

@inproceedings{liu2026prefixalign,
  author        = {Zhenghao Liu and Zhuoyang Wu and Xinze Li and Yukun Yan and Shuo Wang and Zulong Chen and Yu Gu and Ge Yu and Maosong Sun},
  title         = {Long-Chain Reasoning Distillation via Adaptive Prefix Alignment},
  booktitle     = {Proceedings of the 64th Annual Meeting of the {A}ssociation for {C}omputational {L}inguistics (Volume 1: Long Papers)},
  pages         = {18041--18059},
  publisher     = {Association for Computational Linguistics},
  address       = {San Diego, California, United States},
  year          = {2026},
  doi           = {10.18653/v1/2026.acl-long.822},
  url           = {https://aclanthology.org/2026.acl-long.822/},
  isbn          = {979-8-89176-390-6}
}

@inproceedings{liu2026structuredagentdistillation,
  author        = {Jun Liu and Zhenglun Kong and Peiyan Dong and Changdi Yang and Tianqi Li and Yanyue Xie and Yanfan Gong and Xuan Shen and Hao Tang and Pu Zhao and Geng Yuan and Wei Niu and Wenbin Zhang and Xue Lin and Dong Huang and Yanzhi Wang},
  title         = {Structured Agent Distillation for Large Language Model Agents},
  booktitle     = {Proceedings of the 25th International Conference on Autonomous Agents and Multiagent Systems},
  pages         = {3676--3685},
  publisher     = {International Foundation for Autonomous Agents and Multiagent Systems},
  address       = {Paphos, Cyprus},
  year          = {2026},
  doi           = {10.65109/OLHJ8062},
  url           = {https://www.ifaamas.org/Proceedings/aamas2026/pdfs/OLHJ8062.pdf}
}

@misc{lu2026sdar,
  author        = {Zhengxi Lu and Zhiyuan Yao and Zhuowen Han and Zi-Han Wang and Jinyang Wu and Qi Gu and Xunliang Cai and Weiming Lu and Jun Xiao and Yueting Zhuang and Yongliang Shen},
  title         = {Self-Distilled Agentic Reinforcement Learning},
  year          = {2026},
  doi           = {10.48550/arXiv.2605.15155},
  url           = {https://arxiv.org/abs/2605.15155},
  eprint        = {2605.15155},
  archiveprefix = {arXiv},
  primaryclass  = {cs.LG},
  howpublished  = {arXiv preprint}
}

@misc{lyu2026score,
  author        = {Yuanjie Lyu and Chengyu Wang and Jun Huang and Tong Xu},
  title         = {Student-Centered Distillation Narrows the Agentic Gap Between Small and Large {LLM}s},
  year          = {2026},
  url           = {https://openreview.net/forum?id=KaTYG9LGJv},
  eprint        = {2509.14257},
  archiveprefix = {arXiv},
  primaryclass  = {cs.CL},
  note          = {Accepted to ICML 2026}
}

@inproceedings{magister2023teaching,
  author        = {Lucie Charlotte Magister and Jonathan Mallinson and Jakub Adamek and Eric Malmi and Aliaksei Severyn},
  title         = {Teaching Small Language Models to Reason},
  booktitle     = {Proceedings of the 61st Annual Meeting of the Association for Computational Linguistics (Volume 2: Short Papers)},
  pages         = {1773--1781},
  publisher     = {Association for Computational Linguistics},
  address       = {Toronto, Canada},
  year          = {2023},
  doi           = {10.18653/v1/2023.acl-short.151},
  url           = {https://aclanthology.org/2023.acl-short.151/}
}

@inproceedings{qin2024toolllm,
  author        = {Yujia Qin and Shihao Liang and Yining Ye and Kunlun Zhu and Lan Yan and Yaxi Lu and Yankai Lin and Xin Cong and Xiangru Tang and Bill Qian and Sihan Zhao and Lauren Hong and Runchu Tian and Ruobing Xie and Jie Zhou and Mark Gerstein and Dahai Li and Zhiyuan Liu and Maosong Sun},
  title         = {{ToolLLM}: Facilitating Large Language Models to Master 16000+ Real-World {APIs}},
  booktitle     = {International Conference on Learning Representations},
  volume        = {2024},
  pages         = {9695--9717},
  year          = {2024},
  url           = {https://proceedings.iclr.cc/paper_files/paper/2024/hash/28e50ee5b72e90b50e7196fde8ea260e-Abstract-Conference.html}
}

@inproceedings{shridhar2023distilling,
  author        = {Kumar Shridhar and Alessandro Stolfo and Mrinmaya Sachan},
  title         = {Distilling Reasoning Capabilities into Smaller Language Models},
  booktitle     = {Findings of the Association for Computational Linguistics: ACL 2023},
  pages         = {7059--7073},
  publisher     = {Association for Computational Linguistics},
  address       = {Toronto, Canada},
  year          = {2023},
  doi           = {10.18653/v1/2023.findings-acl.441},
  url           = {https://aclanthology.org/2023.findings-acl.441/}
}

@inproceedings{song-etal-2024-agentbank,
  author        = {Yifan Song and Weimin Xiong and Xiutian Zhao and Dawei Zhu and Wenhao Wu and Ke Wang and Cheng Li and Wei Peng and Sujian Li},
  title         = {{A}gent{B}ank: Towards Generalized {LLM} Agents via Fine-Tuning on 50000+ Interaction Trajectories},
  booktitle     = {Findings of the Association for Computational Linguistics: EMNLP 2024},
  pages         = {2124--2141},
  publisher     = {Association for Computational Linguistics},
  address       = {Miami, Florida, USA},
  year          = {2024},
  doi           = {10.18653/v1/2024.findings-emnlp.116},
  url           = {https://aclanthology.org/2024.findings-emnlp.116/}
}

@inproceedings{song-etal-2024-trial,
  author        = {Yifan Song and Da Yin and Xiang Yue and Jie Huang and Sujian Li and Bill Yuchen Lin},
  title         = {Trial and Error: Exploration-Based Trajectory Optimization of {LLM} Agents},
  booktitle     = {Proceedings of the 62nd Annual Meeting of the Association for Computational Linguistics (Volume 1: Long Papers)},
  pages         = {7584--7600},
  publisher     = {Association for Computational Linguistics},
  address       = {Bangkok, Thailand},
  year          = {2024},
  doi           = {10.18653/v1/2024.acl-long.409},
  url           = {https://aclanthology.org/2024.acl-long.409/}
}

@misc{sun2023instruction,
  author        = {Weiwei Sun and Zheng Chen and Xinyu Ma and Lingyong Yan and Shuaiqiang Wang and Pengjie Ren and Zhumin Chen and Dawei Yin and Zhaochun Ren},
  title         = {Instruction Distillation Makes Large Language Models Efficient Zero-Shot Rankers},
  year          = {2023},
  doi           = {10.48550/arXiv.2311.01555},
  url           = {https://arxiv.org/abs/2311.01555},
  eprint        = {2311.01555},
  archiveprefix = {arXiv},
  primaryclass  = {cs.IR},
  howpublished  = {arXiv preprint}
}

@misc{tan2026atod,
  author        = {Qitai Tan and Zefang Zong and Mo Li and Yipeng Shi and Yang Li and Peng Chen},
  title         = {{ATOD}: Annealed Turn-Aware On-Policy Distillation for Multi-Turn Agentic Tasks},
  year          = {2026},
  doi           = {10.48550/arXiv.2606.27814},
  url           = {https://arxiv.org/abs/2606.27814},
  eprint        = {2606.27814},
  archiveprefix = {arXiv},
  primaryclass  = {cs.AI},
  howpublished  = {arXiv preprint}
}

@inproceedings{tang-zhao-2026-smartad,
  author        = {Guokai Tang and Feng Zhao},
  title         = {{S}mart{AD}: Capacity-Aligned Agent Distillation for Small Language Models},
  booktitle     = {Findings of the {A}ssociation for {C}omputational {L}inguistics: {ACL} 2026},
  pages         = {27045--27057},
  publisher     = {Association for Computational Linguistics},
  address       = {San Diego, California, United States},
  year          = {2026},
  doi           = {10.18653/v1/2026.findings-acl.1349},
  url           = {https://aclanthology.org/2026.findings-acl.1349/},
  isbn          = {979-8-89176-395-1}
}

@inproceedings{wadhwa2024mysteries,
  author        = {Somin Wadhwa and Silvio Amir and Byron C. Wallace},
  title         = {Investigating Mysteries of {CoT}-Augmented Distillation},
  booktitle     = {Proceedings of the 2024 Conference on Empirical Methods in Natural Language Processing},
  pages         = {6071--6086},
  publisher     = {Association for Computational Linguistics},
  address       = {Miami, Florida, USA},
  year          = {2024},
  doi           = {10.18653/v1/2024.emnlp-main.349},
  url           = {https://aclanthology.org/2024.emnlp-main.349/}
}

@inproceedings{wang2023scott,
  author        = {Peifeng Wang and Zhengyang Wang and Zheng Li and Yifan Gao and Bing Yin and Xiang Ren},
  title         = {{SCOTT}: Self-Consistent Chain-of-Thought Distillation},
  booktitle     = {Proceedings of the 61st Annual Meeting of the Association for Computational Linguistics (Volume 1: Long Papers)},
  pages         = {5546--5558},
  publisher     = {Association for Computational Linguistics},
  address       = {Toronto, Canada},
  year          = {2023},
  doi           = {10.18653/v1/2023.acl-long.304},
  url           = {https://aclanthology.org/2023.acl-long.304/}
}

@misc{wang2026agentopsd,
  author        = {Zi-Han Wang and Zhengxi Lu and Zhiyuan Yao and Jinyang Wu and Jie Wu and Zhengzhou Cai and Yueqing Sun and Ziang Ye and Linji Hao and Qi Gu and Xunliang Cai and Yongliang Shen and Yujiu Yang},
  title         = {{AgentOPSD}: Recursive Self-Distillation for Agentic Reinforcement Learning},
  year          = {2026},
  doi           = {10.48550/arXiv.2608.05987},
  url           = {https://arxiv.org/abs/2608.05987},
  eprint        = {2608.05987},
  archiveprefix = {arXiv},
  primaryclass  = {cs.AI},
  howpublished  = {arXiv preprint}
}

@misc{wang2026tcod,
  author        = {Jiaqi Wang and Wenhao Zhang and Weijie Shi and Yaliang Li and James Cheng},
  title         = {{TCOD}: Exploring Temporal Curriculum in On-Policy Distillation for Multi-Turn Autonomous Agents},
  year          = {2026},
  doi           = {10.48550/arXiv.2604.24005},
  url           = {https://arxiv.org/abs/2604.24005},
  eprint        = {2604.24005},
  archiveprefix = {arXiv},
  primaryclass  = {cs.LG},
  howpublished  = {arXiv preprint}
}

@inproceedings{xiong-etal-2024-watch,
  author        = {Weimin Xiong and Yifan Song and Xiutian Zhao and Wenhao Wu and Xun Wang and Ke Wang and Cheng Li and Wei Peng and Sujian Li},
  title         = {Watch Every Step! {LLM} Agent Learning via Iterative Step-Level Process Refinement},
  booktitle     = {Proceedings of the 2024 Conference on Empirical Methods in Natural Language Processing},
  pages         = {1556--1572},
  publisher     = {Association for Computational Linguistics},
  address       = {Miami, Florida, USA},
  year          = {2024},
  doi           = {10.18653/v1/2024.emnlp-main.93},
  url           = {https://aclanthology.org/2024.emnlp-main.93/}
}

@misc{yang2026ocsd,
  author        = {Yi Yang and Cong Qin and Xiaodan Liu and Chishui Chen and Qing Dong and Yan Zhang and Cao Liu and Zhao Yang and Lu Pan and Jiaye Lin and Yi Feng},
  title         = {Agentic Reinforcement Learning with Observation-Calibrated Self-Distillation},
  year          = {2026},
  doi           = {10.48550/arXiv.2608.04788},
  url           = {https://arxiv.org/abs/2608.04788},
  eprint        = {2608.04788},
  archiveprefix = {arXiv},
  primaryclass  = {cs.LG},
  howpublished  = {arXiv preprint}
}

@misc{yeo2026hintsd,
  author        = {Woongyeong Yeo and Yumin Choi and Taekyung Ki and Sung Ju Hwang},
  title         = {{HINT-SD}: Targeted Hindsight Self-Distillation for Long-Horizon Agents},
  year          = {2026},
  doi           = {10.48550/arXiv.2605.17873},
  url           = {https://arxiv.org/abs/2605.17873},
  eprint        = {2605.17873},
  archiveprefix = {arXiv},
  primaryclass  = {cs.LG},
  howpublished  = {arXiv preprint},
  note          = {Accepted to Findings of EMNLP 2026}
}

@inproceedings{zeng-etal-2024-agenttuning,
  author        = {Aohan Zeng and Mingdao Liu and Rui Lu and Bowen Wang and Xiao Liu and Yuxiao Dong and Jie Tang},
  title         = {{A}gent{T}uning: Enabling Generalized Agent Abilities for {LLM}s},
  booktitle     = {Findings of the Association for Computational Linguistics: ACL 2024},
  pages         = {3053--3077},
  publisher     = {Association for Computational Linguistics},
  address       = {Bangkok, Thailand},
  year          = {2024},
  doi           = {10.18653/v1/2024.findings-acl.181},
  url           = {https://aclanthology.org/2024.findings-acl.181/}
}

@misc{zhang2026stepopsd,
  author        = {Yanfei Zhang and Xu Lin and Chenglin Wu},
  title         = {{StepOPSD}: Step-Aware Online Preference Self-Distillation for Agent Reinforcement Learning},
  year          = {2026},
  doi           = {10.48550/arXiv.2605.27140},
  url           = {https://arxiv.org/abs/2605.27140},
  eprint        = {2605.27140},
  archiveprefix = {arXiv},
  primaryclass  = {cs.AI},
  howpublished  = {arXiv preprint},
  note          = {Accepted to Findings of EMNLP 2026}
}

@misc{zhao2026opsd,
  author        = {Siyan Zhao and Zhihui Xie and Mengchen Liu and Jing Huang and Guan Pang and Feiyu Chen and Aditya Grover},
  title         = {Self-Distilled Reasoner: On-Policy Self-Distillation for Large Language Models},
  year          = {2026},
  doi           = {10.48550/arXiv.2601.18734},
  url           = {https://arxiv.org/abs/2601.18734},
  eprint        = {2601.18734},
  archiveprefix = {arXiv},
  primaryclass  = {cs.CL},
  howpublished  = {arXiv preprint}
}

@misc{zhong2026sod,
  author        = {Qiyong Zhong and Mao Zheng and Mingyang Song and Xin Lin and Jie Sun and Houcheng Jiang and Xiang Wang and Junfeng Fang},
  title         = {{SOD}: Step-Wise On-Policy Distillation for Small Language Model Agents},
  year          = {2026},
  doi           = {10.48550/arXiv.2605.07725},
  url           = {https://arxiv.org/abs/2605.07725},
  eprint        = {2605.07725},
  archiveprefix = {arXiv},
  primaryclass  = {cs.CL},
  howpublished  = {arXiv preprint}
}

@misc{zhou2026turnopd,
  author        = {Yuhang Zhou and Kai Zheng and Haoling Li and Dengyun Peng and Can Xu and Jingjing Chen},
  title         = {{TurnOPD}: Making On-Policy Distillation Turn-Aware for Efficient Long-Horizon Agent Training},
  year          = {2026},
  doi           = {10.48550/arXiv.2607.05804},
  url           = {https://arxiv.org/abs/2607.05804},
  eprint        = {2607.05804},
  archiveprefix = {arXiv},
  primaryclass  = {cs.AI},
  howpublished  = {arXiv preprint}
}

@inproceedings{zitkovich2023rt2,
  author        = {Zitkovich, Brianna and Yu, Tianhe and Xu, Sichun and Xu, Peng and Xiao, Ted and Xia, Fei and Wu, Jialin and Wohlhart, Paul and Welker, Stefan and Wahid, Ayzaan and Vuong, Quan and Vanhoucke, Vincent and Tran, Huong and Soricut, Radu and Singh, Anikait and Singh, Jaspiar and Sermanet, Pierre and Sanketi, Pannag R. and Salazar, Grecia and Ryoo, Michael S. and Reymann, Krista and Rao, Kanishka and Pertsch, Karl and Mordatch, Igor and Michalewski, Henryk and Lu, Yao and Levine, Sergey and Lee, Lisa and Lee, Tsang-Wei Edward and Leal, Isabel and Kuang, Yuheng and Kalashnikov, Dmitry and Julian, Ryan and Joshi, Nikhil J. and Irpan, Alex and Ichter, Brian and Hsu, Jasmine and Herzog, Alexander and Hausman, Karol and Gopalakrishnan, Keerthana and Fu, Chuyuan and Florence, Pete and Finn, Chelsea and Dubey, Kumar Avinava and Driess, Danny and Ding, Tianli and Choromanski, Krzysztof Marcin and Chen, Xi and Chebotar, Yevgen and Carbajal, Justice and Brown, Noah and Brohan, Anthony and Arenas, Montserrat Gonzalez and Han, Kehang},
  title         = {{RT-2}: Vision-Language-Action Models Transfer Web Knowledge to Robotic Control},
  booktitle     = {Proceedings of The 7th Conference on Robot Learning},
  series        = {Proceedings of Machine Learning Research},
  volume        = {229},
  pages         = {2165--2183},
  publisher     = {PMLR},
  year          = {2023},
  url           = {https://proceedings.mlr.press/v229/zitkovich23a.html}
}

@misc{cai2023latm,
  author        = {Cai, Tianle and Wang, Xuezhi and Ma, Tengyu and Chen, Xinyun and Zhou, Denny},
  title         = {Large Language Models as Tool Makers},
  year          = {2023},
  doi           = {10.48550/arXiv.2305.17126},
  url           = {https://arxiv.org/abs/2305.17126},
  eprint        = {2305.17126},
  archiveprefix = {arXiv},
  primaryclass  = {cs.CL},
  howpublished  = {arXiv preprint}
}

@misc{fu2024autoguide,
  author        = {Fu, Yao and Kim, Dong-Ki and Kim, Jaekyeom and Sohn, Sungryull and Logeswaran, Lajanugen and Bae, Kyunghoon and Lee, Honglak},
  title         = {{AutoGuide}: Automated Generation and Selection of Context-Aware Guidelines for Large Language Model Agents},
  year          = {2024},
  doi           = {10.48550/arXiv.2403.08978},
  url           = {https://arxiv.org/abs/2403.08978},
  eprint        = {2403.08978},
  archiveprefix = {arXiv},
  primaryclass  = {cs.AI},
  howpublished  = {arXiv preprint}
}

@misc{kim2026agentmemory,
  author        = {Kim, Taeil and Kim, Kangsan and Hwang, Sung Ju},
  title         = {Agent Memory Distillation: Empowering Small {LLM} Agents with Hierarchical Teacher Memory},
  year          = {2026},
  doi           = {10.48550/arXiv.2608.07169},
  url           = {https://arxiv.org/abs/2608.07169},
  eprint        = {2608.07169},
  archiveprefix = {arXiv},
  primaryclass  = {cs.AI},
  howpublished  = {arXiv preprint}
}

@inproceedings{qian2023creator,
  author        = {Qian, Cheng and Han, Chi and Fung, Yi and Qin, Yujia and Liu, Zhiyuan and Ji, Heng},
  title         = {{CREATOR}: Tool Creation for Disentangling Abstract and Concrete Reasoning of Large Language Models},
  booktitle     = {Findings of the Association for Computational Linguistics: EMNLP 2023},
  pages         = {6922--6939},
  publisher     = {Association for Computational Linguistics},
  address       = {Singapore},
  year          = {2023},
  doi           = {10.18653/v1/2023.findings-emnlp.462},
  url           = {https://aclanthology.org/2023.findings-emnlp.462/}
}

@misc{qiu2025agentdistill,
  author        = {Qiu, Jiahao and Juan, Xinzhe and Wang, Yimin and Yang, Ling and Qi, Xuan and Zhang, Tongcheng and Guo, Jiacheng and Lu, Yifu and Yao, Zixin and Wang, Hongru and Liu, Shilong and Jiang, Xun and Liu, Leqi and Wang, Mengdi},
  title         = {{AgentDistill}: Training-Free Agent Distillation with Generalizable {MCP} Boxes},
  year          = {2025},
  doi           = {10.48550/arXiv.2506.14728},
  url           = {https://arxiv.org/abs/2506.14728},
  eprint        = {2506.14728},
  archiveprefix = {arXiv},
  primaryclass  = {cs.AI},
  howpublished  = {arXiv preprint}
}

@misc{shi2026skillkd,
  author        = {Shi, Qiming and Dou, Yibo and Zhu, Jiawen and Tao, Yulong and Jin, Linbo and Kang, Zhaolu and Zhou, Yunfan and Weng, Di},
  title         = {{SKILL-KD}: Contrastive Skill Distillation for {LLM} Agents},
  year          = {2026},
  doi           = {10.48550/arXiv.2607.28048},
  url           = {https://arxiv.org/abs/2607.28048},
  eprint        = {2607.28048},
  archiveprefix = {arXiv},
  primaryclass  = {cs.AI},
  howpublished  = {arXiv preprint}
}

@misc{shi2026skillone,
  author        = {Shi, Yaorui and Chen, Yuxin and Lu, Zhengxi and Miao, Yuchun and Liu, Shugui and Gu, Qi and Cai, Xunliang and Wang, Xiang and Zhang, An},
  title         = {{Skill1}: Unified Evolution of Skill-Augmented Agents via Reinforcement Learning},
  year          = {2026},
  doi           = {10.48550/arXiv.2605.06130},
  url           = {https://arxiv.org/abs/2605.06130},
  eprint        = {2605.06130},
  archiveprefix = {arXiv},
  primaryclass  = {cs.AI},
  howpublished  = {arXiv preprint}
}

@inproceedings{tziafas2024lrll,
  author        = {Tziafas, Georgios and Kasaei, Hamidreza},
  title         = {Lifelong Robot Library Learning: Bootstrapping Composable and Generalizable Skills for Embodied Control with Language Models},
  booktitle     = {2024 IEEE International Conference on Robotics and Automation (ICRA)},
  year          = {2024},
  url           = {https://arxiv.org/abs/2406.18746}
}

@misc{wang2024awm,
  author        = {Wang, Zora Zhiruo and Mao, Jiayuan and Fried, Daniel and Neubig, Graham},
  title         = {Agent Workflow Memory},
  year          = {2024},
  doi           = {10.48550/arXiv.2409.07429},
  url           = {https://arxiv.org/abs/2409.07429},
  eprint        = {2409.07429},
  archiveprefix = {arXiv},
  primaryclass  = {cs.CL},
  howpublished  = {arXiv preprint}
}

@misc{xia2026skillrl,
  author        = {Xia, Peng and Chen, Jianwen and Wang, Hanyang and Liu, Jiaqi and Zeng, Kaide and Wang, Yu and Han, Siwei and Zhou, Yiyang and Zhao, Xujiang and Chen, Haifeng and Zheng, Zeyu and Xie, Cihang and Yao, Huaxiu},
  title         = {{SkillRL}: Evolving Agents via Recursive Skill-Augmented Reinforcement Learning},
  year          = {2026},
  doi           = {10.48550/arXiv.2602.08234},
  url           = {https://arxiv.org/abs/2602.08234},
  eprint        = {2602.08234},
  archiveprefix = {arXiv},
  primaryclass  = {cs.AI},
  howpublished  = {arXiv preprint}
}

@misc{xu2026adaskill,
  author        = {Xu, Binyan and Fang, Dong and Li, Haitao and Zhang, Kehuan},
  title         = {From Multi-Agent to Single-Agent: When Is Skill Distillation Beneficial?},
  year          = {2026},
  doi           = {10.48550/arXiv.2604.01608},
  url           = {https://arxiv.org/abs/2604.01608},
  eprint        = {2604.01608},
  archiveprefix = {arXiv},
  primaryclass  = {cs.AI},
  howpublished  = {arXiv preprint}
}

@misc{yuan2026clawtrace,
  author        = {Yuan, Boqin and Su, Yue and Song, Renchu and Yang, Sen and Qin, Jing},
  title         = {{ClawTrace}: Cost-Aware Tracing for {LLM} Agent Skill Distillation},
  year          = {2026},
  doi           = {10.48550/arXiv.2604.23853},
  url           = {https://arxiv.org/abs/2604.23853},
  eprint        = {2604.23853},
  archiveprefix = {arXiv},
  primaryclass  = {cs.AI},
  howpublished  = {arXiv preprint}
}

@misc{zhang2026memskill,
  author        = {Zhang, Haozhen and Long, Quanyu and Bao, Jianzhu and Feng, Tao and Zhang, Weizhi and Yue, Haodong and Wang, Wenya},
  title         = {{MemSkill}: Learning and Evolving Memory Skills for Self-Evolving Agents},
  year          = {2026},
  doi           = {10.48550/arXiv.2602.02474},
  url           = {https://arxiv.org/abs/2602.02474},
  eprint        = {2602.02474},
  archiveprefix = {arXiv},
  primaryclass  = {cs.CL},
  howpublished  = {arXiv preprint}
}

@misc{fan2024workflowllm,
  author        = {Fan, Shengda and Cong, Xin and Fu, Yuepeng and Zhang, Zhong and Zhang, Shuyan and Liu, Yuanwei and Wu, Yesai and Lin, Yankai and Liu, Zhiyuan and Sun, Maosong},
  title         = {{WorkflowLLM}: Enhancing Workflow Orchestration Capability of Large Language Models},
  year          = {2024},
  doi           = {10.48550/arXiv.2411.05451},
  url           = {https://arxiv.org/abs/2411.05451},
  eprint        = {2411.05451},
  archiveprefix = {arXiv},
  primaryclass  = {cs.CL},
  howpublished  = {arXiv preprint}
}

@inproceedings{gou2024critic,
  author        = {Gou, Zhibin and Shao, Zhihong and Gong, Yeyun and Shen, Yelong and Yang, Yujiu and Duan, Nan and Chen, Weizhu},
  title         = {{CRITIC}: Large Language Models Can Self-Correct with Tool-Interactive Critiquing},
  booktitle     = {International Conference on Learning Representations},
  year          = {2024},
  url           = {https://arxiv.org/abs/2305.11738}
}

@misc{hao2026flowscout,
  author        = {Hao, Shuo and Lu, You and Chen, Bihuan and Peng, Xin},
  title         = {{FlowScout}: From Execution Feedback to Reliable Tool-Using Agent Workflows},
  year          = {2026},
  doi           = {10.48550/arXiv.2608.10039},
  url           = {https://arxiv.org/abs/2608.10039},
  eprint        = {2608.10039},
  archiveprefix = {arXiv},
  howpublished  = {arXiv preprint}
}

@inproceedings{hu2024adas,
  author        = {Hu, Shengran and Lu, Cong and Clune, Jeff},
  title         = {Automated Design of Agentic Systems},
  booktitle     = {International Conference on Learning Representations},
  year          = {2025},
  url           = {https://proceedings.iclr.cc/paper_files/paper/2025/hash/36b7acf6f6010652b3f2a433774a66fe-Abstract-Conference.html}
}

@misc{liu2023dylan,
  author        = {Liu, Zijun and Zhang, Yanzhe and Li, Peng and Liu, Yang and Yang, Diyi},
  title         = {A Dynamic {LLM}-Powered Agent Network for Task-Oriented Agent Collaboration},
  year          = {2023},
  doi           = {10.48550/arXiv.2310.02170},
  url           = {https://arxiv.org/abs/2310.02170},
  eprint        = {2310.02170},
  archiveprefix = {arXiv},
  primaryclass  = {cs.CL},
  howpublished  = {arXiv preprint}
}

@misc{ong2024routellm,
  author        = {Ong, Isaac and Almahairi, Amjad and Wu, Vincent and Chiang, Wei-Lin and Wu, Tianhao and Gonzalez, Joseph E. and Kadous, M. Waleed and Stoica, Ion},
  title         = {{RouteLLM}: Learning to Route {LLM}s with Preference Data},
  year          = {2024},
  doi           = {10.48550/arXiv.2406.18665},
  url           = {https://arxiv.org/abs/2406.18665},
  eprint        = {2406.18665},
  archiveprefix = {arXiv},
  primaryclass  = {cs.LG},
  howpublished  = {arXiv preprint}
}

@misc{zhang2024aflow,
  author        = {Zhang, Jiayi and Xiang, Jinyu and Yu, Zhaoyang and Teng, Fengwei and Chen, Xionghui and Chen, Jiaqi and Zhuge, Mingchen and Cheng, Xin and Hong, Sirui and Wang, Jinlin and Zheng, Bingnan and Liu, Bang and Luo, Yuyu and Wu, Chenglin},
  title         = {{AFlow}: Automating Agentic Workflow Generation},
  year          = {2024},
  doi           = {10.48550/arXiv.2410.10762},
  url           = {https://arxiv.org/abs/2410.10762},
  eprint        = {2410.10762},
  archiveprefix = {arXiv},
  primaryclass  = {cs.AI},
  howpublished  = {arXiv preprint}
}

@misc{zhang2024agentprune,
  author        = {Zhang, Guibin and Yue, Yanwei and Li, Zhixun and Yun, Sukwon and Wan, Guancheng and Wang, Kun and Cheng, Dawei and Yu, Jeffrey Xu and Chen, Tianlong},
  title         = {Cut the Crap: An Economical Communication Pipeline for {LLM}-Based Multi-Agent Systems},
  year          = {2024},
  doi           = {10.48550/arXiv.2410.02506},
  url           = {https://arxiv.org/abs/2410.02506},
  eprint        = {2410.02506},
  archiveprefix = {arXiv},
  primaryclass  = {cs.MA},
  howpublished  = {arXiv preprint}
}

@misc{zhao2025a2flow,
  author        = {Zhao, Mingming and Wei, Xiaokang and Shao, Yuanqi and Zhou, Kaiwen and Yang, Lin and Rao, Siwei and Zhan, Junhui and Chen, Zhitang},
  title         = {{$A^2$Flow}: Automating Agentic Workflow Generation via Self-Adaptive Abstraction Operators},
  year          = {2025},
  doi           = {10.48550/arXiv.2511.20693},
  url           = {https://arxiv.org/abs/2511.20693},
  eprint        = {2511.20693},
  archiveprefix = {arXiv},
  primaryclass  = {cs.AI},
  howpublished  = {arXiv preprint}
}

@misc{chen2024magdi,
  author        = {Chen, Justin Chih-Yao and Saha, Swarnadeep and Stengel-Eskin, Elias and Bansal, Mohit},
  title         = {{MAGDi}: Structured Distillation of Multi-Agent Interaction Graphs Improves Reasoning in Smaller Language Models},
  year          = {2024},
  doi           = {10.48550/arXiv.2402.01620},
  url           = {https://arxiv.org/abs/2402.01620},
  eprint        = {2402.01620},
  archiveprefix = {arXiv},
  primaryclass  = {cs.CL},
  howpublished  = {arXiv preprint}
}

@inproceedings{andriushchenko2025agentharm,
  author        = {Andriushchenko, Maksym and Souly, Alexandra and Dziemian, Mateusz and Duenas, Derek and Lin, Maxwell and Wang, Justin and Hendrycks, Dan and Zou, Andy and Kolter, Zico and Fredrikson, Matt and Gal, Yarin and Davies, Xander},
  title         = {AgentHarm: A Benchmark for Measuring Harmfulness of LLM Agents},
  booktitle     = {International Conference on Learning Representations},
  volume        = {2025},
  pages         = {79185--79220},
  year          = {2025},
  url           = {https://proceedings.iclr.cc/paper_files/paper/2025/hash/c493d23af93118975cdbc32cbe7323f5-Abstract-Conference.html}
}

@article{bender2018datastatements,
  author        = {Bender, Emily M. and Friedman, Batya},
  title         = {Data Statements for Natural Language Processing: Toward Mitigating System Bias and Enabling Better Science},
  journal       = {Transactions of the Association for Computational Linguistics},
  volume        = {6},
  pages         = {587--604},
  publisher     = {MIT Press},
  address       = {Cambridge, MA},
  year          = {2018},
  doi           = {10.1162/tacl_a_00041},
  url           = {https://aclanthology.org/Q18-1041/}
}

@misc{cemri2025mast,
  author        = {Cemri, Mert and Pan, Melissa Z. and Yang, Shuyi and Agrawal, Lakshya A. and Chopra, Bhavya and Tiwari, Rishabh and Keutzer, Kurt and Parameswaran, Aditya and Klein, Dan and Ramchandran, Kannan and Zaharia, Matei and Gonzalez, Joseph E. and Stoica, Ion},
  title         = {Why Do Multi-Agent {LLM} Systems Fail?},
  year          = {2025},
  doi           = {10.48550/arXiv.2503.13657},
  url           = {https://arxiv.org/abs/2503.13657},
  eprint        = {2503.13657},
  archiveprefix = {arXiv},
  howpublished  = {arXiv preprint}
}

@inproceedings{debenedetti2024agentdojo,
  author        = {Debenedetti, Edoardo and Zhang, Jie and Balunovic, Mislav and Beurer-Kellner, Luca and Fischer, Marc and Tramer, Florian},
  title         = {{AgentDojo}: A Dynamic Environment to Evaluate Prompt Injection Attacks and Defenses for {LLM} Agents},
  booktitle     = {Advances in Neural Information Processing Systems, Datasets and Benchmarks Track},
  volume        = {37},
  pages         = {82895--82920},
  year          = {2024},
  doi           = {10.52202/079017-2636},
  url           = {https://proceedings.neurips.cc/paper_files/paper/2024/hash/97091a5177d8dc64b1da8bf3e1f6fb54-Abstract-Datasets_and_Benchmarks_Track.html}
}

@inproceedings{deyoung2020eraser,
  author        = {DeYoung, Jay and Jain, Sarthak and Rajani, Nazneen Fatema and Lehman, Eric and Xiong, Caiming and Socher, Richard and Wallace, Byron C.},
  title         = {{ERASER}: A Benchmark to Evaluate Rationalized {NLP} Models},
  booktitle     = {Proceedings of the 58th Annual Meeting of the Association for Computational Linguistics},
  pages         = {4443--4458},
  publisher     = {Association for Computational Linguistics},
  address       = {Online},
  year          = {2020},
  doi           = {10.18653/v1/2020.acl-main.408},
  url           = {https://aclanthology.org/2020.acl-main.408/}
}

@misc{exgentic2026,
  author        = {Bandel, Elron and Yehudai, Asaf and Eden, Lilach and Sagron, Yehoshua and Perlitz, Yotam and Venezian, Elad and Razinkov, Natalia and Ergas, Natan and Ifergan, Shlomit Shachor and Shlomov, Segev and Jacovi, Michal and Choshen, Leshem and Ein-Dor, Liat and Katz, Yoav and Shmueli-Scheuer, Michal},
  title         = {General Agent Evaluation},
  year          = {2026},
  doi           = {10.48550/arXiv.2602.22953},
  url           = {https://arxiv.org/abs/2602.22953},
  eprint        = {2602.22953},
  archiveprefix = {arXiv},
  howpublished  = {arXiv preprint}
}

@article{gebru2021datasheets,
  author        = {Gebru, Timnit and Morgenstern, Jamie and Vecchione, Briana and Vaughan, Jennifer Wortman and Wallach, Hanna and Daume III, Hal and Crawford, Kate},
  title         = {Datasheets for Datasets},
  journal       = {Communications of the ACM},
  volume        = {64},
  number        = {12},
  pages         = {86--92},
  publisher     = {Association for Computing Machinery (ACM)},
  year          = {2021},
  doi           = {10.1145/3458723},
  url           = {https://doi.org/10.1145/3458723}
}

@misc{harnessbench2026,
  author        = {Yao, Yilun and Tan, Xinyu and Liu, Chao-Hsuan and Li, Yaoming and Wang, Zhengyang and Yu, Wenhan and Tan, Zhewen and Tian, Yuxuan and Zhao, Guangxiang and Sun, Lin and Zhang, Xiangzheng and Yang, Tong},
  title         = {{Harness-Bench}: Measuring Harness Effects across Models in Realistic Agent Workflows},
  year          = {2026},
  doi           = {10.48550/arXiv.2605.27922},
  url           = {https://arxiv.org/abs/2605.27922},
  eprint        = {2605.27922},
  archiveprefix = {arXiv},
  howpublished  = {arXiv preprint}
}

@misc{hu2025memoryagentbench,
  author        = {Hu, Yuanzhe and Wang, Yu and McAuley, Julian},
  title         = {Evaluating Memory in {LLM} Agents via Incremental Multi-Turn Interactions},
  year          = {2025},
  doi           = {10.48550/arXiv.2507.05257},
  url           = {https://arxiv.org/abs/2507.05257},
  eprint        = {2507.05257},
  archiveprefix = {arXiv},
  howpublished  = {arXiv preprint}
}

@inproceedings{kornblith2019similarity,
  author        = {Kornblith, Simon and Norouzi, Mohammad and Lee, Honglak and Hinton, Geoffrey},
  title         = {Similarity of Neural Network Representations Revisited},
  booktitle     = {International Conference on Machine Learning},
  year          = {2019},
  url           = {https://arxiv.org/abs/1905.00414}
}

@inproceedings{levy2024stwebagentbench,
  author        = {Levy, Ido and Wiesel, Ben and Marreed, Sami and Oved, Alon and Yaeli, Avi and Mashkif, Nir and Shlomov, Segev},
  title         = {{ST-WebAgentBench}: A Benchmark for Evaluating Safety and Trustworthiness in Web Agents},
  booktitle     = {International Conference on Learning Representations},
  year          = {2026},
  url           = {https://arxiv.org/abs/2410.06703}
}

@misc{li2026skillsbench,
  author        = {Li, Xiangyi and Liu, Yimin and Chen, Wenbo and You, Bingran and Di, Zonglin and He, Yifeng and Zheng, Shenghan and Choe, Kyoung Whan and Sun, Jiankai and Wang, Shuyi and Tao, Chujun and Li, Binxu and Zhao, Xuandong and Geng, Hejia and Wu, Xiaojun and Zhou, Junwei and Chen, Xiaokun and Xing, Hanwen and Li, Yubo and Zeng, Qunhong and Wang, Di and Wang, Yuanli and Chaim, Roey Ben and Jiang, Penghao and Shen, Haotian and Kong, Luyang and Liu, Xinyi and Wang, Runhui and Liu, Xuanqing and Li, Jiachen and Lan, Xin and Lin, Yueqian and Ye, Wengao and He, Junwei and Li, Songlin and Zhang, Yue and Gao, Yipeng and Li, Yijiang and Ma, Ze and Jing, Liqiang and Wang, Tianyu and Li, Kaixin and Xue, Yiqi and Lyu, Haoran and He, Yizhuo and Tian, Yuchen and Wu, Shutong and Wang, Bowei and Gao, Yixuan and Chen, Bo and Liu, Litong and Cheng, Sikai and Bao, Jiajun and Tong, Shuaicheng and Xu, Shuwen and Zhuo, Terry Yue and Ye, Tinghan and Qi, Qi and Li, Miao and Liao, Longtai and Tan, Zelin and Shi, Chang and Tang, Xilin and Tankasala, Srinath and Yuan, Boqin and Qian, Yaoyao and Tu, Jianhong and Wang, Chenguang and Sun, Yizhou and Wang, Wei and Taylor, Aaron and Yang, Ziyue and Guan, Changkun and Dong, Zhikang and Zhang, Xinyu and Dillmann, Steven and Lee, Han-chung and Song, Dawn},
  title         = {{SkillsBench}: Benchmarking How Well Agent Skills Work Across Diverse Tasks},
  year          = {2026},
  doi           = {10.48550/arXiv.2602.12670},
  url           = {https://arxiv.org/abs/2602.12670},
  eprint        = {2602.12670},
  archiveprefix = {arXiv},
  howpublished  = {arXiv preprint}
}

@inproceedings{liu2023geval,
  author        = {Liu, Yang and Iter, Dan and Xu, Yichong and Wang, Shuohang and Xu, Ruochen and Zhu, Chenguang},
  title         = {{G-Eval}: {NLG} Evaluation Using {GPT-4} with Better Human Alignment},
  booktitle     = {Proceedings of the 2023 Conference on Empirical Methods in Natural Language Processing},
  pages         = {2511--2522},
  publisher     = {Association for Computational Linguistics},
  address       = {Singapore},
  year          = {2023},
  doi           = {10.18653/v1/2023.emnlp-main.153},
  url           = {https://aclanthology.org/2023.emnlp-main.153/}
}

@inproceedings{lu2025toolsandbox,
  author        = {Lu, Jiarui and Holleis, Thomas and Zhang, Yizhe and Aumayer, Bernhard and Nan, Feng and Bai, Haoping and Ma, Shuang and Ma, Shen and Li, Mengyu and Yin, Guoli and Wang, Zirui and Pang, Ruoming},
  title         = {{ToolSandbox}: A Stateful, Conversational, Interactive Evaluation Benchmark for {LLM} Tool Use Capabilities},
  booktitle     = {Findings of the Association for Computational Linguistics: NAACL 2025},
  pages         = {1160--1183},
  publisher     = {Association for Computational Linguistics},
  address       = {Albuquerque, New Mexico},
  year          = {2025},
  doi           = {10.18653/v1/2025.findings-naacl.65},
  url           = {https://aclanthology.org/2025.findings-naacl.65/},
  isbn          = {979-8-89176-195-7}
}

@inproceedings{ma2024agentboard,
  author        = {Ma, Chang and Zhang, Junlei and Zhu, Zhihao and Yang, Cheng and Yang, Yujiu and Jin, Yaohui and Lan, Zhenzhong and Kong, Lingpeng and He, Junxian},
  title         = {{AgentBoard}: An Analytical Evaluation Board of Multi-turn {LLM} Agents},
  booktitle     = {Advances in Neural Information Processing Systems, Datasets and Benchmarks Track},
  volume        = {37},
  pages         = {74325--74362},
  year          = {2024},
  doi           = {10.52202/079017-2365},
  url           = {https://proceedings.neurips.cc/paper_files/paper/2024/hash/877b40688e330a0e2a3fc24084208dfa-Abstract-Datasets_and_Benchmarks_Track.html}
}

@inproceedings{patil2025bfcl,
  author        = {Patil, Shishir G. and Mao, Huanzhi and Yan, Fanjia and Ji, Charlie Cheng-Jie and Suresh, Vishnu and Stoica, Ion and Gonzalez, Joseph E.},
  title         = {The Berkeley Function Calling Leaderboard ({BFCL}): From Tool Use to Agentic Evaluation of Large Language Models},
  booktitle     = {Proceedings of the 42nd International Conference on Machine Learning},
  series        = {Proceedings of Machine Learning Research},
  volume        = {267},
  pages         = {48371--48392},
  publisher     = {PMLR},
  year          = {2025},
  url           = {https://proceedings.mlr.press/v267/patil25a.html}
}

@inproceedings{pushkarna2022datacards,
  author        = {Pushkarna, Mahima and Zaldivar, Andrew and Kjartansson, Oddur},
  title         = {Data Cards: Purposeful and Transparent Dataset Documentation for Responsible AI},
  booktitle     = {Proceedings of the 2022 ACM Conference on Fairness, Accountability, and Transparency},
  pages         = {1776--1826},
  publisher     = {ACM},
  year          = {2022},
  doi           = {10.1145/3531146.3533231},
  url           = {https://doi.org/10.1145/3531146.3533231}
}

@inproceedings{stanton2021does,
  author        = {Stanton, Samuel and Izmailov, Pavel and Kirichenko, Polina and Alemi, Alexander A. and Wilson, Andrew G.},
  title         = {Does Knowledge Distillation Really Work?},
  booktitle     = {Advances in Neural Information Processing Systems},
  volume        = {34},
  pages         = {6906--6919},
  year          = {2021},
  url           = {https://proceedings.neurips.cc/paper/2021/hash/376c6b9ff3bedbbea56751a84fffc10c-Abstract.html}
}

@inproceedings{trivedi2024appworld,
  author        = {Trivedi, Harsh and Khot, Tushar and Hartmann, Mareike and Manku, Ruskin and Dong, Vinty and Li, Edward and Gupta, Shashank and Sabharwal, Ashish and Balasubramanian, Niranjan},
  title         = {AppWorld: A Controllable World of Apps and People for Benchmarking Interactive Coding Agents},
  booktitle     = {Proceedings of the 62nd Annual Meeting of the Association for Computational Linguistics (Volume 1: Long Papers)},
  pages         = {16022--16076},
  publisher     = {Association for Computational Linguistics},
  address       = {Bangkok, Thailand},
  year          = {2024},
  doi           = {10.18653/v1/2024.acl-long.850},
  url           = {https://aclanthology.org/2024.acl-long.850/}
}

@inproceedings{turpin2023unfaithful,
  author        = {Turpin, Miles and Michael, Julian and Perez, Ethan and Bowman, Samuel R.},
  title         = {Language Models Don't Always Say What They Think: Unfaithful Explanations in Chain-of-Thought Prompting},
  booktitle     = {Advances in Neural Information Processing Systems},
  volume        = {36},
  pages         = {74952--74965},
  year          = {2023},
  doi           = {10.52202/075280-3275},
  url           = {https://proceedings.neurips.cc/paper_files/paper/2023/hash/ed3fea9033a80fea1376299fa7863f4a-Abstract.html}
}

@inproceedings{wu2025longmemeval,
  author        = {Wu, Di and Wang, Hongwei and Yu, Wenhao and Zhang, Yuwei and Chang, Kai-Wei and Yu, Dong},
  title         = {{LongMemEval}: Benchmarking Chat Assistants on Long-Term Interactive Memory},
  booktitle     = {International Conference on Learning Representations},
  year          = {2025},
  url           = {https://arxiv.org/abs/2410.10813}
}

@inproceedings{xiao2024flowbench,
  author        = {Xiao, Ruixuan and Ma, Wentao and Wang, Ke and Wu, Yuchuan and Zhao, Junbo and Wang, Haobo and Huang, Fei and Li, Yongbin},
  title         = {{FlowBench}: Revisiting and Benchmarking Workflow-Guided Planning for {LLM}-based Agents},
  booktitle     = {Findings of the Association for Computational Linguistics: EMNLP 2024},
  pages         = {10883--10900},
  publisher     = {Association for Computational Linguistics},
  address       = {Miami, Florida, USA},
  year          = {2024},
  doi           = {10.18653/v1/2024.findings-emnlp.638},
  url           = {https://aclanthology.org/2024.findings-emnlp.638/}
}

@inproceedings{xie2024osworld,
  author        = {Xie, Tianbao and Zhang, Danyang and Chen, Jixuan and Li, Xiaochuan and Zhao, Siheng and Cao, Ruisheng and Hua, Toh Jing and Cheng, Zhoujun and Shin, Dongchan and Lei, Fangyu and Liu, Yitao and Xu, Yiheng and Zhou, Shuyan and Savarese, Silvio and Xiong, Caiming and Zhong, Victor and Yu, Tao},
  title         = {{OSWorld}: Benchmarking Multimodal Agents for Open-Ended Tasks in Real Computer Environments},
  booktitle     = {Advances in Neural Information Processing Systems, Datasets and Benchmarks Track},
  volume        = {37},
  pages         = {52040--52094},
  year          = {2024},
  doi           = {10.52202/079017-1650},
  url           = {https://proceedings.neurips.cc/paper_files/paper/2024/hash/5d413e48f84dc61244b6be550f1cd8f5-Abstract-Datasets_and_Benchmarks_Track.html}
}

@inproceedings{yao2025taubench,
  author        = {Yao, Shunyu and Shinn, Noah and Razavi, Pedram and Narasimhan, Karthik},
  title         = {{$\tau$-bench: A Benchmark for Tool-Agent-User Interaction in Real-World Domains}},
  booktitle     = {International Conference on Learning Representations},
  volume        = {2025},
  pages         = {9965--10017},
  year          = {2025},
  url           = {https://proceedings.iclr.cc/paper_files/paper/2025/hash/1b126cc38b8638e07bef37e7b2bb72bf-Abstract-Conference.html}
}

@inproceedings{zhang2025asb,
  author        = {Zhang, Hanrong and Huang, Jingyuan and Mei, Kai and Yao, Yifei and Wang, Zhenting and Zhan, Chenlu and Wang, Hongwei and Zhang, Yongfeng},
  title         = {Agent Security Bench ({ASB}): Formalizing and Benchmarking Attacks and Defenses in {LLM}-Based Agents},
  booktitle     = {International Conference on Learning Representations},
  year          = {2025},
  url           = {https://openreview.net/forum?id=V4y0CpX4hK}
}

@article{zhang2026generalizability,
  author        = {Zhang, Minxing and Yang, Yi and Xie, Roy and Dhingra, Bhuwan and Zhou, Shuyan and Pei, Jian},
  title         = {Generalizability of Large Language Model-Based Agents: A Comprehensive Survey},
  journal       = {ACM Computing Surveys},
  volume        = {58},
  number        = {10},
  pages         = {263:1--263:44},
  publisher     = {Association for Computing Machinery (ACM)},
  year          = {2026},
  doi           = {10.1145/3794858},
  url           = {https://doi.org/10.1145/3794858}
}

@inproceedings{zheng2023judging,
  author        = {Zheng, Lianmin and Chiang, Wei-Lin and Sheng, Ying and Zhuang, Siyuan and Wu, Zhanghao and Zhuang, Yonghao and Lin, Zi and Li, Zhuohan and Li, Dacheng and Xing, Eric P. and Zhang, Hao and Gonzalez, Joseph E. and Stoica, Ion},
  title         = {Judging {LLM}-as-a-Judge with {MT-Bench} and Chatbot Arena},
  booktitle     = {Advances in Neural Information Processing Systems},
  volume        = {36},
  pages         = {46595--46623},
  year          = {2023},
  doi           = {10.52202/075280-2020},
  url           = {https://papers.neurips.cc/paper_files/paper/2023/hash/91f18a1287b398d378ef22505bf41832-Abstract-Datasets_and_Benchmarks.html}
}

@inproceedings{zheng2025processbench,
  author        = {Zheng, Chujie and Zhang, Zhenru and Zhang, Beichen and Lin, Runji and Lu, Keming and Yu, Bowen and Liu, Dayiheng and Zhou, Jingren and Lin, Junyang},
  title         = {{ProcessBench}: Identifying Process Errors in Mathematical Reasoning},
  booktitle     = {Proceedings of the 63rd Annual Meeting of the Association for Computational Linguistics (Volume 1: Long Papers)},
  pages         = {1009--1024},
  publisher     = {Association for Computational Linguistics},
  address       = {Vienna, Austria},
  year          = {2025},
  doi           = {10.18653/v1/2025.acl-long.50},
  url           = {https://aclanthology.org/2025.acl-long.50/},
  isbn          = {979-8-89176-251-0}
}

@inproceedings{zhong2026skilllearnbench,
  author        = {Zhong, Shanshan and Lu, Yi and Ning, Jingjie and Wan, Yibing and Feng, Lihan and Ao, Yuyi and Ribeiro, Leonardo F. R. and Dreyer, Markus and Ammirati, Sean and Xiong, Chenyan},
  title         = {{SkillLearnBench}: Benchmarking Continual Learning Methods for Agent Skill Generation on Real-World Tasks},
  booktitle     = {Conference on Language Modeling},
  year          = {2026},
  url           = {https://arxiv.org/abs/2604.20087}
}

@inproceedings{zhu2025multiagentbench,
  author        = {Zhu, Kunlun and Du, Hongyi and Hong, Zhaochen and Yang, Xiaocheng and Guo, Shuyi and Wang, Zhe and Wang, Zhenhailong and Qian, Cheng and Tang, Xiangru and Ji, Heng and You, Jiaxuan},
  title         = {{MultiAgentBench}: Evaluating the Collaboration and Competition of {LLM} Agents},
  booktitle     = {Proceedings of the 63rd Annual Meeting of the Association for Computational Linguistics (Volume 1: Long Papers)},
  pages         = {8580--8622},
  publisher     = {Association for Computational Linguistics},
  address       = {Vienna, Austria},
  year          = {2025},
  doi           = {10.18653/v1/2025.acl-long.421},
  url           = {https://aclanthology.org/2025.acl-long.421/},
  isbn          = {979-8-89176-251-0}
}

\end{document}